\documentclass{article}

\usepackage[dvipsnames]{xcolor}
\definecolor{linkcolor}{rgb}{0,0.6,0.6}
\usepackage[colorlinks=true,
    linkcolor=linkcolor,
    citecolor=linkcolor,
    filecolor=linkcolor,
    urlcolor=linkcolor,
    pagebackref=true]{hyperref} 
\hypersetup{
    colorlinks=true,
    linkcolor=linkcolor,
    filecolor=magenta,      
    urlcolor=ForestGreen,
    }
\renewcommand*\backref[1]{\ifx#1\relax \else (Cit. on p. #1) \fi}

\usepackage{amsfonts}
\usepackage{nicefrac}       %
\usepackage{amsmath}
\usepackage{amssymb}
\usepackage{cancel}
\usepackage{mathtools}
\usepackage{bbm}
\usepackage{bm}

\usepackage{microtype}      %

\usepackage{enumitem}

\DeclarePairedDelimiter{\norm}{\lVert}{\rVert}

\newcommand{\bg}[1]{\boldsymbol{#1}}

\newcommand{\vect}[1]{\mathbf{#1}}

\newcommand{\vz}{\vect{z}}
\newcommand{\vx}{\vect{x}}

\newcommand{\vthetat}{\bg{\tilde{\theta}}}
\newcommand{\vthetah}{\bg{\hat{\theta}}}
\newcommand{\vtheta}{\bg{\theta}}
\newcommand{\vphi}{\boldsymbol{\phi}}
\newcommand{\vlambda}{\bg{\lambda}}

\newcommand{\fobj}{\mathfrak{f}_{\text{obj}}}
\newcommand{\gconst}{\mathfrak{g}_{\text{const}}}

\newcommand{\Lag}{\mathfrak{L}}

\newcommand{\lambdapen}{\lambda_{\text{pen}}}
\newcommand{\lambdaco}{\lambda_{\text{co}}}

\newcommand{\gatelr}{\eta_{\text{primal}}^{\vphi}}
\newcommand{\weightlr}{\eta_{\text{primal}}^{\vthetat}}
\newcommand{\duallr}{\eta_{\text{dual}}}

\newcommand{\blobletter}[1]{\raisebox{.5pt}{\textcircled{\raisebox{-.8pt}{{\hspace{-1mm} \small #1}}}}}

\usepackage{microtype}
\usepackage{graphicx}
\usepackage{subfigure}
\usepackage{booktabs} %

\usepackage{amsmath}
\usepackage{amssymb}
\usepackage{mathtools}
\usepackage{amsthm}

\usepackage[capitalize]{cleveref}

\theoremstyle{plain}

\theoremstyle{definition}

\theoremstyle{remark}

\usepackage{multirow}
\usepackage{float}
\floatstyle{plaintop}
\restylefloat{table}
\usepackage[tableposition=top]{caption}

\usepackage[symbol]{footmisc}

\usepackage[textsize=tiny, colorinlistoftodos]{todonotes}

\newcounter{todocounter}

\usepackage{tikz}
\usepackage{pgfplots}
\pgfplotsset{compat=1.17}
\usetikzlibrary{decorations.markings}

\usepackage{scrwfile}
\TOCclone[\appendixname]{toc}{atoc}
\newcommand\StartAppendixEntries{}
\AfterTOCHead[toc]{%
  \renewcommand\StartAppendixEntries{\value{tocdepth}=-10000\relax}%
}
\AfterTOCHead[atoc]{%
  \edef\maintocdepth{\the\value{tocdepth}}%
  \value{tocdepth}=-10000\relax%
  \renewcommand\StartAppendixEntries{\value{tocdepth}=\maintocdepth\relax}%
}
\newcommand*\appendixwithtoc{%
  \appendix
  \renewcommand{\baselinestretch}{1.3}\normalsize  %
  \addtocontents{toc}{\protect\StartAppendixEntries}
  \listofatoc
  \renewcommand{\baselinestretch}{1.0}\normalsize %
}
\PassOptionsToPackage{numbers}{natbib}
    \usepackage[final]{neurips2022}

\usepackage[utf8]{inputenc} %
\usepackage[T1]{fontenc}    %
\usepackage{hyperref}       %
\usepackage{url}            %
\usepackage{booktabs}       %
\usepackage{amsfonts}       %
\usepackage{nicefrac}       %
\usepackage{microtype}      %
\usepackage{xcolor}         %

\title{Controlled Sparsity via Constrained Optimization or: {\Large How I Learned to Stop Tuning Penalties and Love Constraints}}

\author{
    \vspace{-7mm} \\
   \textbf{Jose Gallego-Posada}\thanks{Correspondence to: \texttt{\{gallegoj, juan.ramirez, akram.erraqabi\}@mila.quebec}} \hspace{5mm}  \textbf{Juan Ramirez} \hspace{5mm} \textbf{Akram Erraqabi} \\ 
     \vspace{-3mm} \\
   \textbf{Yoshua Bengio}$^\ddagger$ \hspace{5mm} \textbf{Simon Lacoste-Julien}$^\ddagger$\\
   \vspace{-3mm} \\
   Mila and DIRO, Université de Montréal, Canada \\
   $^\ddagger$ Canada CIFAR AI Chair
}

\begin{document}

\maketitle

\vspace{-3ex}

\begin{abstract}
\vspace{-1ex}
The performance of trained neural networks is robust to harsh levels of pruning. Coupled with the ever-growing size of deep learning models, this observation has motivated extensive research on learning sparse models. In this work, we focus on the task of controlling the level of sparsity when performing sparse learning. 
Existing methods based on sparsity-inducing penalties involve expensive trial-and-error tuning of the penalty factor, thus lacking direct control of the resulting model sparsity.
In response, we adopt a \emph{constrained} formulation: using the gate mechanism proposed by \citet{louizos2017learning}, we formulate a constrained optimization problem where sparsification is guided by the training objective and the desired sparsity target in an end-to-end fashion. 
Experiments on CIFAR-\{10, 100\}, TinyImageNet, and ImageNet using WideResNet and ResNet\{18, 50\} models validate the effectiveness of our proposal and demonstrate that we can reliably achieve pre-determined sparsity targets without compromising on predictive performance.
\end{abstract}

\vspace{-3ex}

\section{Introduction}
\label{sec:introduction}

Commonly used neural networks result in \textit{overparametrized} models, whose performance is robust to harsh levels of parameter pruning \citep{han2015deepcompression, ullrich2017soft, frankle2019lottery, gale2019state}. Thus, regularization techniques aimed at learning sparse models can drastically reduce the computational cost associated with the learnt model by removing unnecessary parameters, and retain good performance in the learning task. Given the recent research trends which explore the capabilities of ever more ambitious large-scale models \citep{brown2020language}, developing techniques which provide reliable training of \textit{sparsified} models becomes crucial for deploying them in massively-used systems, or on resource-constrained devices.

Pruning methods aim to reduce the storage and/or computational footprint of a model by discarding individual parameters \citep{han2015deepcompression, molchanov2017variational} or groups thereof \citep{li2017l1pruning, lemaire2019structured, neklyudov2017structured}, while inducing minimal distortion in the model's predictions. These methods can be further categorized based on whether the sparse model is obtained \textit{while} or \textit{after} training the model (also known as \textit{in-training} and \textit{post-training} sparsification).

Traditional post-training methods rely on heuristic rankings of the weights or filters to be pruned, often based on parameter magnitudes \citep{lecun1990optimal, han2015deepcompression}. Despite their simplicity, these methods usually require retraining the weights to maintain high accuracy after pruning, and thus incur in additional computational overhead. On the other hand, in-training methods which \textit{learn} a good sparsity pattern by augmenting the training loss with sparsity-inducing penalties \citep{louizos2017learning, lemaire2019structured} do not perform fine-tuning, but face challenges regarding the tuning and interpretability of the penalty hyperparameter.

In this work\footnote{Our code is available at: {\small \texttt{\url{https://github.com/gallego-posada/constrained_sparsity}}}}, we focus on the task of learning models with \emph{controlled} levels of sparsity while performing in-training pruning. We tackle two central issues of the popular penalized method of \citet{louizos2017learning}: \blobletter{1} tuning the $L_0$-penalty coefficient to achieve a desired sparsity level is non trivial and can involve computationally wasteful trial-and-error attempts; \blobletter{2} in the worst case the penalized method can outright fail at producing any sparsity, as documented by \citet{gale2019state}. 

\newpage

\begin{figure}[h]
\vspace{-8ex}
\setlength\tabcolsep{0pt}%
\centering
\begin{tabular}{llccc}
 \centering
 & \hspace{2mm} & \hspace{-2mm} \textbf{$L_0$-density} & \hspace{6mm}\textbf{Parameters} & \hspace{5mm} \textbf{MACs} \\
 
\rotatebox{90}{\hspace{1.3cm} \textbf{Layer-wise}} & & \includegraphics[width=0.345\textwidth, trim={0mm 2mm 0mm 2mm}, clip]{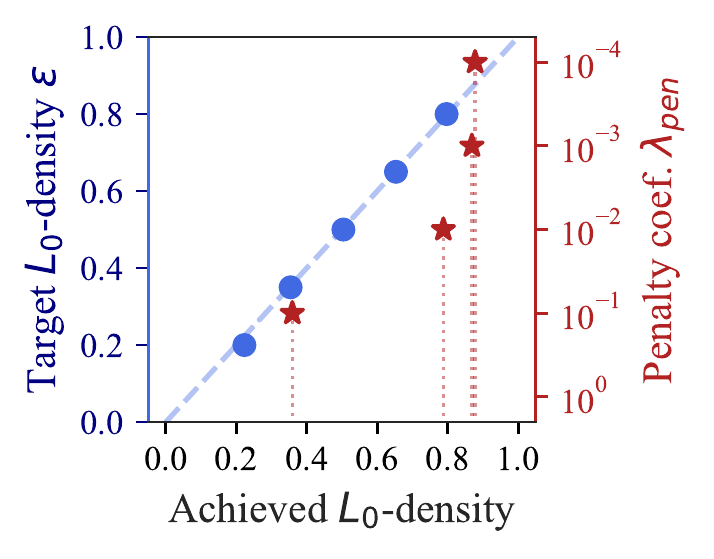}
 & \includegraphics[width=0.27\textwidth, trim={0mm 2mm 0mm 2mm}, clip]{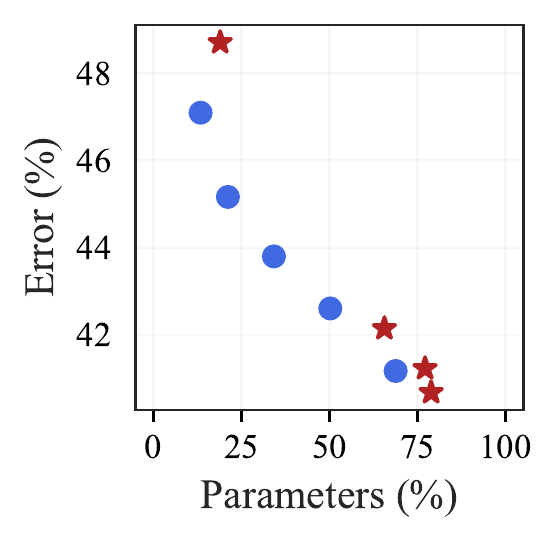}
 & \includegraphics[width=0.27\textwidth, trim={0mm 2mm 0mm 2mm}, clip]{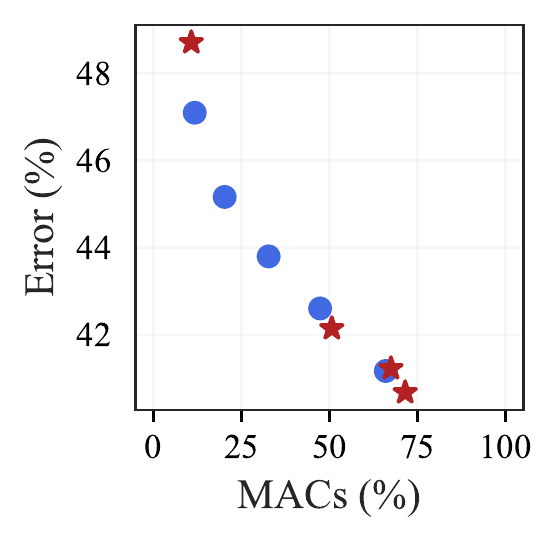} \\
\rotatebox{90}{\hspace{1.3cm} \textbf{Model-wise}} &  & \includegraphics[width=0.345\textwidth, trim={0mm 2mm 0mm 2mm}, clip]{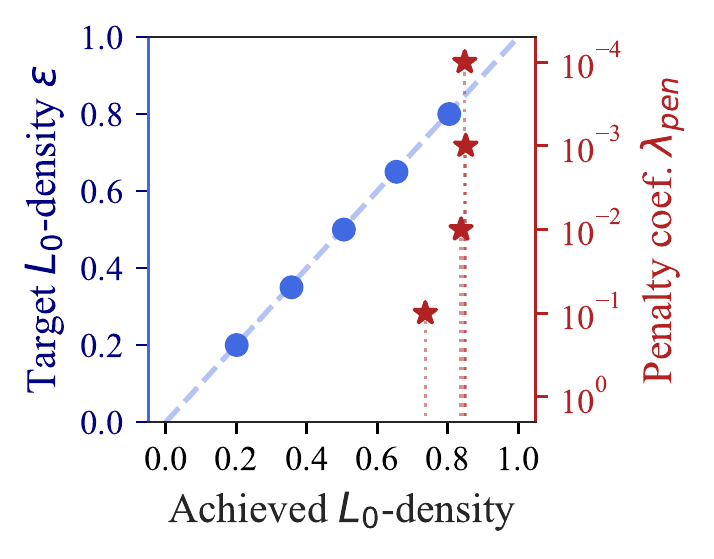}
 & \includegraphics[width=0.27\textwidth, trim={0mm 2mm 0mm 2mm}, clip]{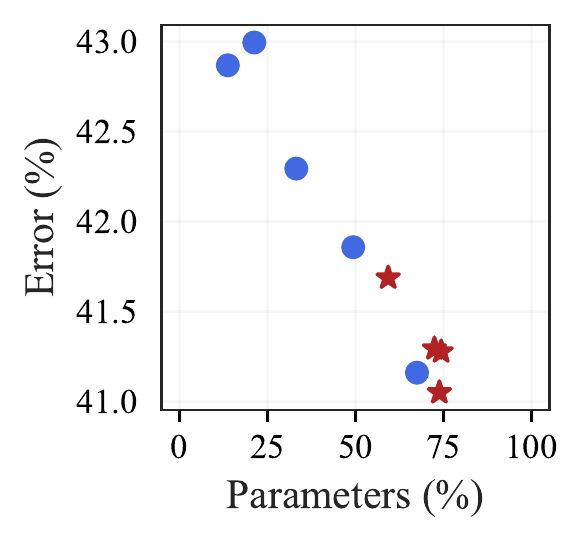}
 & \includegraphics[width=0.27\textwidth, trim={0mm 2mm 0mm 2mm}, clip]{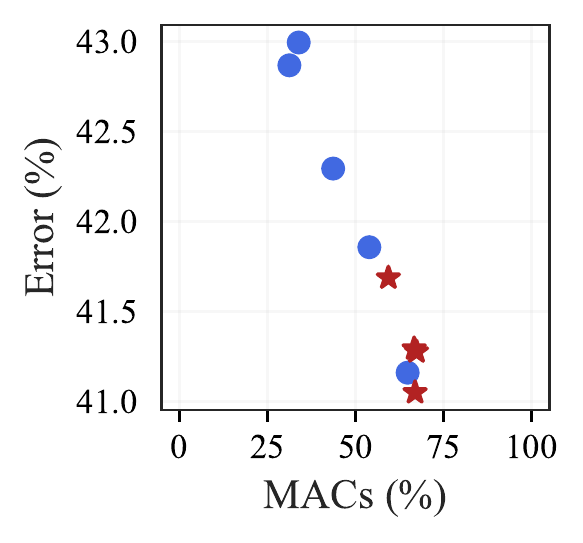}
\end{tabular}
    \vspace{-1ex}
    \caption[]{Training sparse ResNet18 models on TinyImageNet \citep{tinyImagenet}. Density denotes the proportion of active gates in the model. The penalty-based method (red) shows a stagnating-then-overshooting behavior, making it difficult to tune. In contrast, our proposed constrained approach (blue) reliably achieves the desired target $L_0$-densities. The diagonal denotes the ideal case in which the achieved density exactly matches the target density used in the constrained setting. Parameters and MACs are computed for the corresponding test-time \textit{purged} networks following the procedure described in \cref{app:purging_models}; the $L_0$-density (see \cref{eq:const_problem}) is computed for the train-time model.}
    \label{fig:tiny_imagenet_comparison}
\end{figure}

To address these limitations, we propose a constrained optimization approach in which \textbf{\emph{arbitrary sparsity targets are expressed as constraints}} on the $L_0$-norm of the parameters.
Formally, we consider constraints of the type $\norm{\vtheta_g}_0 \le K$, where $\vtheta_g$ represents a group $g$ of parameters of the network (e.g. individual layers, or the whole model), and resort to well established gradient-based methods for optimizing the Lagrangian associated with the constrained optimization problem. 

Adopting this constrained formulation provides several advantages:
\begin{itemize}[noitemsep,topsep=-2pt]
    \item Unlike the multiplicative factor $\lambda$ of a penalty term, the constraint level $\epsilon_g$ has straight-forward and \textbf{interpretable semantics} associated with the \textit{density} of a block of parameters $\vtheta_g$, i.e. the percentage of active parameters. 
    \item Requiring different density levels for different parameter groups (e.g. lower density for network modules with a larger computational or memory footprint), simply amounts to specifying several constraints with levels matching these desired densities, thus \textbf{avoiding the costly process of trial-and-error tuning}\footnote{\cref{app:bisection} shows that the tuning challenges of penalized methods exist even for simple MLP tasks.} and re-balancing various penalty factors. 
    \item Much like the penalized approach in which additional regularizers can be ``stacked'' as other additive terms in the objective, new desired properties can be expressed in the constrained formulation in a \textbf{modular and extensible} fashion as additional constraints. 
    \item In non-convex problems, the constrained formulation \textbf{can be strictly more powerful} than the penalized approach: there may be constraint levels that \textit{cannot be achieved by any value of the penalty coefficient} \citep[$\S 4.7.4$]{boyd2004convex}.
\end{itemize}

The left column of \cref{fig:tiny_imagenet_comparison} illustrates the interpretability and controllability advantages of the constrained approach when training a sparse ResNet18 model on TinyImageNet. We vary the constraint level (left axis) and the penalty coefficient (right axis) and compare the achieved parameter density at the end of training. Note how the penalized approach results in an very dense model ($>80\%$) across several orders of magnitude of the penalty factor, and then suddenly drops to $< 40\%$ density. This behavior is in stark contrast with our proposed constrained approach, which \textit{consistently achieves the desired target density}, across a wide range of values. See \cref{sec:experiments} for further discussion.

The purpose of our paper is to illustrate the feasibility and advantages of using constrained formulations in the study of sparse learning. We favor Lagrangian, gradient-based methods for tackling the constrained optimization problem due to their ease of use and scalability in the context of machine learning models. Exploring alternative constrained optimization techniques is an interesting direction for future studies, but lies beyond the scope of our work.

The main contributions of this work are:
\begin{itemize}[noitemsep,topsep=-2pt]
    \item Building on the work of \citet{louizos2017learning}, we propose a constrained approach for learning models with controllable levels of sparsity (\cref{sec:co_formulation}).
    \item We introduce a \textit{dual restart} heuristic to avoid the excessive regularization caused by the accumulation of constraint violations in gradient-based Lagrangian optimization (\cref{sec:co_formulation}).
    \item Previous studies~\citep{louizos2017learning,gale2019state} have been unsuccessful at training sparse ResNets~\citep{zagoruykoK16} based on $L_0$ regularization without significantly damaging performance. We propose two simple adjustments to the implementation of \citet{louizos2017learning}, allowing us to overcome these challenges (\cref{app:resnet_control_fixes}).
    \item We provide empirical evidence that we can reliably achieve controllable sparsity across many different architectures and datasets. Moreover, the controlability and interpretability benefits of the constrained approach do not come at the expense of achieving competitive predictive performance (\cref{sec:experiments}).
\end{itemize}

\section{Sparsity via \texorpdfstring{$L_0$}{L0} Penalties}
\label{sec:l0_paper}

\citet{louizos2017learning} propose a framework for learning sparse models using the $L_0$-``norm'' of the model parameters as an additive penalty to the usual training objective. The $L_0$-norm counts the \textit{number} of non-zero entries in the parameter vector, and ignores the magnitude of said entries. Consider $h(x;\ \vtheta)$ be a predictor with parameters $\vtheta$ and a supervised learning problem defined by a dataset of $N$ i.i.d. pairs $\mathcal{D} = \{(x_i,\ y_i)\}_{i=1}^N$, a loss function $\ell$ and a regularization coefficient $\lambdapen \geq 0$. \citet{louizos2017learning} formulate the $L_0$-regularized empirical risk objective: 
\vspace{-1ex}
\begin{equation}
    \label{eq:pen_original}
    \mathcal{R}(\vtheta, \lambdapen) = \mathcal{L}_{\mathcal{D}}(\vtheta) + \lambdapen \|\vtheta\|_0 = \frac{1}{N}\left( \sum_{i=1}^N \ell\left( h(x_i;\ \vtheta),\ y_i\right) \right) 
    + \lambdapen \sum_{j=1}^{|\vtheta|} \mathbbm{1}\{\theta_j \neq 0 \}
\end{equation}

\vspace{-2ex}

The non-differentiability of the $L_0$-norm makes it poorly suited for gradient-based optimization. The authors propose a reparametrization $\vtheta = \vthetat \odot \vz$, where $\vthetat$ are free (signed) parameter magnitudes, and $\vz$ are independent stochastic \emph{gates} indicating whether a parameter is active\footnote{Note that the $L_0$-norm of $\vtheta$ is determined by that of $\vz$. This is because for commonly used weight initialization and optimization schemes, $\vthetat \neq 0$ almost surely.}. The authors model the gates using a modified version of the concrete distribution \citep{maddison2016concrete, jang2016categorical}, with parameters denoted by $\vphi$. 

This reparametrization allows for gradient-based optimization procedures, while retaining the possibility of achieving \textit{exact} zeros in the parameters values. We provide a brief overview of the properties of the concrete distribution in \cref{app:gates}.

Moreover, this stochastic reparametrization induces a distribution over the network parameters $\vtheta$. In consequence, the authors propose to re-define the training objective as the expectation (under the distribution of the gates) of the $L_0$-regularized empirical risk in Eq. \eqref{eq:pen_original}:  
\begin{equation}
    \label{eq:pen_expected}
    \mathcal{R}(\vthetat, \vphi, \lambdapen)  \triangleq \mathbb{E}_{\vz \ | \ \vphi} \left[ \mathcal{R}(\vthetat \odot \vz, \lambdapen) \right]  = \mathbb{E}_{\vz \ | \ \vphi} \left[ \mathcal{L}_{\mathcal{D}}(\vthetat \odot \vz) \right] + \lambdapen \mathbb{E}_{\vz \ | \ \vphi} \left[ \|\vz\|_0 \right]
\end{equation}

\paragraph{Test-time model.} Since the stochastic reparametrization induces a distribution over models, \citet{louizos2017learning} propose a protocol to choose a sparse network at test time. We employ a slightly modified version of their strategy, based on the \textit{medians} of the gates (see \cref{app:gates:validation_gates}).

\paragraph{Parameter grouping.} Rather than considering a gate for each individual parameter (which would double the number of trainable parameters), several parameters may be gathered under a shared gate. We match the setup of \citet{louizos2017learning} who focus on \textit{neuron sparsity}: using
\raisebox{.5pt}{\textcircled{\raisebox{-.8pt}{{\hspace{-1mm} \small 1}}}} one gate per \textit{input neuron} for fully connected layers; and
\raisebox{.5pt}{\textcircled{\raisebox{-.8pt}{{\hspace{-1mm} \small 2}}}} one gate per \textit{output feature map} for convolutional layers. This use of \textit{structured sparsity} results in practical storage and computation improvements since entire parameter groups (e.g. slice of convolution kernels/activation) can be discarded.

\paragraph{Combining other norms.} \citet{louizos2017learning} show that their reparametrization can used in conjunction with other commonly used norms for regularization, such as the $L_2$-norm. One can express $\mathbb{E}_{\vz|\vphi}\left[ \|\vthetah\|^2_2 \right] = \sum_{j=1}^{|\vtheta|} \mathbb{P}[ z_j \neq 0 ] \,  \tilde{\theta}^2_j$, where $\vthetah$ is a \textit{gate-rescaled} version of $\vtheta$ in order to ``avoid extra shrinkage for the gates''. Further discussion on the challenges of combining weight-decay and their proposed reparametrization can be found in \cref{app:resnet_control_fixes,,app:test_time_gates}.

\section{Sparsity via \texorpdfstring{$L_0$}{L0} Constraints}
\label{sec:co_formulation}

We favor formulating regularization goals as constraints, rather than as additive penalties with fixed scaling factors. We refer to these two approaches as \textit{constrained} and \textit{penalized}, respectively. Although a ubiquitous tool in machine learning, penalized formulations may come at the cost of hyper-parameter interpretability and are susceptible to intricate dynamics when incorporating multiple, potentially conflicting, sources of regularization.

\subsection{Constrained Formulation}
  
In contrast to the penalized objective of \citet{louizos2017learning} presented in \cref{eq:pen_expected}, we propose to incorporate sparsity through constraints on the $L_0$-norm. We formulate an optimization problem that aims to minimize the model's expected empirical risk, subject to constraints on the expected $L_0$-norm of pre-determined parameter groups:
\vspace{-1.5ex}
\begin{align}
    \label{eq:const_problem}
    \underset{\boldsymbol{\tilde{\theta}}, \vphi}{\text{min}} \, \fobj(\vthetat, \vphi) \triangleq \mathbb{E}_{\vz | \vphi} \left[ \mathcal{L}_{\mathcal{D}}(\vthetat \odot \vz) \right] \hspace{2mm}
    \text{s.t.} \hspace{2mm}  \gconst(\vphi_g) \triangleq \overbrace{\frac{\mathbb{E}_{\vz_g | \vphi_g} \left[ \|\vz_g\|_0 \right]}{\#(\vthetat_g)}}^{L_0-\text{density}}  \le \epsilon_g \hspace{2mm} \text{for } g \in [1:G], 
\end{align}
where $g$ denotes a subset of gates, $\#(\vx)$ counts the total number of entries in $\vx$, and $\vx_g$ denotes the entries of a vector $\vx$ associated with the group $g$. See \cref{app:grouping} for details on parameter grouping.

Note how the $\#(\vthetat_g)$ factor in the constraint levels allows us to \textbf{interpret $\epsilon_g$} as the maximum \textit{proportion} of gates that are allowed to be active within group $g$, in expectation. We refer to $\epsilon_g$ as the \textbf{\textit{target density}} of group $g$. Lowering the target density demands a sparser model and thus a (not necessarily strictly) more challenging optimization problem in terms of the best \textit{feasible} empirical risk. Moreover, for any choice of $\epsilon_g \ge 0$, the feasible set in \cref{eq:const_problem} is always non-empty; while values of $\epsilon_g \ge 1$ result in vacuous constraints.

We highlight one important difference between the constrained and penalized formulations. The penalized approach is \emph{jointly} optimizing the training loss $\fobj(\vthetat, \vphi)$ \emph{and} the expected $L_0$-norm $\gconst(\vphi)$, due to their additive combination (mediated by $\lambdapen$). Meanwhile, the constrained method focuses on obtaining the best possible model within a prescribed density level $\epsilon$: given two \textit{feasible} solutions, the constrained formulation in \cref{eq:const_problem} only discriminates based on the training loss. In other words, \textbf{we aim to \textit{satisfy} the constraints, not to \textit{optimize} them}.

\subsection{Solving the Constrained Optimization Problem}
\label{sec:solve_const}

We start by considering the (nonconvex-concave) Lagrangian associated with the constrained formulation in \cref{eq:const_problem}, along with the corresponding min-max game:
\vspace{-1ex}
\begin{equation}
    \label{eq:minmax_game}
    \vthetat^*, \vphi^*, \vlambda_{\text{co}}^* \triangleq    \underset{\vthetat, \vphi}{\text{argmin}}\ \underset{\vlambda_{\text{co}} \ge 0}{\text{argmax}} \, \,
    \Lag(\vthetat, \vphi, \vlambda_{\text{co}}) \triangleq \fobj(\vthetat, \vphi)  + \sum_{g=1}^G \lambdaco^g \left( \gconst(\vphi_g) - \epsilon_g \right),
\end{equation}
where $\vlambda_{\text{co}} = [\lambdaco^g]_{g=1}^G$ are the 
(non-negative) Lagrange multipliers associated with each constraint.

A commonly used approach to optimize this Lagrangian is \textit{simultaneous} gradient descent on ($\vthetat,\vphi$) and projected (to $\mathbb{R}^+)$ gradient ascent on $\vlambda_{\text{co}}$ \citep{lin2020gradient}:

\vspace{-4ex}

\begin{align}
    \label{eq:gda_updates}
    [\vthetat^{t+1}, \vphi^{t+1}] &\triangleq [\vthetat^t, \vphi^t] \ {\color{red} -}\ \eta_{\text{primal}} \,  \nabla_{[\vthetat, \vphi]} \Lag(\vthetat^t, \vphi^t, \vlambda_{\text{co}}^t) \nonumber \\
    \hat{\vlambda}^{t+1} &\triangleq  \vlambda_{\text{co}}^{t}\ {\color{red} +} \ \eta_{\text{dual}} \,  \nabla_{\vlambda_{\text{co}}} \Lag(\vthetat^t, \vphi^t, \vlambda_{\text{co}}^t) = \vlambda_{\text{co}}^{t} + \eta_{\text{dual}} \, \left[ \gconst(\vphi_g^t) - \epsilon_g \right]_{g=1}^G \\
    \vlambda_{\text{co}}^{t+1} &\triangleq \max \left( 0, \hat{\vlambda}^{t+1} \right) \nonumber
\end{align}

\vspace{-2ex}

 The gradient update for $\vlambda_{\text{co}}$ matches the value of the violation of each constraint. When a constraint is satisfied, the gradient for its corresponding Lagrange multiplier is non-positive, leading to a reduction in the value of the multiplier.

\textbf{Negligible computational overhead.} Just as the penalized formulation of \citet{louizos2017learning}, the update for $\vthetat$ and $\vphi$ requires the gradient of the training loss and that of the expected $L_0$-norm. Hence, the cost of executing this update scheme is the same as the cost of a gradient descent update on the penalized formulation in \cref{eq:pen_original}, up to the negligible cost of updating the multipliers. 

\textbf{Choice of optimizers.} We present simple gradient descent-ascent (GDA) updates in \cref{eq:gda_updates}. However, our proposed framework is compatible with different choices for the primal and dual optimizers, including stochastic methods. Throughout our experiments, we opt for primal (model) optimizers which match standard choices for the different architectures. A choice of gradient ascent for the dual optimizer provided consistently robust optimization dynamics across all tasks. Detailed experimental configurations are provided in  \cref{app:exp_details}. The evaluation and design of other optimizers, especially those for updating the Lagrange multipliers, is an interesting direction for future research.

\textbf{Oscillations.} The non-convexity of the optimization problem in \cref{eq:minmax_game} implies that a saddle point (pure strategy Nash Equilibrium) might not exist. In general, this can lead to oscillations and unstable optimization dynamics. \cref{app:constrained_optimization} provides pointers to more sophisticated constrained optimization algorithms which achieve better convergence guarantees on nonconvex-concave problems than GDA. Fortunately, throughout our experiments we observed oscillatory behavior that quickly settled around feasible solutions. Empirical evidence of this claim is presented in \cref{sec:experiments:no_oscillations}.

\textbf{Extensibility.} Our proposed constrained formulation is ``modular'' in the sense that it is easy to induced other properties in the model's behavior beside sparsity (e.g. fairness \citep{hooker2019compressednnsforget, cotter2019}) by prescribing them as \textit{additional constraints}; much like extra additive terms in the penalized formulation. However, the improved interpretability and control afforded by the constrained approach removes the need to perform extensive tuning of the hyper-parameters to balance these potentially competing demands.

\subsection{Dual Restarts}
\label{sec:dual_restarts}

A drawback of gradient-based updates for optimizing the Lagrangian in \cref{eq:minmax_game} is that the constraint violations accumulate in the value of the Lagrange multipliers throughout the optimization, and continue to affect the optimization dynamics, \textit{even after a constraint has been satisfied}. This results in an excessive regularization effect, which forces the primal parameters towards the \textit{interior} of the feasible set. This behavior can be detrimental if we are concerned about minimizing the objective function and \textit{satisfying} (but not minimizing!) the constraints.

To address this, we propose a \textbf{\textit{dual restart scheme}} in which the Lagrange multiplier $\lambdaco^g$ associated with a constraint $\gconst(\vphi_g) \le \epsilon_g$ is set to $0$ whenever the constraint is satisfied; rather than waiting for the ``negative'' gradient updates ($\gconst(\vphi_g) - \epsilon_g < 0$ when feasible) to reduce its value. Formally,
\begin{equation}
    \label{eq:dual_restart}
    \left[ \vlambda_{\text{co}}^{t+1} \right]_g \triangleq  \begin{cases}
\text{max} \left( 0, \, \left[ \vlambda_{\text{co}}^{t} \right]_g + \eta_{\text{dual}} \, \left( \gconst(\vphi_g^t) - \epsilon_g \right) \right), & \text{if} \quad \gconst(\vphi_g^t) > \epsilon_g \\
 0, \hspace{5mm} \text{otherwise}  &
\end{cases}
\end{equation}

\vspace{-1.5ex}
Dual restarts remove the contribution of the expected $L_0$-norm to the Lagrangian for groups $g$ whose constraints are satisfied, so that the optimization may focus on improving the predictive performance of the model. In fact, this dual restart strategy can be theoretically characterized as a \textit{best response} (in the game-theoretic sense) by the dual player. The effect of dual restarts in the optimization dynamics is illustrated in \cref{sec:experiments:dual_restarts,,app:tr_dynamics}.

\vspace{-1ex}

\section{Related Work}

\textbf{Min-max optimization.} Commonly used methods for solving constrained convex optimization problems \citep{boyd2004convex, frankwolfe, jaggi2013revisiting} make assumptions on the properties of the objective function, constraints or feasible set. In this work, we focus on applications involving neural networks, leading to the violation of such assumptions. We rely on a GDA-like updates for optimizing the associated Lagrangian. However, our proposed formulation can be readily integrated with more sophisticated/theoretically supported algorithms for constrained optimization of non-convex-concave objectives, such as the extragradient method \citep{korpelevich1976extragradient}. Further discussion on guarantees and alternative algorithms for min-max optimization is provided in \cref{app:constrained_optimization}.

\textbf{Model sparsity.} Learning sparse models is a rich research area in machine learning. There exist many different approaches for obtaining sparse models. Magnitude-based methods \citep{thimm1995evaluating,han2015deepcompression} perform one or more rounds of pruning, by removing the parameters with the lowest magnitudes. Popular non-magnitude based techniques include \citep{lecun1990optimal, guo2016dynamic, molchanov2017variational}. Structured pruning methods \citep{dai2018vib, neklyudov2017structured, louizos2017learning, li2017l1pruning}, remove entire neurons/channels rather than \textit{individual} parameters.  More recently, the Lottery Ticket Hypothesis \citep{frankle2019lottery} has sparked interest in techniques that provide the storage and computation benefits of sparse models \textit{directly during training} \citep{mostafa2019parameter, evci2020rigl}. However, finding ``good'' sparse sub-networks at initialization remains a central challenge for these techniques \citep{frankle2020,malach2020}.

\textbf{Controllable sparsity.} 
Magnitude pruning \citep{han2015deepcompression, li2017l1pruning} can achieve  arbitrary levels of sparsity ``by design'' since it removes \textit{exactly} the proportion of parameters with lowest magnitudes in order to match the desired density. However, the magnitude pruning method experiences certain shortcomings: \blobletter{1} retaining performance usually involves several round of fine-tuning\footnote{This re-training overhead makes magnitude pruning less appealing compared with in-training alternatives, since magnitude pruning is typically performed given an \textit{already fully trained} model.}; \blobletter{2} it relies on the \textit{assumption} that magnitude (of filters or activations) is a reasonable surrogate for parameter importance; and \blobletter{3} it lacks the ``extensibility'' property of our constrained formulation: it is not immediately evident how to induce other desired properties in the model, besides sparsity.

Note that several extensions of the basic magnitude pruning method have been proposed.  \citet{zhuPruneNotPrune2017} start from a partially or fully pre-trained model and consider a sparsification scheme in which the network density is gradually reduced, while fine-tuning the model to compensate for any potential loss in performance due to pruning. \citet{wang2021NeuralPruning} start by \textit{identifying} the parameters to be removed by applying magnitude pruning on a pre-trained model. However, rather than pruning the model immediately, the authors propose to fine-tune the model with an adaptive $L_2$-penalty. The weight of this penalty is increased over time for the previously identified parameters, leading their magnitudes to decrease during the fine-tuning process.

\textbf{Sparsity via constrained optimization.}  Previous works have cast the task of learning sparse models as the solution of a constrained optimization problem. 
\citet{carreira2018learningcompression} consider a reformulation of the constrained optimization problem using ``auxiliary variables'', and assume that the constraints enjoy an efficient proximal operator. Their empirical evaluation is limited to low-scale models and datasets. 

\citet{zhou2021effective} adopt a constrained formulation similar to ours, although based on a different reparametrization of the gates. The authors tackle the constrained problem via projected gradient descent by cleverly exploiting the existence of an efficient projection of the gate parameters onto their feasible set. However, the applicability of their method is limited to constraints with an efficiently-computable projection operator.

\citet{lemaire2019structured} consider ``budget-aware regularization'' and tackle the constrained problem using a barrier method. Although originally inspired by a constrained approach, their resulting training objective corresponds to a penalized method with a penalty factor that requires tuning, in addition to the choice of barrier function. 

\textbf{Other constrained formulations in ML.} Constrained formulations can be used to prescribe desired behaviors or properties in machine learning models. \citet{nandwani2019PrimalDual} study the problem of training deep models under constraints on the network's predicted labels, and approach the constrained problem in practice through a min-max Lagrangian formulation. Incorporating these constraints during training allows them to inject domain-specific knowledge into their models across several tasks in natural language processing. 

\citet{fioretto2020LagrangianDuality} consider a wide range of applications spanning from optimal power flow in energy grids, to the training of fair classification models. Their work demonstrates how Lagrangian-based methods can be complementary to deep learning by effectively enforcing complex physical and engineering constraints.

\citet{cotter2019} train models under constraints on the prediction rates of the model over different datasets. 
Note that the sparsity constraints we study in this paper depends only on properties of the \textit{parameters} and not on the \textit{predictions} of the model. 
We would like to highlight that the notion of \textit{proxy constraints} introduced by \citet{cotter2019} can enable training models based on constraints on their actual test time density, rather than the surrogate expected $L_0$-norm metric.

\vspace{-2ex}
\section{Experiments}
\label{sec:experiments}

\vspace{-2ex}
The main goal of our work is to train models that attain good predictive performance, while having a fine-grained command on the sparsity of the resulting model. In this section we present a comparison with the work of \citet{louizos2017learning}\footnote{See a comparison to other sparsity methods therein, along with the survey of \citet{gale2019state}.}; we explore the stability and controllability properties of our Lagrangian-based constrained approach, along with the effect of our proposed dual restarts heuristic. Finally, we present empirical evidence which demonstrates that our method successfully retains its interpretability and controllability advantages when applied to large-scale models and datasets.

\vspace{-2ex}
\subsection{Experimental Setup}
\label{sec:experiments:intro}

\vspace{-1ex}
\textbf{Experiment configuration and hyperparameters.} Details on our implementation, hyperparameter choices and information on the network architectures can be found in \cref{app:gates,,app:grouping,,app:normalized_l0,,app:purging_models,,app:exp_details}.  

\textbf{Model- and layer-wise settings.} We present experiments using two kinds of constraints: one \textit{global} constraint on the proportion of active gates throughout the entire model; or several \textit{local} constraints prescribing a maximum density at each layer. Note that for models such as ResNet50, the layer-wise setting involves handling 48 constraints. The experiments below demonstrate that \textbf{our constrained approach can gracefully handle from a single constraint up to dozens of constraints} in a unified way and still achieve controllable sparsity for \textit{each} of the layers/model. This level of control is an intractable goal for penalized methods: as demonstrated in \cref{app:bisection}, even trying to tame \textit{one} constraint via a penalty factor can be prohibitively expensive.

\textbf{$L_0$-regularization for residual models.} ResNets have been a challenging setting for $L_0$-penalty based methods. \citet{gale2019state} trained WideResNets \citep{zagoruykoK16} and ResNet50 \citep{he2015resnets} using the penalized $L_0$-regularization framework of \citet{louizos2017learning}, and reported being unable to produce sparse ResNets without significantly degrading performance.

We propose two simple adjustments that enable us to successfully train WRNs and ResNets with controllable sparsity, while retaining competitive performance: \raisebox{.5pt}{\textcircled{\raisebox{-.8pt}{{\hspace{-1mm} \small 1}}}}  increasing the learning rate of the stochastic gates; and \raisebox{.5pt}{\textcircled{\raisebox{-.8pt}{{\hspace{-1mm} \small 2}}}}  removing the gradient contribution of the weight decay penalty towards the gates.  \cref{app:resnet_control_fixes,,app:test_time_gates} provide detailed analysis and empirical validation of these two modifications. We integrate these adjustments in all experiments involving residual models below.

\textbf{Obtaining test-time models.} \cref{app:purging_models} describes our procedure to transform a model with stochastic gates into a deterministic, test-time model. The measurements of retained parameters and MACs (multiply-accumulate operations) percentages reported in the tables and figures below, are computed for the deterministic, purged, test-time models.

\subsection{Proof of Concept Experiments on MNIST}
\label{sec:experiments:mnist}

We begin by comparing the behavior of our method with that of \citet{louizos2017learning} in the simple setting of training MLP and LeNet5 architectures on the MNIST dataset. The authors report the size of their pruned architectures found using the penalized formulation. In this section we aim to showcase the \textit{controllability} advantages of our constrained approach. We manually computed the corresponding model-wise or layer-wise density levels achieved by the reported architectures of \citet{louizos2017learning} and used these values as the target density levels for our constrained formulation.

\begin{table*}[ht]
    \renewcommand{\arraystretch}{1.05}
    \centering

    \caption{Achieved density levels and performance for sparse MLP and LeNet5 models     trained on MNIST for 200 epochs. Metrics aggregated over 5 runs. {\small $^\dag$Results by \citet{louizos2017learning} with $N$ representing the training set size (see \cref{app:normalized_l0}).}
    }
    \label{tab:mnist}

    \resizebox{\textwidth}{!}{%
    \begin{tabular}{ccclccc}
      \hline
      \multirow{3}{*}{\textbf{Architecture}} & \multirow{3}{*}{\textbf{Grouping}} & \multirow{3}{*}{\textbf{Method}}  & \multirow{3}{*}{\textbf{\hspace{5mm}Hyper-parameters}} & \textbf{Pruned} & \multicolumn{2}{c}{\textbf{Val. Error} (\%)}  \\
      \cline{6-7} & & &  &  \multirow{2}{*}{\textbf{architecture}} & \multirow{2}{*}{best} & at 200 epochs \\
       & & & & &  & (avg {\color{gray} $\pm$ 95\% CI)}\\
      \hline
      \hline
       & \multirow{2}{*}{Model} & Pen.  &  $^\dag\lambda_{pen} = 0.1/N$ &  219-214-100 & 1.4 & --\\
       MLP & & Const. &  $\epsilon = 33 \%$ &  198-233-100 & 1.36 & 1.77 {\color{gray} $\pm$ 0.08}\\
       \cline{2-7}
       784-300-100 & \multirow{2}{*}{Layer} & Pen. &  $^\dag\lambda_{pen} = [0.1,0.1,0.1]/N$ &  266-88-33 & 1.8 & -- \\
      & & Const. &  $\epsilon = [30 \%, 30\%, 30\%]$ &  243-89-29 & 1.58 & 2.19 {\color{gray} $\pm$ 0.12 }\\
       \hline
       & \multirow{2}{*}{Model} & Pen. &   $^\dag\lambda_{pen} = 0.1/N$ &  20-25-45-462 & 0.9 & --\\
       LeNet5 & & Const. &  $\epsilon = 10\%$ &  20-21-34-407 & 0.56 & 1.01 {\color{gray} $\pm$ 0.05}\\
       \cline{2-7}
       20-50-800-500 & \multirow{2}{*}{Layer} & Pen. &  $^\dag\lambda_{pen} = [10,0.5,0.1,0.1]/N$ &  9-18-65-25 & 1.0 & -- \\
       & & Const. &  $\epsilon = [50\%, 30\%, 70\%, 10\%]$ &  10-14-224-29 & 0.7 & 0.91 {\color{gray} $\pm$ 0.05}\\
       \hline
    \end{tabular}
    } %

\end{table*}

\cref{tab:mnist} displays the results of our constrained method and the reported metrics for the penalized approach.
Note that, as desired, the pruned models obtained using the constrained formulation resemble closely the ``target architecture sizes'' reported by \citet{louizos2017learning}.
Moreover, our method does not cause any loss in performance with respect to the penalized approach. This final observation will be confirmed for larger-scale tasks in later sections.

Note that the goal of this section is to demonstrate that our constrained approach can achieve arbitrary sparsity targets ``in one shot'' (i.e. without trial-and-error tuning) and without inducing any compromise in the predictive performance of the resulting models. Comprehensive experiments for MLP and LeNet5 models on MNIST across a wide range of sparsity levels for model- and layer-wise constraints are presented in \cref{app:all_comparisons:mnist}.

\subsection{Training Dynamics and Dual Restarts}
\label{sec:experiments:dual_restarts}

\begin{figure}[h]
    \setlength\tabcolsep{0pt}%
    \centering
    \includegraphics[width=0.97\textwidth]{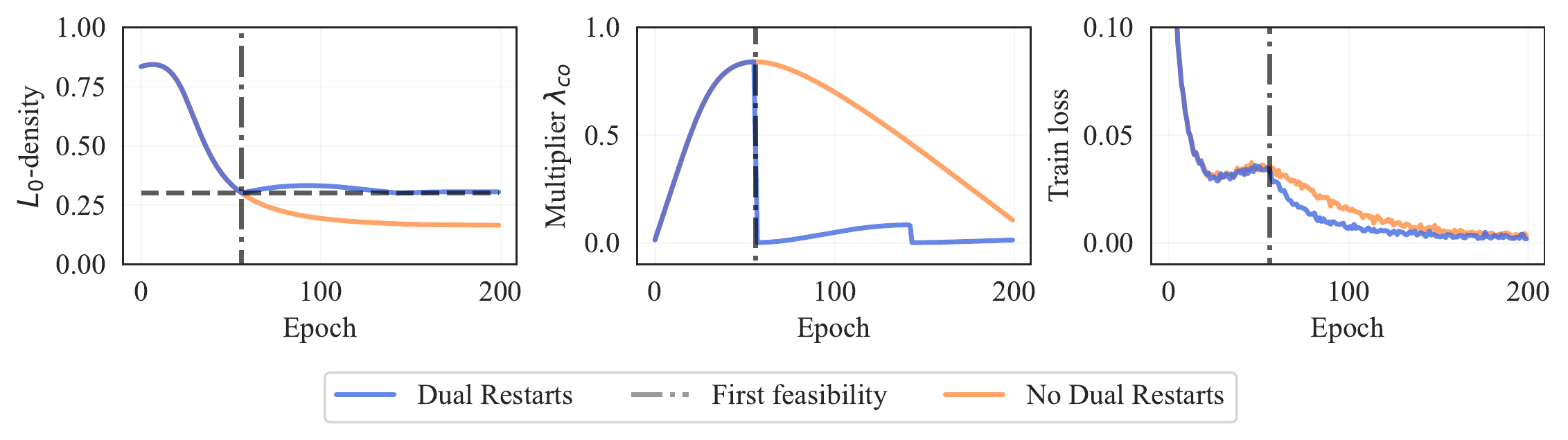}
    \vspace{-1ex}
    \caption[]{Effect of dual restarts for training a LeNet5 on MNIST with a model-wise target density $\epsilon_g = 30\%$ (horizontal dashed line). The accumulation of the constraint violations in the Lagrange multiplier leads to excessive sparsification when the model satisfies the constraint. Restarting the Lagrange multiplier allows the model to concentrate on improving the training loss.}
    \label{fig:dual_restarts}
\end{figure}

We now discuss the effect of the dual restarts scheme introduced in \cref{sec:dual_restarts} on the training dynamics of our constrained formulation. \cref{fig:dual_restarts} illustrates the training of a convolutional network on MNIST under a $30\%$ model-wise density constraint when using (blue) or not (orange) dual restarts. 

\vspace{-1ex}

We initialize the Lagrange multipliers to zero. Therefore, at the beginning of optimization there is no contribution from the $L_0$-norm in the Lagrangian (see \cref{eq:minmax_game}), and the optimization focuses on improving the training loss. As the optimization progresses, the constraint violations are accumulated in the value of the Lagrange multiplier. When the Lagrange multiplier is sufficiently large, the importance of satisfying the constraints outweighs that of optimizing the training loss.\footnote{Note that the Lagrange multiplier influences the update of the model parameters by \textit{dynamically adjusting} the relative importance of the gradient of the training loss with respect to the gradient of the constraint. In the penalized method this relative importance is fixed.}. In consequence, the model density decreases. As the model reaches the desired sparsity level, $\lambdaco$ stops increasing. 

Up until the time at which the model is first feasible, the multiplier value accumulates the constraint violations (scaled by the dual learning rate). Once the model is feasible, the constraint violation $\gconst(\vphi_g) - \epsilon < 0$ becomes negative, leading to a \textit{decrease} in the Lagrange multiplier. However, at this stage, the Lagrange multiplier is large due to the accumulated constraint violations. This confers a higher relative importance to the gradient of the constraints over that of the training loss: the larger multiplier encourages to reduce the \textit{constraints even if they are already being satisfied}. 

Our proposed dual restart heuristic reduces the Lagrange multiplier to zero whenever the constrained is satisfied, allowing the training to focus on minimizing the training loss faster. Although this heuristic may lead to slightly unfeasible solutions, as demonstrated throughout our experiments, our models remain consistently below (or close to) the required $L_0$-density levels.

\subsection{Stable Constraint Dynamics}
\label{sec:experiments:no_oscillations}

\begin{figure}[h]
\vspace{-2ex}
\setlength\tabcolsep{2pt}%
\centering
\begin{tabular}{cc}
 \centering
\includegraphics[width=0.37\textwidth]{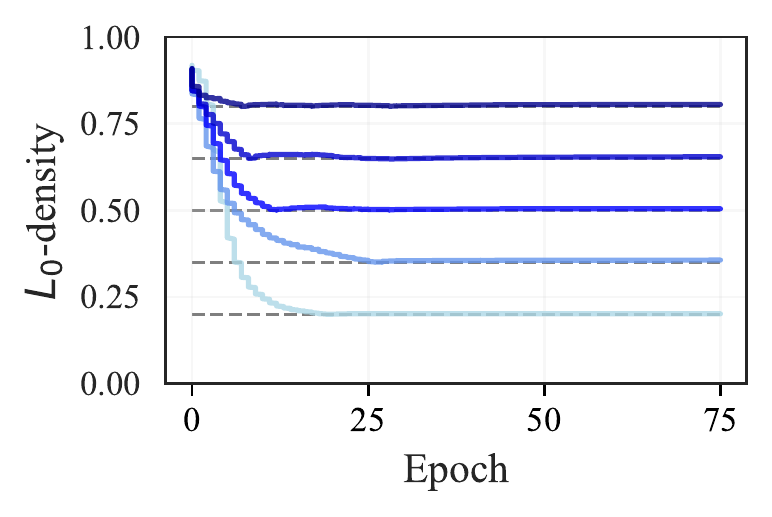} & \includegraphics[width=0.48\textwidth]{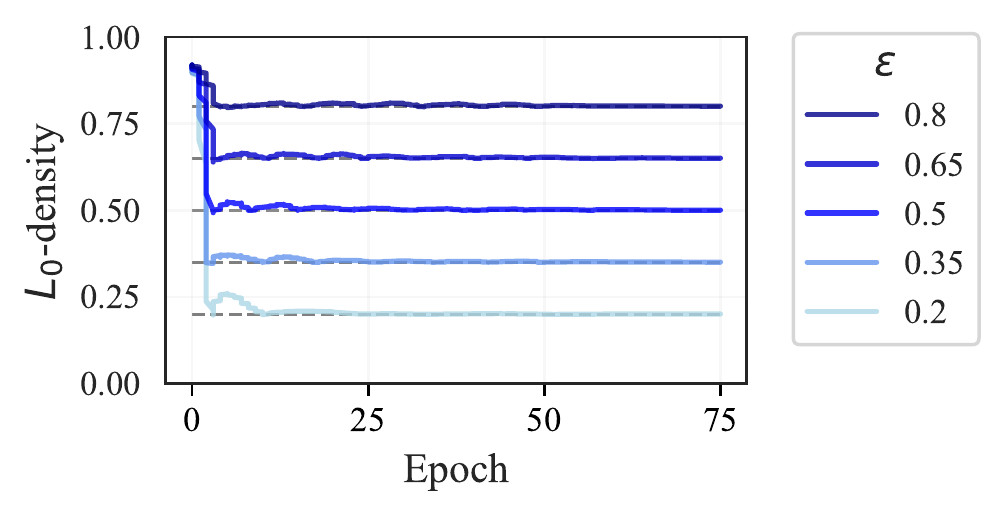} \\
\end{tabular}
    \vspace{-3ex}
    \caption[]{Density levels for a ResNet18 model (left; trained with a model-wise constraint) and the last sparsifiable layer of a WideResNet-28-10 model (right; trained with layer-wise constraints).}
    \label{fig:stable_training}
\end{figure}

Despite the theoretical risk of oscillatory dynamics commonly associated with iterative constrained optimization methods, we consistently observed quickly stabilizing behavior in our experiments. \cref{fig:stable_training} shows the density levels throughout training for a layer of a WideResNet-28-10 trained on CIFAR-10 (right), and the model-wise density of a ResNet18 trained on TinyImageNet (left).

The desired density levels are successfully achieved over a wide range of targets, and the constraint dynamics stabilize quickly.
These dynamics were consistent across all our architectures and datasets.

\subsection{Large-scale Experiments}
\label{sec:experiments:large_models}

We now demonstrate the scalability of our method to more challenging settings: we consider (Wide)ResNet models on the CIFAR-\{10, 100\}, TinyImageNet \citep{tinyImagenet} and ImageNet \citep{imagenet} datasets. Comprehensive experiment are provided in \cref{app:all_comparisons}.

\begin{figure}[h]
\setlength\tabcolsep{0pt}%
\centering
\begin{tabular}{llccc}
 \centering
 & \hspace{2mm} & \hspace{-2mm} \textbf{$L_0$-density} & \hspace{6mm}\textbf{Parameters} & \hspace{5mm} \textbf{MACs} \\
 
\rotatebox{90}{\hspace{1.2cm} \textbf{Layer-wise}} & & \includegraphics[width=0.345\textwidth, trim={0mm 2mm 0mm 2mm}, clip]{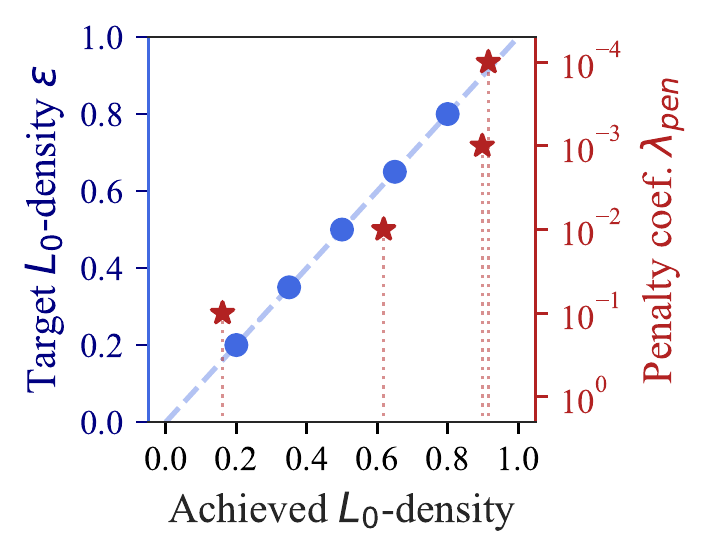}
 & \includegraphics[width=0.27\textwidth, trim={0mm 2mm 0mm 2mm}, clip]{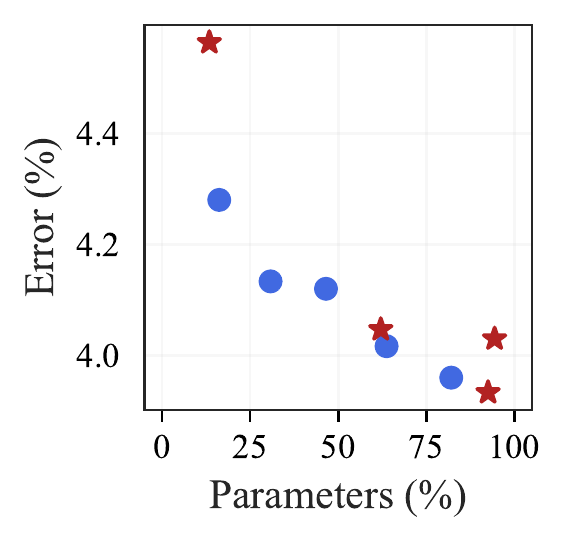}
 & \includegraphics[width=0.27\textwidth, trim={0mm 2mm 0mm 2mm}, clip]{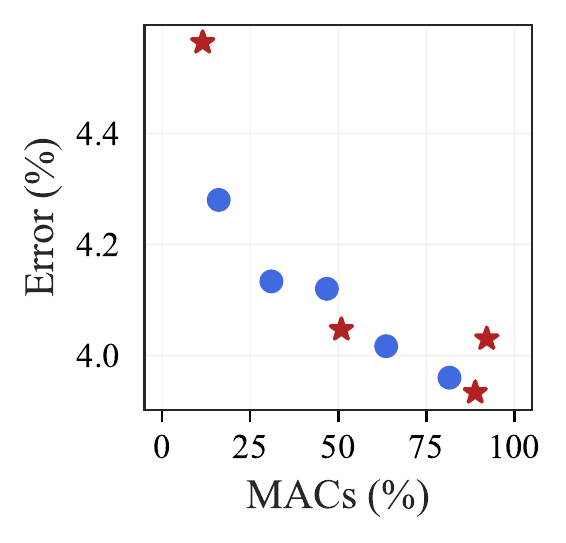} \\
\rotatebox{90}{\hspace{1.2cm} \textbf{Model-wise}} &  & \includegraphics[width=0.345\textwidth, trim={0mm 2mm 0mm 2mm}, clip]{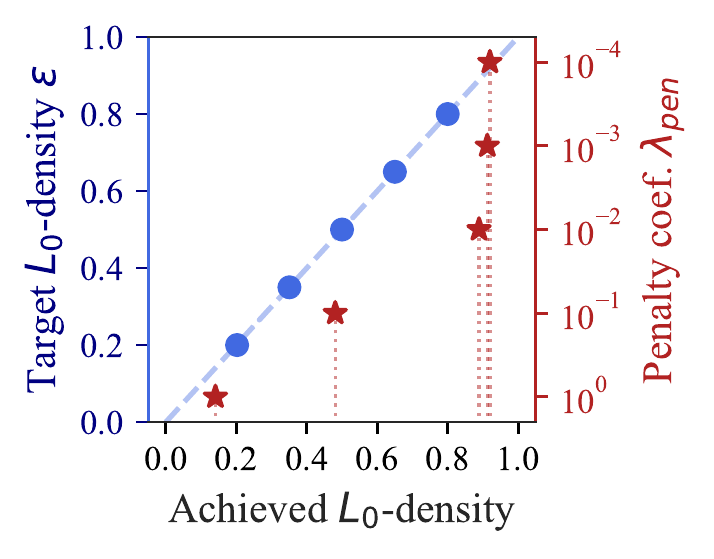}
 & \includegraphics[width=0.27\textwidth, trim={0mm 2mm 0mm 2mm}, clip]{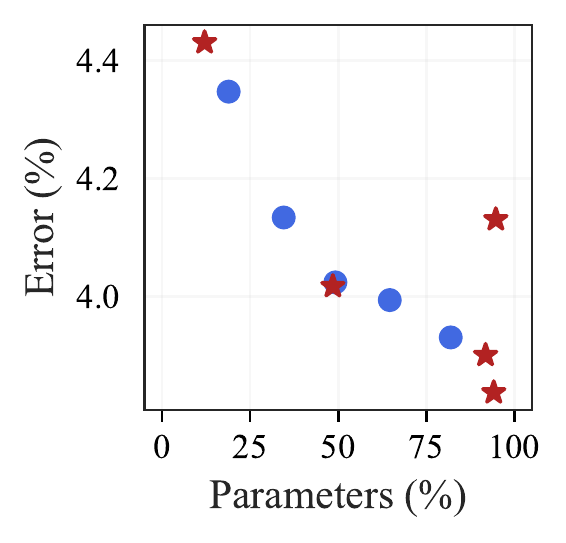}
 & \includegraphics[width=0.27\textwidth, trim={0mm 2mm 0mm 2mm}, clip]{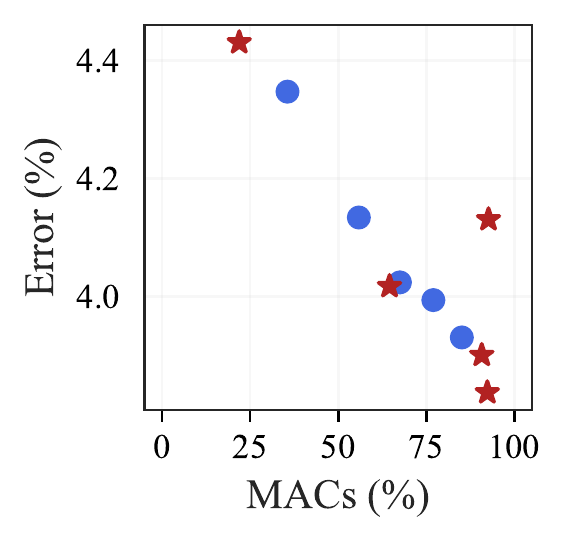} \\
\end{tabular}
    \caption[]{Training sparse WideResNet-28-10 models on CIFAR-10.}
    \label{fig:cifar10_comparison}
\end{figure}

\textbf{CIFAR-\{10, 100\} and TinyImageNet.} Figures \ref{fig:tiny_imagenet_comparison} and \ref{fig:cifar10_comparison} display the results for a ResNet18 model trained on Imagenet, and a WideResNet-28-10 trained on CIFAR-10, respectively. The left column shows the alignment between the achieved and desired densities (as expected proportion of \textit{active gates} in the model). Our method (in blue) provides a  robust control over the range of densities. In contrast, the penalized method (in red) exhibits an unreliable dependency between the penalty coefficient and the achieved density: when increasing the coefficient $\lambdapen$, the achieved density seems to be insensitive to $\lambdapen$ for several orders of magnitude until it starts considerably changing. This brittle sensitivity profile limits the potential of the penalized method for controlling sparsity.

Columns two and three display the number of parameters and MACs (multiply-accumulate operations) of the resulting purged models, as a proportion of those of the fully-dense baseline model. Note that, while retaining a similar proportion of parameters, layer-wise constraints lead to a larger reduction in the number of MACs, compared to the model-wise case. This is because layer-wise constraints induce a strict, homogeneous sparsification of all the modules of the network; while the model-wise setting can allow for a more flexible allocation of the parameter budget across different layers.

\textbf{ImageNet.} We conducted experiments on ImageNet \citep{imagenet} with a ResNet50 architecture. We compare with layer-wise structured magnitude pruning  \citep{li2017l1pruning}\footnote{For \textit{each layer}, we remove the filters with the $(1-\epsilon)$ lowest $L_1$-norms to achieved the desired $\epsilon$ density.}. The results are presented in \cref{tab:mp_comparison_imagenet}.  

Note that experiments with layer-wise constraints correspond to optimization problems with 48 constraints (one for each sparsifiable layer). We highlight the number of constraints since tuning such a large number of penalty coefficients is an intractable challenge when using the penalized method.

\vspace{-2ex}

\begin{table}[h!]
\small
\centering
\renewcommand{\arraystretch}{1.1}

\caption{ResNet50 models on ImageNet with structured sparsity. 
``Fine-tuning'' for zero epochs means \textit{no} fine-tuning.}

\label{tab:mp_comparison_imagenet}
\vspace{1ex}

\resizebox{\textwidth}{!}{%
\begin{tabular}{ccccccccc}
\hline
\multirow{2}{*}{\textbf{Target}} & \multirow{3}{*}{\textbf{Method}}  & \multirow{2}{*}{$L_0$-\textbf{density}} & \multirow{2}{*}{\textbf{Params}} & \multirow{2}{*}{\textbf{MACs}} & \multicolumn{4}{c}{\textbf{Best Val. Error} (\%)} \\
\cline{6-9}
\multirow{2}{*}{\textbf{Density}} & & \multirow{2}{*}{(\%)} & \multirow{2}{*}{(\%)} & \multirow{2}{*}{(\%)} &  \multicolumn{4}{c}{After fine-tuning for \# epochs} \\
& & & & & 0 & 1 & 10 & 20 \\
\hline
\hline
  \multirow{1}{*}{$-$} & \multirow{1}{*}{Pre-trained Baseline} & \multirow{1}{*}{$100$} & \multirow{1}{*}{[25.5M]}  & \multirow{1}{*}{[$4.12 \cdot 10^9$]}  & {\multirow{1}{*}{$23.90$}} & \multicolumn{3}{c}{\multirow{1}{*}{{\color{gray} ----------------------}}} \\
 \hline
\multirow{3}{*}{$\epsilon = 90 \%$} & Const. {\color{gray} \textit{Model-wise}} & \multirow{1}{*}{$90.36$} & \multirow{1}{*}{$88.06$} & \multirow{1}{*}{$91.62$} & {\multirow{1}{*}{$\textbf{24.68}$}} & \multicolumn{3}{c}{\multirow{1}{*}{{\color{gray} ----------------------}}} \\
& Const. {\color{gray} \textit{Layer-wise}} & \multirow{1}{*}{$90.58$} & \multirow{1}{*}{$87.07$} & \multirow{1}{*}{$85.97$} & {\multirow{1}{*}{$24.97$}} & \multicolumn{3}{c}{\multirow{1}{*}{{\color{gray} ----------------------}}} \\
 & L1-MP {\color{gray} \textit{Layer-wise}} & \multirow{1}{*}{$-$} & \multirow{1}{*}{$85.94$} & \multirow{1}{*}{$84.99$} & \multirow{1}{*}{$38.74$} & \multirow{1}{*}{$25.38$} & \multirow{1}{*}{$24.69$} &\multirow{1}{*}{$\textbf{24.68}$} \\
\hline
\multirow{3}{*}{$\epsilon = 70 \%$} & Const. {\color{gray} \textit{Model-wise}} & \multirow{1}{*}{$70.78$} & \multirow{1}{*}{$64.41$} & \multirow{1}{*}{$76.50$} & {\multirow{1}{*}{$\textbf{25.53}$}} & \multicolumn{3}{c}{\multirow{1}{*}{{\color{gray} ----------------------}}} \\
& Const. {\color{gray} \textit{Layer-wise}}& \multirow{1}{*}{$70.36$} & \multirow{1}{*}{$61.91$} & \multirow{1}{*}{$58.59$} & {\multirow{1}{*}{$26.98$}} & \multicolumn{3}{c}{\multirow{1}{*}{{\color{gray} ----------------------}}} \\
& L1-MP {\color{gray} \textit{Layer-wise}} & \multirow{1}{*}{$-$} & \multirow{1}{*}{$62.15$} & \multirow{1}{*}{$59.85$} & \multirow{1}{*}{$97.78$} & \multirow{1}{*}{$29.04$} & \multirow{1}{*}{$26.80$} & \multirow{1}{*}{$26.14$} \\
\hline
\multirow{3}{*}{$\epsilon = 50 \%$} & Const. {\color{gray} \textit{Model-wise}} & \multirow{1}{*}{50.18} & \multirow{1}{*}{42.47} & \multirow{1}{*}{58.00} & {\multirow{1}{*}{\textbf{27.51}}} & \multicolumn{3}{c}{\multirow{1}{*}{{\color{gray} ----------------------}}} \\
& Const. {\color{gray} \textit{Layer-wise}} & \multirow{1}{*}{50.70} & \multirow{1}{*}{43.15} & \multirow{1}{*}{38.25} & {\multirow{1}{*}{27.89}} & \multicolumn{3}{c}{\multirow{1}{*}{{\color{gray} ----------------------}}}\\
& L1-MP {\color{gray} \textit{Layer-wise}} & \multirow{1}{*}{$-$} & \multirow{1}{*}{43.47} & \multirow{1}{*}{39.76} & \multirow{1}{*}{99.75} & \multirow{1}{*}{36.21} & \multirow{1}{*}{29.98} & \multirow{1}{*}{29.16} \\
\hline
\multirow{3}{*}{$\epsilon = 30 \%$} & Const. {\color{gray} \textit{Model-wise}} & \multirow{1}{*}{30.31} & \multirow{1}{*}{31.81} & \multirow{1}{*}{42.05} & {\multirow{1}{*}{\textbf{29.65}}} & \multicolumn{3}{c}{\multirow{1}{*}{{\color{gray} ----------------------}}} \\
& Const. {\color{gray} \textit{Layer-wise}} & \multirow{1}{*}{31.44} & \multirow{1}{*}{30.16} & \multirow{1}{*}{23.74} & {\multirow{1}{*}{31.71}} & \multicolumn{3}{c}{\multirow{1}{*}{{\color{gray} ----------------------}}}\\
& L1-MP {\color{gray} \textit{Layer-wise}} & \multirow{1}{*}{$-$} & \multirow{1}{*}{29.86} & \multirow{1}{*}{24.80} & \multirow{1}{*}{99.89} & \multirow{1}{*}{56.11} & \multirow{1}{*}{36.90} & \multirow{1}{*}{34.74} \\
\hline
\end{tabular}
}
\end{table}

Just like the magnitude pruning method, our proposed approach successfully delivers the desired levels of sparsity in this challenging task. To the best of our knowledge, our work constitutes the first instance of successfully learning ResNet50 models using the $L_0$ reparametrization of \citet{louizos2017learning} for structured sparsity while retaining high accuracy. 

Our results clearly demonstrate that the constrained $L_0$ formulations can obtain large levels of structured parameter reduction while preserving performance. \cref{tab:mp_comparison_imagenet} shows a quick degradation in performance for the magnitude pruning method, and highlights the need for fine-tuning in heuristic-based pruning techniques.

\subsection{Unstructured Sparsity}
\label{sec:experiments:unstructured}

\cref{app:unstructured} contains experiments with unstructured sparsity (i.e. one gate per parameter, rather than per neuron/activation map) for the MNIST and TinyImageNet datasets. These experiments show that the controllability advantages of our constrained formulation apply in the unstructured regime. Recall that \citet{gale2019state} report an apparent dichotomy between sparsity and performance when training (residual) models with unstructured sparsity using the $L_0$ reparametrization of \citet{louizos2017learning}. Our experimental results demonstrate that it is in fact possible to achieve high levels of sparsity \textit{and} predictive performance.

\vspace{-1ex}

\section{Conclusion}

We resort to a constrained optimization approach as a tool to overcome the controllability shortcomings faced by penalty-based sparsity methods. Along with a reliable control of the model density, this technique provides a more interpretable hyper-parameter and removes the need for expensive iterative tuning. We adopt the $L_0$ reparametrization framework of \citet{louizos2017learning} and integrate simple adjustments to remedy their challenges at training (Wide)ResNet models. Our proposed method succeeds at achieving the desired sparsity with no compromise on the model's performance for a broad range of architectures and datasets. These observations position the constrained approach as a solid, practical alternative to popular penalty-based methods in modern machine learning tasks.

\begin{ack}
This work was partially supported by the Canada CIFAR AI Chair Program and by an IVADO PhD Excellence Scholarship.
Simon Lacoste-Julien is a CIFAR Associate Fellow and Yoshua Bengio is a Program Co-Director in the Learning in Machines \& Brains program.
\end{ack}

\bibliography{neurips2022/references.bib}
\bibliographystyle{support_files/abbrvnatClean}

\section*{Checklist}

\begin{enumerate}

\item For all authors...
\begin{enumerate}
  \item Do the main claims made in the abstract and introduction accurately reflect the paper's contributions and scope?
    \answerYes{}
  \item Did you describe the limitations of your work?
    \answerYes{}
  \item Did you discuss any potential negative societal impacts of your work?
    \answerNA{}
  \item Have you read the ethics review guidelines and ensured that your paper conforms to them?
    \answerYes{}
\end{enumerate}

\item If you are including theoretical results...
\begin{enumerate}
  \item Did you state the full set of assumptions of all theoretical results?
    \answerNA{}
        \item Did you include complete proofs of all theoretical results?
    \answerNA{}
\end{enumerate}

\item If you ran experiments...
\begin{enumerate}
  \item Did you include the code, data, and instructions needed to reproduce the main experimental results (either in the supplemental material or as a URL)?
    \answerYes{} Code provided in the supplementary material.
  \item Did you specify all the training details (e.g., data splits, hyperparameters, how they were chosen)?
    \answerYes{} See \cref{app:exp_details}.
    \item Did you report error bars (e.g., with respect to the random seed after running experiments multiple times)?
    \answerYes{}
        \item Did you include the total amount of compute and the type of resources used (e.g., type of GPUs, internal cluster, or cloud provider)?
    \answerYes{} See \cref{app:exp_details,,app:all_comparisons}.
\end{enumerate}

\item If you are using existing assets (e.g., code, data, models) or curating/releasing new assets...
\begin{enumerate}
  \item If your work uses existing assets, did you cite the creators?
    \answerYes{}
  \item Did you mention the license of the assets?
    \answerNA{}
  \item Did you include any new assets either in the supplemental material or as a URL?
    \answerNA{}
  \item Did you discuss whether and how consent was obtained from people whose data you're using/curating?
    \answerNA{}
  \item Did you discuss whether the data you are using/curating contains personally identifiable information or offensive content?
    \answerNA{}
\end{enumerate}

\item If you used crowdsourcing or conducted research with human subjects...
\begin{enumerate}
  \item Did you include the full text of instructions given to participants and screenshots, if applicable?
    \answerNA{}
  \item Did you describe any potential participant risks, with links to Institutional Review Board (IRB) approvals, if applicable?
    \answerNA{}
  \item Did you include the estimated hourly wage paid to participants and the total amount spent on participant compensation?
    \answerNA{}
\end{enumerate}

\end{enumerate}

\newpage

\appendixwithtoc

\section{Reparametrization of the Gates}
\label{app:gates}

\citet{louizos2017learning} introduce the hard concrete distribution for modeling the stochastic gates $\vz$. 
Consider a concrete random variable $s_j \sim q(\cdot\ |\ (\phi_j, \beta))$, given a fixed $0 < \beta < 1$. This variable is then stretched and clamped, resulting in a mixed distribution with point masses at $0$ and $1$, and a continuous density over $(0, 1)$. 

Formally, given $U_j \sim \text{Unif}(0, 1)$ and hyper-parameters $\gamma < 0 < 1 < \zeta$,
 \begin{align}
    \label{eq:hard_concrete}
     &s_j = \text{Sigmoid} \left( \frac{1}{\beta} \log \left( \frac{\phi_j \, U_j}{1-U_j} \right) \right); \hspace{5mm}
     \vz = \text{clamp}_{[0, 1]}( \vect{s} (\zeta - \gamma) + \gamma))
 \end{align}

The stochastic nature of $\vz$ entails a model which is itself \emph{stochastic}. Therefore, both its $L_0$-norm and predictions are random quantities. Nonetheless, the specific choice of re-parameterization in Eq. \eqref{eq:hard_concrete} allows for the training loss in \cref{eq:pen_expected} to be estimated using Monte Carlo samples. 

Moreover, \citet{louizos2017learning} show that the expected $L_0$-norm can be expressed in closed-form as:
\begin{equation}
    \label{eq:exp_l0_closed_form}
    \mathbb{E}_{\vz|\bg{\phi}}\left[\ \|\vtheta\|_0 \right] = \sum_{j=1}^{|\vtheta|} \mathbb{P}[ z_j \neq 0 ]  = \sum_{j=1}^{|\vtheta|} \text{Sigmoid} \left( \log \phi_j - \beta \log \frac{- \gamma}{\zeta} \right)
\end{equation}

The probability distribution of the gates has both learnable and fixed parameters.
\cref{tab:gates_params} specifies the values of the fixed parameters employed throughout this work, following \citet{louizos2017learning}. 

\begin{table}[h]
    \renewcommand{\arraystretch}{1.1}
    \centering
    \begin{tabular}{|c|c|c|c|}
      \hline
      \textbf{Parameter} & $\gamma$ & $\zeta$ & $\beta$ \\
      \textbf{Value} & -0.1 & 1.1 & 2/3 \\
      \hline
    \end{tabular}
    \caption{Fixed parameters of the hard concrete distribution.}
    \label{tab:gates_params}
\end{table}

\subsection{Choice of Gates at Test-Time}
\label{app:gates:validation_gates}

Recall that the stochastic reparametrization induces a distribution over models. We suggest a natural way of ``freezing'' the network gates so as to obtain a \textit{deterministic} predictor when evaluating the model on unseen datapoints: replace each stochastic gate by its median.
\begin{equation}
    \label{eq:validation_gates}
    \hat{z}_j = \text{min}\left(1,\text{ max}\left(0,\ \text{Sigmoid}\left(\frac{\text{log}(\phi_j)} {\beta}\right) (\zeta - \gamma) + \gamma \right)\right)
\end{equation}

Note that these medians may be \textit{fractional}, i.e.,  $z_j \in [0, 1]$. Nonetheless, as shown in \cref{app:test_time_gates}, for trained sparse models we observe the medians to be highly concentrated at 0 and 1. 

\citet[Eq. ($13$)]{louizos2017learning} originally proposed a similar approach for obtaining a deterministic test-time model (without an explicit motivation for their choice). Their proposal differs from ours in that they do not perform a division by $\beta$, and thus their test-time gates are not the median (nor the mean) of the gate distributions. 

In our preliminary experiments the division by $\beta$ (under the settings of \cref{tab:gates_params}) did not induce significant changes in behavior. However, we opt for \cref{eq:validation_gates} in our experiments based on its concise theoretical motivation.

\subsection{Initialization of the Gates}
\label{app:rho_init}

\citet{louizos2017learning} introduce a hyper-parameter $\rho_{\text{init}} \in (0,1)$ which determines the initialization of the parameter $\phi_j$ of the hard concrete distribution of the gates (see \cref{app:gates}). 
Concretely, the gate parameters $\phi_j$ are initialized as:
\begin{equation}
    \log \phi_j = \log \left( \frac{1-\rho_{\text{init}}}{\rho_{\text{init}}} \right) + \mathcal{N}(0, 10^{-2})
\end{equation}

Note that in practice, the optimization variable is $\log \phi_j$ (rather than $\phi_j$) as this sidesteps having to preserve the non-negativity of $\phi_j$.

The choice of $\rho_{\text{init}}$ has an inverse relationship with the probability of a gate being active at initialization. For simplicitly, we ignore the small additive noise in the initialization of $\log \phi_j$, and let $\psi = \left(- \gamma / \zeta \right)^{\beta}$. Formally, the influence of the hyper-parameter $\rho_{\text{init}}$ at initialization is given by:
\begin{equation}
    \label{eq:rho_init_prob}
    \mathbb{P}[ z_j \neq 0 ] = \text{Sigmoid} \left( \log \left( \frac{1-\rho_{\text{init}}}{\rho_{\text{init}}} \right) - \log  \left(\frac{- \gamma}{\zeta} \right)^{\beta} \right) = \frac{1 - \rho_{\text{init}}}{1 - (1- \psi) \rho_{\text{init}}}.
\end{equation}

\section{Schemes for Grouping Parameters}
\label{app:grouping}

We consider two schemes for grouping gates: a) with a single constraint/penalty on the proportion of active gates of the model, or b) with separate constraints/penalties for each layer. These groupings are referred to as \emph{model-wise} and \emph{layer-wise}.

For the penalized method, the corresponding optimization problems are given by:
\vspace{-2ex}

\begin{minipage}{0.45\textwidth}
\begin{align*}
    & \textbf{\text{Model-wise grouping}} & \\
    \underset{\boldsymbol{\tilde{\theta}}, \vphi}{\text{min}}\quad & \fobj(\vthetat, \vphi) + \lambdapen \, \, \gconst(\vphi) 
\end{align*}
\end{minipage}
\begin{minipage}{0.45\textwidth}
\begin{align*}
    & \textbf{\text{Layer-wise grouping}} & \\
    \underset{\boldsymbol{\tilde{\theta}}, \vphi}{\text{min}}\quad & \fobj(\vthetat, \vphi) + \sum_{g = 1}^{\texttt{num\_layers}} \lambdapen^g \, \, \gconst(\vphi_g) 
\end{align*}
\end{minipage}

For the constrained method, the corresponding optimization problems are given by:

\vspace{-2ex}

\begin{minipage}{0.45\textwidth}
\begin{align*}
    & \textbf{\text{Model-wise grouping}} & \\
    \underset{\boldsymbol{\tilde{\theta}}, \vphi}{\text{min}}\quad & \fobj(\vthetat, \vphi)
    \\ \text{s.t.} \hspace{3mm} & \gconst(\vphi) \le \epsilon  
\end{align*}
\end{minipage}
\begin{minipage}{0.45\textwidth}
\begin{align*}
    & \textbf{\text{Layer-wise grouping}} & \\
    \underset{\boldsymbol{\tilde{\theta}}, \vphi}{\text{min}}\quad & \fobj(\vthetat, \vphi)
    \\ \text{s.t.} \hspace{3mm} & \gconst(\vphi_g) \le \epsilon_g \hspace{2mm} \text{for } g \in [1:\texttt{num\_layers}]  
\end{align*}
\end{minipage}

For example, consider a [$d_{\text{in}}, d_{\text{hid}}, d_{\text{out}}]$ 1-hidden layer MLP with input-neuron sparsity on both of its two fully connected layers. For simplicity, we ignore the bias in the description below.
\begin{itemize}
    \item \textbf{Grouping at the layer level} (akin to ``$\lambda$ sep.'' in \citep{louizos2017learning}), would yield $G=2$ groups, with $d_{\text{in}}$ gates in group $g=1$.
    Each gate in group $1$ is shared across $d_{\text{hid}}$ parameters in $\vthetat$, thus $ \#(\vthetat_1) =     d_{\text{in}} \cdot d_{\text{hid}}$. Similarly for $g=2$.
    \item \textbf{Grouping at the model level} (akin to ``$\lambda \propto \frac{1}{N}$'' in \citep{louizos2017learning}). corresponds to the case of  $G=1$ group comprising with 
    $d_{\text{in}} + d_{\text{hid}}$ gates. Finally, $\#(\vthetat_1) =     d_{\text{in}} \cdot d_{\text{hid}} + d_{\text{hid}} \cdot d_{\text{out}}$ gives the total number of parameters in the network.
\end{itemize} 

A similar analysis holds for the case of output feature-map sparsity used in convolutional layers.

\section{Normalizing the \texorpdfstring{$L_0$}{L0}-norm}
\label{app:normalized_l0}

\citet{louizos2017learning} normalize the expected $L_0$-norm of model parameters with respect to the \textit{training set size} $N$, and not with respect to the \textit{number of parameters}. This is done by selecting a $\lambdapen = \lambda/N$.  In contrast, as stated in \cref{eq:const_problem}, we favor normalizing by the total number of parameters in the model $\# (\vthetat)$. This yields an expected $L_0$-density consistently in the range $[0,\,1]$ regardless of model architecture. 

Optimization problems corresponding to each of these normalization schemes (grouping gates model-wise, for illustration) are given by:

\vspace{-2ex}

\hspace{1cm}
\begin{minipage}{0.45\textwidth}
\begin{align*}
    & \textbf{\text{Ours (Penalized)}} & \\
    \underset{\boldsymbol{\tilde{\theta}}, \vphi}{\text{min}}\quad & \fobj(\vthetat, \vphi) + \lambdapen \, \, \frac{\mathbb{E}_{\vz \ | \ \vphi} \left[ \|\vz\|_0 \right]}{\#(\vthetat)} 
\end{align*}
\end{minipage}
\begin{minipage}{0.45\textwidth}
\begin{align*}
    & \textbf{\citet{louizos2017learning}} & \\
    \underset{\boldsymbol{\tilde{\theta}}, \vphi}{\text{min}}\quad & \fobj(\vthetat, \vphi) + \lambdapen \, \, \frac{\mathbb{E}_{\vz \ | \ \vphi} \left[ \|\vz\|_0 \right]}{N} 
\end{align*}
\end{minipage}

Therefore, a $\lambdapen = \lambda$ in the context of our work does \emph{not} correspond to the same regularization factor as choosing $\lambdapen = \lambda/N$ in \citet{louizos2017learning}. For details on the number of parameters of each architecture and the associated number of training examples, see \cref{app:exp_details:model_stats}. Note that in certain settings (e.g. CIFAR-10/100) the number of training points and number of parameters of the model can differ by several orders of magnitude.

\section{Purging Models}
\label{app:purging_models}

In this section, we describe how we transform a model with stochastic gates $\vz$ and free parameters $\vthetat$ into a deterministic test-time model.  For conciseness, we present the procedure for convolutional layers. The case of fully connected layers is analogous: simply consider the  parameter groups to be ``all those weights connecting an input neuron to any neuron in the next layer''.

The procedure for removing filters from the $i$-th convolutional layer is as follows:
\begin{enumerate}
    \item For each filter in the layer, compute the test-time value of its associated gate as described in \cref{app:gates:validation_gates}.
    \item Gates with medians $\hat{z}_j=0$ are considered inactive. The value of an active gate $\hat{z}_j > 0$ is ``absorbed'' multiplicatively by its associated weights.
    \item Prune the filters associated with inactive gates and their corresponding activation maps. The kernel entries in the next convolutional layer associated with the pruned channels in layer $i$ are also removed. If present, weights of the adjacent batch
normalization layer are removed.
    \item A new kernel matrix is created for both the $i$-th and $(i + 1)$-th layer, and the remaining kernel weights are copied to the new model.
\end{enumerate}

The pruning procedure described above guarantees equivalent outputs for the network before and after pruning under the assumption that the element-wise activation function $h$ used in the network satisfies $h(0) = 0$, as is the case for ReLU activations. 

\textbf{Double sparsification.} We highlight that the pruning of filters of the subsequent layer in step 3 happens \textit{regardless of whether the following layer is sparsifiable} or not. This observation implies that if a model contains adjacent sparsifiable convolutional layers, the resulting number of active parameters in the second layer will be affected by its output sparsity rate, \textit{as well as the output sparsity rate of the first layer}.

This ``double sparsification'' leads to a subtle issue when studying the relationship between the proportion of active gates in the model and the \textit{effective} number of active parameters. For example, if $80\%$ of the (output) gates of layer $i$ are active, and $70\%$ of the (output) gates of layer $i+1$ are active, the effective number of active parameters in layer $i+1$ will be $\sim 56\%=(0.8 \cdot 0.7)$ and not $70\%$!

Note that this procedure is identical to that of \citet[$\S 3.1$]{li2017l1pruning}. However, our selection of filters to remove is based a sparsity pattern learned during training, rather than motivated by a heuristic ranking of the filter norms. We use a similar language and presentation to facilitate the comparison. 

The double sparsification effect arises due to performing structured pruning in adjacent layers, while considering groupings (i.e. one gate per output channel) which disregard the sparsity from the previous and next layers. We discuss this issue to provide clarity when analyzing our results regarding controllability: the goal of our constrained formulation is to achieve (at most) a certain proportion of expected active \textit{gates}. Our experiments demonstrate that we can indeed achieve said control (compare target density $\epsilon$ with $L_0$-density columns throughout all tables). Addressing the issue of ``double sparsification'' in structured pruning is beyond the scope of our work.

\textbf{Parameters and MACs.} To achieve fair comparisons between models trained using different techniques (e.g. penalized, constrained or magnitude pruning) we apply the same purging procedure to all models.  For all of the experiments presented in the paper, the reported parameter and MAC\footnote{A MAC operation modifies an accumulator $a$ as $a \leftarrow a + (b \times c)$.} counts are calculated based on the deterministic, purged models. The number of MACs corresponds to the number of multiply-accumulate operations involved in a \textit{forward} pass through the network.

In the case of magnitude pruning, steps 1 is replaced with the ranking-and-thresholding operation of \citet{li2017l1pruning}. This results (conceptually) in binary 0-1 values for the gates associated with each of the filters, required in step 2.

\section{Bisection Search}
\label{app:bisection}

\begin{figure}[ht]
\setlength\tabcolsep{7mm}%
\centering
\begin{tabular}{cc}
 \includegraphics[width=0.3\textwidth, trim={.25cm .25cm .25cm .25cm}, clip]{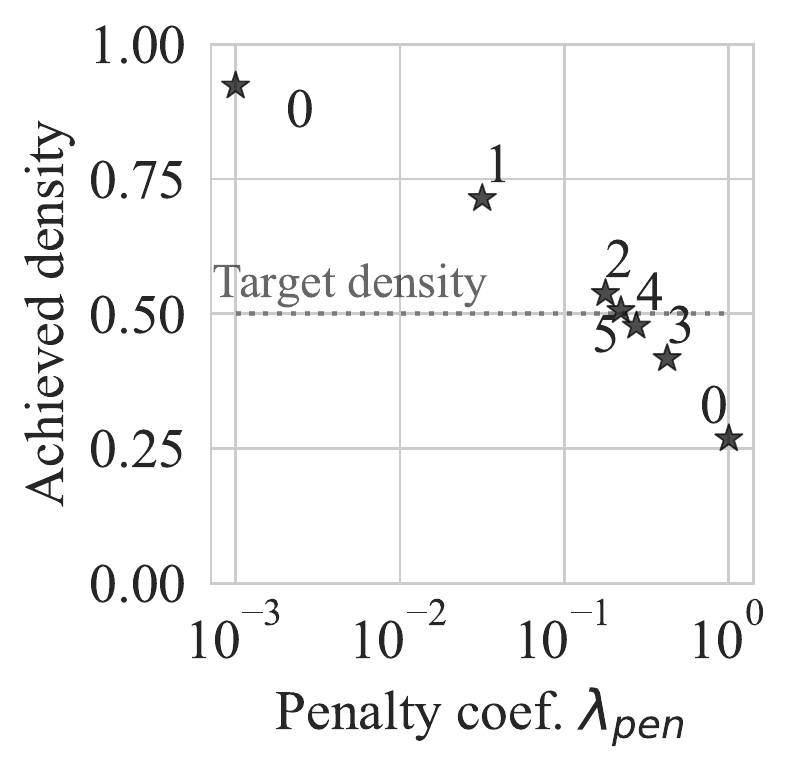} & 
  \includegraphics[width=0.3\textwidth, trim={.25cm .25cm .25cm .25cm}, clip]{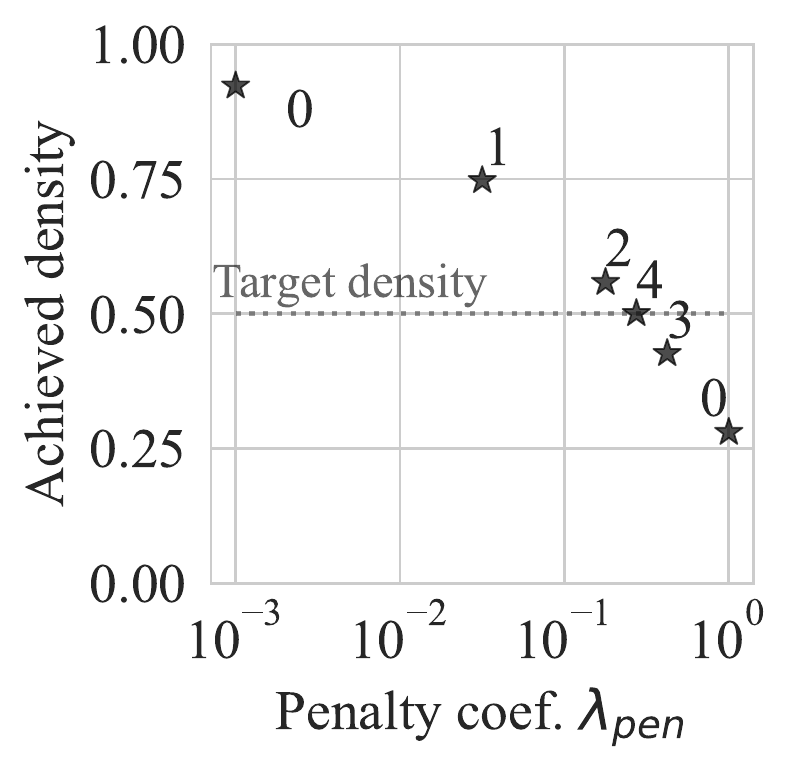}\\
 a) Layer-wise & (b) Model-wise \\
\end{tabular}
    \caption[]{Iterations of bisection search on the logarithmic $\lambdapen$ space for achieving a model density of $50\%(\pm1\%$). Annotations represent iteration indices, with endpoints labelled as 0. We report the $L_0$-density of MLPs after 150 training epochs. On the left, parameters are grouped layer-wise and groups share a fixed $\lambdapen$; on the right, parameters are grouped at the model level.}
    \label{fig:bisection}
\end{figure}

We execute a bisection search algorithm on the (logarithmic) value of $\lambdapen$, aiming to achieve a model $L_0$-density of $50\%(\pm1\%)$.
\cref{fig:bisection} shows the results for experiments with parameter groupings model- and layer-wise.

Although bisection search successfully finds a penalty value which achieves the desired density, it required the execution of at least 6 complete training cycles to be within $1\%$ of the target. Performing such a high number of repeated experiments for tuning $\lambdapen$ can be in-admissibly costly in real applications. Moreover, we chose the endpoint values such that their resulting densities enclosed the target, reducing the difficulty of the search problem. While bisection search is by no means the optimal approach to adjust $\lambdapen$, these experiments highlight the tunability challenges associated with penalized methods.

\section{Constrained Optimization: Theory and Further Algorithms}
\label{app:constrained_optimization}

Recall that a pure Nash equilibrium of the min-max game in \cref{eq:minmax_game} corresponds to a saddle point of the Lagrangian and determines an optimal, feasible solution \cite{neumann1928theorie}. However, for non-convex problems such pure Nash equilibria might not exist, and thus simultaneous gradient descent-ascent (GDA) updates can \textit{potentially} lead to oscillations in the parameters. \citep{cotter2019, lin2020gradient}.

There has been extensive research in the saddle-point optimization community studying in these type of problems. In particular, for convex-concave problems there are  (non-)asymptotic convergence guarantees for averaged iterates from GDA with equal step-sizes \citep{korpelevich1976extragradient, chen1997convergence, nemirovski2004prox}. \citet{lin2020gradient} present a comprehensive bibliography of studies focusing on nonconvex-concave problems, as is the case for the Lagrangian in \cref{eq:minmax_game}. The authors also present non-asymptotic complexity results showing that two-timescale GDA can find stationary points for nonconvex-concave minimax problems efficiently. 

\citet{cotter2019} propose an algorithm for non-convex constrained problems that returns an approximately optimal and feasible solution, consisting of a pair of mixed strategies (with support size of at most $m + 1$ for a problem with $m$ constraints). Experimentally we observed convergent, non-oscillatory behavior when using simultaneous updates for solving the problem in \cref{eq:minmax_game} employing \textit{pure strategies}, i.e., a single instance of primal and dual variables. This is discussed in detail in \cref{sec:experiments:no_oscillations} and \cref{app:tr_dynamics}.

Other approaches, such as extra-gradient \citep{korpelevich1976extragradient}, provide better convergence guarantees for games like \cref{eq:minmax_game}, compared to GDA. However, extra-gradient requires twice as many gradient computations per parameter update and the storage of an auxiliary copy of all trainable parameters. Nonetheless, extrapolation from the past \citep{gidel2018variational} enjoys similar convergence properties to extra-gradient without requiring a second gradient computation. Our preliminary experiments showed no significant difference in performance when using extra-gradient-based updates. These techniques can be useful for mitigating oscillatory behavior when applying Lagrangian-based optimization to other constrained problems.

\section{Training Dynamics and Dual Restarts}
\label{app:tr_dynamics}

\begin{figure*}[h]
    \centering
    \vspace{-1ex}
    \hspace{-.7cm}\includegraphics[scale=0.67]{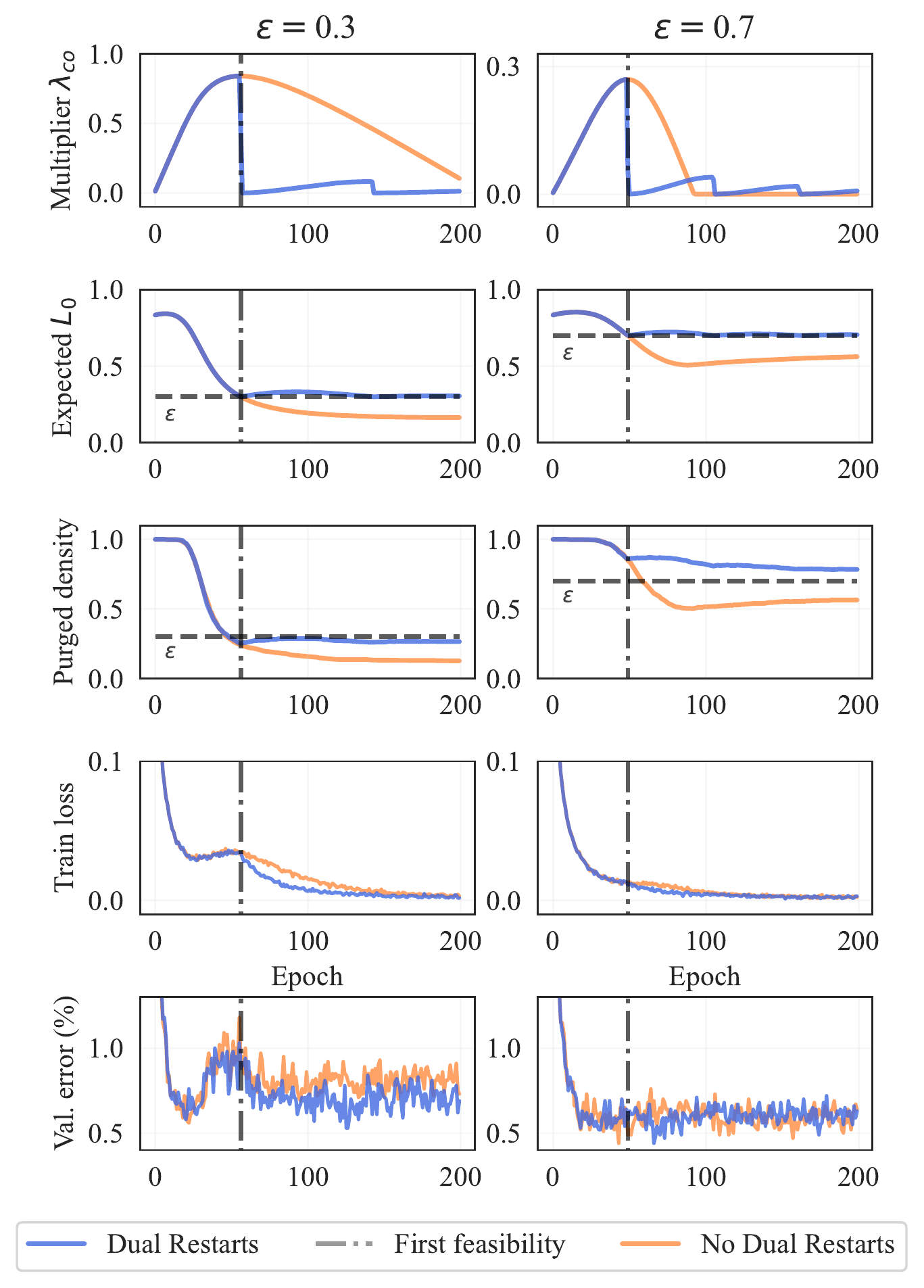}
    \vspace{-0.1in}
    \caption{
    Effect of the dual restarts scheme on the training dynamics for LeNet models on MNIST using model-wise constraints with target densities of $30\%$ and $70\%$.}
    \label{fig:dynamics}
\end{figure*}

In this section we provide further details on the training dynamics of our gradient descent-ascent approach for solving the Lagrangian in \cref{eq:minmax_game}. \cref{fig:dynamics} displays the training dynamics for a LeNet model on the MNIST dataset using model-wise density constraints of $30\%$ and $70\%$, as well as whether or not using dual restarts. The experimental setup for this section matches that of \cref{app:exp_details:mnist}.

We employ the same learning rate for both cases. Since the left column corresponds to a constraint yielding a more sparse model, the initial constraint violations are larger. In consequence, the magnitude of the Lagrange multiplier, which accumulates the constraint violations, is also larger for the $30\%$ case (compare the scale of the vertical axis in the plots of the second row). 

The horizontal dashed line in the first row signals the desired density for each of the cases. Note that all models become feasible: blue and orange lines are at or below the density target.  

However, \textit{not} employing dual restarts leads to the model being "excessively" sparsified: the orange line overshoots past the desired density level. While in principle this behavior is not ``wrong'' since the constraint is satisfied, it can lead to slow learning and decreased performance. Note that our constrained formulation leads to a natural monotonicity property in the constraints: if $\epsilon' < \epsilon$, the best performance achievable by a $\epsilon$-dense model is greater than or equal to the best performance achievable by an $\epsilon'$-dense model. 

When dual restarts are applied, the contribution of the accumulated constraint violation to the Lagrangian is removed once the constraints are satisfied. Thus, the optimization is mainly guided towards minimizing the training loss (see plots in third row). 

This ability to focus in reducing the loss usually come at the expense of increased density: note the slight "bounces" in model density. After reaching feasibility, models trained with dual restarts present small increases in density which are \textit{quickly mitigated} by further growth of the multiplier. As demonstrated throughout our experiments, we can reliably achieve models that are feasible or within ($\sim 1\%$) of the desired target level.

\subsection{Dual Restarts as Best Responses}
\label{app:dual_restart_best_response}

Our proposed ``dual restart'' scheme is theoretically motivated as a choice of best-response from the dual player when the constraints are satisfied. Without loss of generality, we present the argument below in the case of a single inequality constraint. When there are multiple inequality constraints, the best response problem for the dual player decouples into \textit{individual} problems for each of the Lagrange multipliers.

Given choices $[\theta,\, \phi]$ by the primal player, consider the optimization problem faced by the dual player:
\begin{equation}
    \lambda_{\text{co}}^{\text{BR}}(\tilde{\theta}, \phi)  = \underset{\lambda_{\text{co}} \geq 0}{\text{argmax}} \, \mathfrak{L}(\tilde{\theta}, \phi, \lambda_{\text{co}})  = \underset{\lambda_{\text{co}} \geq 0}{\text{argmax}} \quad \fobj(\vthetat, \vphi) + \lambda_{\text{co}} \left( \gconst(\phi) - \epsilon \right)
\end{equation}
This is a linear optimization problem with a trivial solution: if the constraint is being satisfied $(\gconst(\phi) - \epsilon < 0)$, then $\lambda_{\text{co}}^{\text{BR}} = 0$. If the constraint is satisfied with equality, $\lambda_{\text{co}}^{\text{BR}} = \mathbb{R}^+$. Finally, if the constraint is violated $(\gconst(\phi) - \epsilon > 0)$, then $\lambda_{\text{co}}^{\text{BR}} = \infty$.
 
In summary, applying dual restarts corresponds to updating the value of the Lagrange multiplier following a best response for the dual player, regardless of the current value of the Lagrange multiplier! However, note that the same reasoning cannot be applied to the case of violated constraints: stability and overflow issues render a choice of $\infty$ for a Lagrange multiplier to be impractical for a numerical implementation.

Finally, we emphasize that while gradient ascent is an effective and simple tool for updating the Lagrange multipliers, further empirical and theoretical investigation on the influence of the dual update scheme could be beneficial.

\section{Learning Sparsity-Controlled (Wide)ResNets}
\label{app:resnet_control_fixes}

ResNets have been a challenging setting for $L_0$-penalty based methods. \citet{gale2019state} trained WideResNets (WRNs) \citep{zagoruykoK16} and ResNet50 \citep{he2015resnets} using the penalized $L_0$-regularization framework of \citet{louizos2017learning}, and reported being unable to produce sparse residual models without significantly compromising performance.\footnote{\citet{gale2019state} state: \textit{``Across hundreds of experiments, our [ResNet50] models were either able to achieve full test set performance with no sparsification, or sparsification with test set performance akin to random guessing.''}; and \textit{``Applying our weight-level $L_0$ regularization implementation to WRN produces a model with comparable training time sparsity, but with no sparsity in the test-time parameters. For models that achieve test-time sparsity, we observe significant accuracy degradation on CIFAR-10.''}}
Our initial experiments on WRNs confirmed a similar behavior. 

\begin{figure}[h]
\centering

\begin{minipage}{.47\textwidth}
    \centering
    \includegraphics[width=0.8\textwidth,trim={.25cm .3cm .25cm .8cm}, clip]{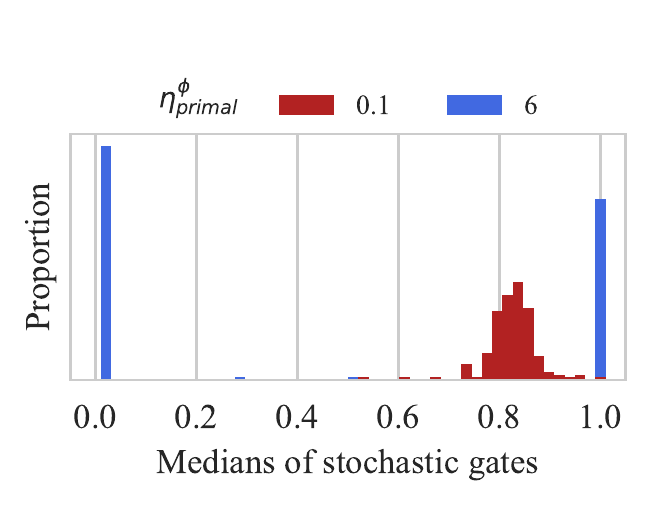}
    \captionof{figure}{Distribution of gate medians for the first layer of a WRN, at the end of (penalized) training using $\lambdapen=10^{-3}$.}
    \label{fig:increase_lr}
\end{minipage}%
\hfill
\begin{minipage}{.47\textwidth}
    \centering
    \includegraphics[width=0.85\textwidth,trim={.25cm .3cm .25cm .2cm}, clip]{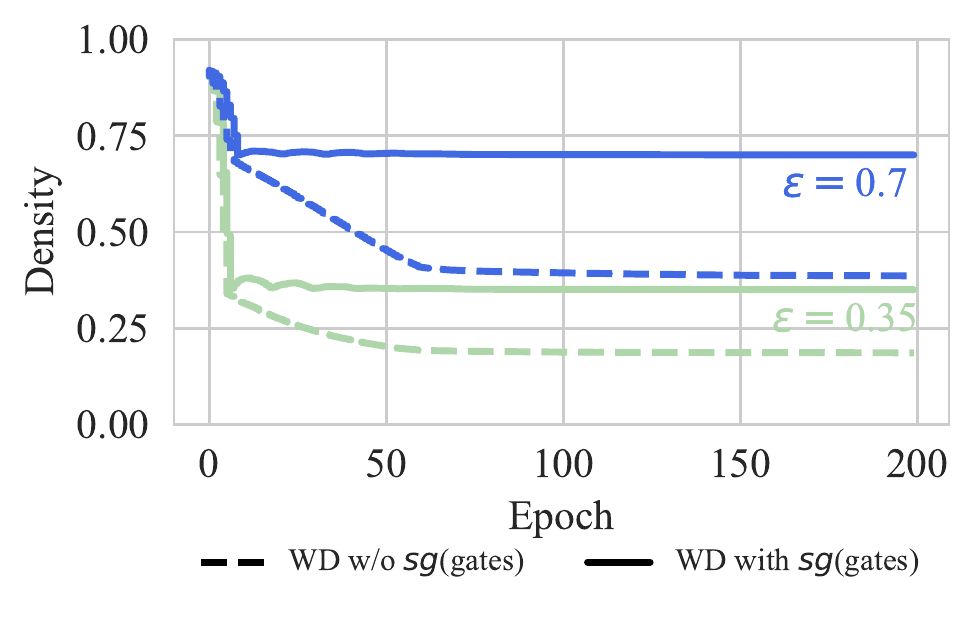}
    \captionof{figure}{Not detaching the gradient contribution of the weight decay (WD) penalty leads to excesive sparsification.}
    \label{fig:wd_overpen}
\end{minipage}

\end{figure}

We detect two main modifications that enable us to learn WRNs and ResNets with controllable sparsity, while retaining good performance: \blobletter{1}  increasing the learning rate of the stochastic gates, shown in \cref{fig:increase_lr}; and \blobletter{2} removing the gradient contribution of the weight decay penalty towards the gates, displayed in \cref{fig:wd_overpen}.

For conciseness, we present these observations in the case of WideResNets. We adopt the two adjustments presented in this section for our experiments involving WideResNet-28-10, ResNet18 and ResNet50 models. These two simple modifications allowed us to achieve reliable controllability for (Wide)ResNets without performance degradation, just as with the MLP and LeNet architectures. 

Further exploration of these modifications, and their influence in the resulting sparsity of the model are provided in \cref{app:test_time_gates}.

\subsection{Sparsity in (Wide)ResNets by Tuning the Learning Rate of the Gates.} 
\label{app:resnet_control_fixes:adjust_lr}

Replicating the CIFAR10 experiments of \citet{louizos2017learning} on WRNs using their choice of regularization parameter and learning rate for the stochastic gates, results in a distribution of the stochastic gates which does not induce sparsity in the model.
\cref{fig:increase_lr} shows (in red) the distribution of the gate medians in the first layer of a WRN trained using a learning rate of $\eta_{\text{primal}}^{\vphi}=0.1$ for the gates parameters, as in \citet{louizos2017learning}. We observed that this distribution of medians did not change significantly during training. Thus, since the model has a high initial density, the gate parameters at the end of training do not induce any sparsity.

To enable the gate parameters to effectively change during training, we decoupled the learning rate of the gates $\gatelr$, from that of the model weights $\weightlr$. \cref{fig:increase_lr} illustrates how increasing $\gatelr$ from $0.1$ to $6$ leads to a drastically different distribution for the gate medians. This simple change results in a distribution of medians which exhibits the desired concentration behavior: a non-negligible proportion of gates have a median of zero, and are therefore inactive (see \cref{app:gates:validation_gates}).

We adopt this learning rate adjustment in all our WRN experiments. \cref{tab:cifar10} presents the performance of WRNs trained using different values of $\gatelr$, for both the constrained and penalized settings. Note how the models using a higher learning rate for the gates parameters successfully achieve sparsity \textit{without any compromise in performance}.

\subsection{A Loophole in Weight Decay Leads to Excessive Regularization.}  
\label{app:resnet_control_fixes:detach_gates}

This section includes the adjustment to the learning rate of the gates presented above. We noticed a systematic over-sparsification behavior in WRNs when solving the constrained formulation: the constraint is satisfied, well beyond the prescribed density level. 

This issue is illustrated in \cref{fig:wd_overpen}. Dashed lines correspond to the weight decay from \citet{louizos2017learning} and solid lines correspond to our method with the modified weight decay as in \cref{eq:detach_gate}. Experimental details for CIFAR-10 experiments are provided in \cref{app:exp_details:cifar10}.

We identified the cause of this phenomenon to be the $L_2$ weight decay term in the training objective of WRNs from \citet{louizos2017learning} (see \cref{sec:l0_paper}). This penalty term depends both on the probability of gates being active $\pi_j = \mathbb{P}[ z_j \neq 0 ]$, as well as the norm of the signed magnitudes $\tilde{\theta}^2_j$. Reducing the value of this penalty could be achieved by turning off gates, such that the contribution of their associated magnitudes is ignored. This behavior is undesirable for the purpose of controllability. 

We propose to restrict the effect of the weight decay penalty to $\vthetat$, as a way to reduce the parameter magnitudes, and keep the gates deactivation under the \emph{sole influence} of the constraint violation term.
We achieve this by stopping (also known as detaching) the gradients from propagating through the gate-dependent terms in the $L_2$ norm:
\vspace{-2ex}
\begin{equation}
    \mathbb{E}_{\vz|\vphi}\left[ \|\vthetah\|^2_2 \right] = \sum_{j=1}^{|\vtheta|} \texttt{stop-grad}(\pi_j) \,  \tilde{\theta}^2_j.
    \label{eq:detach_gate}
\end{equation}

\cref{fig:wd_overpen} shows the effect of this simple adjustment when applying a model-wise constraint on a WRN for the CIFAR10 dataset. Note how, removing the influence of weight decay on the gate parameters allows us to reliably achieve the desired target density, without over-sparsifying the model.

\section{Test-Time Gates and Influence of Learning Rate and Weight Decay}
\label{app:test_time_gates}

\begin{figure}[h]
    \centering
    \includegraphics[width=0.95\textwidth]{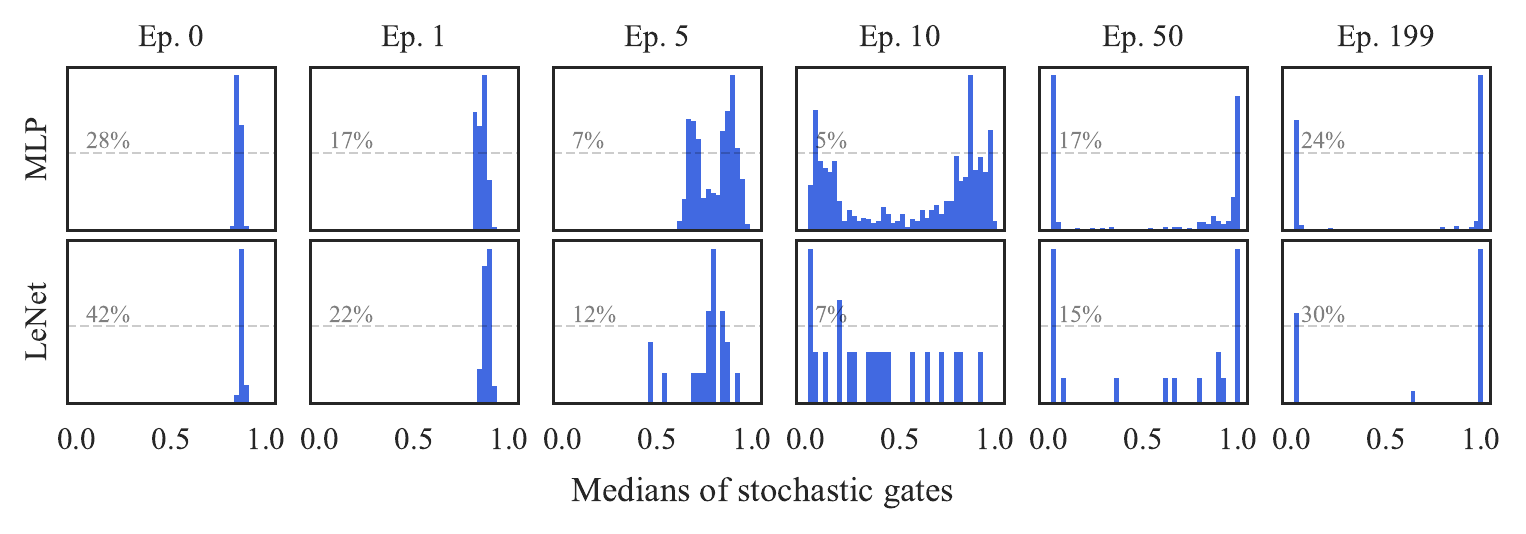}
    \caption{
        Histograms of \textit{medians} of the gates in first layer of MLP and LeNet models. These models were trained on MNIST with a layer-wise target density of $70\%$. Note the transition from highly concentrated around a fractional value of $~0.8$ at the first epochs, towards an approximately binary distribution peaking at 0 and 1 at the last epoch.}
        \label{fig:mnist_gate_histograms}
\end{figure}

Recall that we make the gates ``freeze'' the gates at their medians to obtain a deterministic model to evaluate on unseen data (\cref{app:gates:validation_gates}). We analyze the behavior of gates at test-time by considering histograms of their medians across specific layers of models.\footnote{Note the subtle detail that these histograms are based on a statistic (the median) of a probability distribution and do not represent distributions of gates by themselves.} This section provides further empirical evidence to support the hypothesis presented in \cref{app:resnet_control_fixes} (see \cref{app:exp_details:cifar10} for experiment setup).

\cref{fig:mnist_gate_histograms} contains histograms for the first layer of an MLP and a LeNet trained to classify MNIST digits. 
These correspond to a fully connected and a convolutional layer, respectively. 
Sparsity requirements are specified via layer-wise constraints with a target density of $70\%$ on both cases. 
Experimental details for these runs match those presented in \cref{app:exp_details}. 
Test-time gate medians are measured at the \textit{end} of the training epochs shown in the panel titles.

At initialization, the distributions of different gates are highly similar and yield closely packed medians. 
These are mostly fractional in value, away from being 0 or 1. 
As training progresses, the medians drift apart. 
Histograms peak at 0 and 1 as of the 50th epoch.
As expected, approximately $30\% = (100 - 70)\%$ of gates are inactive at the end of training. 

\begin{figure}[h]
    \vspace{-8ex}
    \centering
\includegraphics[width=\textwidth]{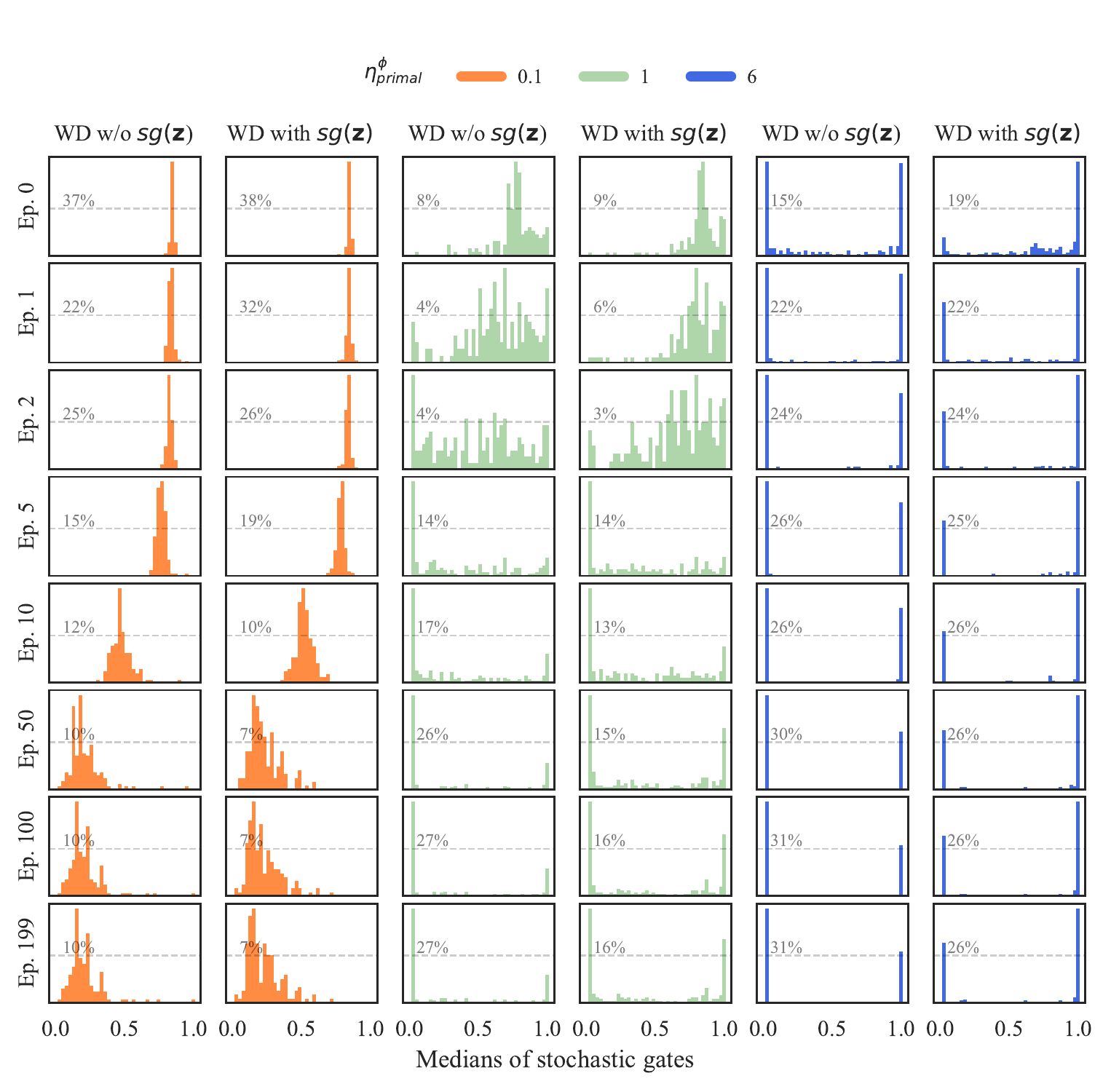}
    \caption{
        Histograms of gate medians for the first convolutional layer of WRN models trained on CIFAR10 under $70\%$ layer-wise density constraints. Configurations comprise three learning rates $\gatelr$, and removing or not the gradient contribution of the weight decay towards the gate updates. 
        \label{fig:wrn_gate_histograms}
    }
\end{figure}

\cref{fig:wrn_gate_histograms} contains histograms of the test-time gates associated with the first convolutional layer of various WRNs-28-10 trained on CIFAR10. The proportion of gates at zero specifies the amount of inactive gates, while bars in $(0,1]$ correspond to active gates. Since all histograms have different scales, we provide a reference line for each, corresponding to \textit{half} the height of the tallest bar in said histogram. For example, if the dashed line is labeled as $15\%$, then the highest bar in that histogram corresponds to $30\%$.

We consider 6 configurations spanning: three learning rates for the gate parameters $\vphi$; and whether or not to remove the gradient contribution of the weight decay towards the gates (see \cref{app:resnet_control_fixes}).
All experiments use the same initialized model. Sparsity requirements are specified via layer-wise constraints with a target density of $70\%$ on all cases. Measurements of the gate medians are made at the \textit{end} of each of the presented epochs.

Medians do not drift apart noticeably when employing $\gatelr = 0.1$. 
In addition, they maintain fractional values (i.e., remain away from 0 and 1). 
Given our protocol for choosing gates at test-time in \cref{app:gates:validation_gates}, this would lead to a ``fully dense'' test-time network, which violates the required sparsity constraints by a wide margin. 
Similar to \citet{gale2019state}, we observed that when a WRN trained with $\gatelr = 0.1$ achieved \textit{any} significant degree of sparsity, it led to a performance akin to random guessing.
Note that, unsurprisingly, the bulk of the medians gets closer to zero: this is a consequence of the model aiming to satisfy the constraint on the expectation of the $L_0$-norm.

The behavior when training with $\gatelr = 1$ and $\gatelr = 6$ stands in clear contrast to that of $\gatelr = 0.1$. The former two resemble more closely the dynamics observed in \cref{fig:mnist_gate_histograms}, where medians disperse and tend to accumulate and saturate at 0 or 1. Unsurprisingly, this accumulation happens more quickly for experiments with $\gatelr = 6$. This setting also has the smallest proportion of fractional medians at the end of training.  Note that in the case of $\gatelr = 1$, longer training could result in a similar outlook to that of $\gatelr = 6$. Choosing $\gatelr > 1$ in our experiments led to better performance in terms of test-time error. 

Finally, note how\footnote{Except for the undesirable setting $\gatelr = 0.1$, as discussed in \cref{app:resnet_control_fixes}.} removing the gradient contribution of the weight decay towards the gates (WD \textbf{with} $\texttt{sg}(z)$) yields yields test-time models which have approximately $30\%$ of their parameters \textit{inactive}, as desired. Not removing this contribution (WD \textbf{without} $\texttt{sg}(z)$) results in over-sparsification: the distribution of gate medians is shifted towards zero for all learning rate choices. For example, consider the panels at the last epoch for $\gatelr = 1$ and $\gatelr = 6$. The experiments \textit{with} detaching result in models whose sparsity is close to the desired sparsity level (peak at 0 close to $30\%$). In contrast, when not detaching, the rate of \textit{inactive} gates is around $60\%$, twice as sparse as required.

\section{Experimental Details}
\label{app:exp_details}

Our implementation is developed in Python 3.8, using Pytorch 1.11 \citep{pytorch} and the Cooper constrained optimization library \citep{cooper}. We provide scripts to replicate the experiments in this paper at: \\
\texttt{\url{https://github.com/gallego-posada/constrained_sparsity}}.

All our models use ReLU activations. Throughout our experiments \blobletter{1} we decouple the learning rates used for the weights and gates of the network, and \blobletter{2} when employing weight decay, we remove the gradient contribution of the penalty towards the gates, as explained in \cref{app:resnet_control_fixes}. 

\subsection{Model Statistics}
\label{app:exp_details:model_stats}

\cref{tab:model_stats} describes the different architectures used throughout our experiments in terms of their total number of trainable parameters and the computational cost involved in a \textit{forward} calculation in MACs (multiply-accumulate operations). We also provide details on the input size and the number of training examples for their respective datasets.

\begin{table}[h]
\renewcommand{\arraystretch}{1.1}
\centering
\caption{Count of parameters and MACs for all architectures used in this paper, along with dimension and number of training examples.}
\label{tab:model_stats}
    \vspace{1ex}
    \begin{tabular}{cccccc}
        \textbf{Model Type} & \textbf{Parameters} & \textbf{MACs} & \textbf{Input size} & \textbf{Train set size} &  \textbf{Dataset}\\ 
        \hline
        \hline
        MLP & 266k & 267k & (28, 28) & 50k & MNIST\\
        LeNet & 431k & 2,327M & (28, 28) & 50k & MNIST\\
        WideResNet-28-10 & 36.5M & 5,959M & (3, 32, 32) & 50k & CIFAR-10/100\\
        ResNet18 & 11.3M & 6,825M & (3, 64, 64) & 100k & TinyImageNet\\
        ResNet50 & 25.5M & 4,120M & (3, 224, 224) & 1.2M & ImageNet
    \end{tabular}
\vspace{-2.5ex}
\end{table}

\subsection{Dual Optimizer}
\label{app:exp_details:dual_optimizer}

Note that the constraint functions considered throughout this work involve expectations but can be computed in closed-form based on the parameters of the gates (\cref{app:gates}). Therefore, the computation of constraint violations is deterministic. We employ gradient \textit{ascent} on the Lagrange multipliers. We initialize all Lagrange multipliers at zero. Details on the chosen dual learning rate, along with the use of dual restarts are provided for each experiment below.

\subsection{MNIST}
\label{app:exp_details:mnist}

Following \citet{louizos2017learning}, our experiments on MNIST classification consider two different architectures: i) an MLP with 2 hidden layers with 300 and 100 units respectively, and ii) a LeNet-5 network, consisting on convolutional layers of 20 and 50 output channels, each succeeded by a max-pooling layer with stride 2; followed by two fully connected layers of 800 and 500 input dimensions. All fully connected layers in these models use input neuron sparsity, and all convolutional layers (for LeNet models) employ output feature map sparsity.

\begin{table}[h]
    \vspace{-2.5ex}
    \renewcommand{\arraystretch}{1.1}
    \centering
    \caption{Default configurations for MLP and LeNet5 experiments on MNIST.}
    \label{tab:optim_mnist}
    \vspace{1ex}
    \resizebox{0.9\textwidth}{!}{%
    \begin{tabular}{cccccccc}
    \hline
      \multirow{2}{*}{\textbf{Approach}} & \multicolumn{2}{c}{\textbf{Weights}} & \multicolumn{2}{c}{\textbf{Gates}} & \multicolumn{3}{c}{\textbf{Lagrange Multipliers}} \\
      \cline{2-8} 
      & Optim. & $\weightlr$ & Optim. & $\gatelr$ & Optim. & $\duallr$ & Restarts \\ 
      \hline
      \hline
      Constrained & \multirow{2}{*}{Adam} & \multirow{2}{*}{$7 \cdot 10^{-4}$} & \multirow{2}{*}{Adam} & \multirow{2}{*}{$7 \cdot 10^{-4}$} & Grad. Ascent & $10^{-3}$ & Yes \\ 
      Penalized &  & &  &  & - & - & - \\ 
       \hline
      
    \end{tabular}
    }
\end{table}

\cref{tab:optim_mnist} presents the hyper-parameters used for learning sparse MLPs and LeNets. Both cases employ the same configuration for the primal optimizer: Adam \citep{adam} with  $(\beta_1,\ \beta_2) = (0.9,\ 0.99)$, as provided by default in Pytorch, with a batch size of 128. These experiments do not use weight decay.

\subsection{CIFAR-10 and CIFAR-100}
\label{app:exp_details:cifar10}

We employ WideResNet-28-10 (WRN) models \citep{zagoruykoK16} for the tasks of classifying CIFAR-10 and CIFAR-100 images. 
Akin to \citet{louizos2017learning}, the first convolutional layer in each residual block uses output feature map sparsity, whereas the following convolutional layer and the residual connection are kept to be fully dense. This model counts with 12 sparsifiable convolutional layers.

\cref{tab:optim_wrn} presents the hyper-parameters used for learning sparse WRNs. We use SGD with a momentum coefficient of 0.9 for the weights and gates. We use a batch size of 128 for 200 epochs. The primal learning rate is multiplied by 0.2 at 60, 120 and 160 epochs. This mimics the training procedure of \citet{zagoruykoK16} and \citet{louizos2017learning}. These experiments use $\rho_{\text{init}} = 0.3$ (see  \cref{app:rho_init}).

\begin{table}[h]
    \vspace{-2.5ex}
    \renewcommand{\arraystretch}{1.1}
    \centering
    \caption{Default configurations for WideResNet-28-10 experiments on CIFAR-10 and CIFAR-100.}
    \vspace{1ex}
    \label{tab:optim_wrn}
    \resizebox{\textwidth}{!}{%
    \begin{tabular}{ccccccccc}
    \hline
      \multirow{2}{*}{\textbf{Approach}} & \multicolumn{2}{c}{\textbf{Weights}} & \multicolumn{2}{c}{\textbf{Gates}}   & \multicolumn{3}{c}{\textbf{Lagrange Multipliers}} & \textbf{Weight decay} \\
      \cline{2-8} 
       & Optim. & $\weightlr$ & Optim. & $\gatelr$ & Optim. & $\duallr$ & Restarts & \textbf{coefficient} \\ 
      \hline
      \hline
       Constrained & \multirow{2}{*}{SGDM} & \multirow{2}{*}{\textbf{$0.1$}} & \multirow{2}{*}{SGDM} & \multirow{2}{*}{\textbf{$6$}} & Grad. Ascent & $7 \cdot 10^{-4}$ & Yes & \multirow{2}{*}{\textbf{$5 \cdot 10^{-4}$}} \\ 
       Penalized & & & & & - & - & - & \\ 
       \hline
    \end{tabular}
    }
\vspace{-2.5ex}
\end{table}

\subsection{TinyImageNet}
\label{app:exp_details:tiny_imagenet}

We employ ResNet18 models for the task of classifying TinyImageNet \citep{tinyImagenet} images. 
The model's initial convolutional and final fully connected layers are kept fully dense.
The residual connection of each \texttt{BasicBlock} in the model is kept fully dense, while all other convolutional layers employ output feature-map sparsity. This model thus counts with 16 sparsifiable convolutional layers. 

\cref{tab:optim_r18} presents the hyper-parameters used for learning sparse ResNet18s. We use SGD with a momentum coefficient of 0.9 for the weights and gates. We use a batch size of 100 for 120 epochs. The learning rate of the weights $\weightlr$ is multiplied by 0.1 at 30, 60 and 90 epochs. This mimics the training procedure of previous works \citep{kundu2020pre}. This experiment uses $\rho_{\text{init}} = 0.3$ (see  \cref{app:rho_init}).

\begin{table}[h]
    \vspace{-2.5ex}
    \renewcommand{\arraystretch}{1.1}
    \centering
    \caption{Default configurations for ResNet18 experiments on TinyImageNet.}
    \vspace{1ex}
    \label{tab:optim_r18}
    \resizebox{\textwidth}{!}{%
    \begin{tabular}{cccccccccc}
      \hline
     \multirow{2}{*}{\textbf{Approach}} & \multirow{2}{*}{\textbf{Grouping}} & \multicolumn{2}{c}{\textbf{Weights}} & \multicolumn{2}{c}{\textbf{Gates}} & \multicolumn{3}{c}{\textbf{Lagrange Multipliers}} & \textbf{Weight decay} \\
    \cline{3-9} 
      & & Optim. & $\weightlr$ & Optim. & $\gatelr$ & Optim. & $\duallr$ & Restarts & \textbf{coefficient} \\ 
      \hline
      \hline
      \multirow{2}{*}{Constrained} & Model & \multirow{4}{*}{SGDM} & \multirow{4}{*}{\textbf{$0.1$}} & \multirow{4}{*}{SGDM} & \multirow{4}{*}{\textbf{$1$}} & \multirow{2}{*}{Grad. Ascent} & $8 \cdot 10^{-4}$ & \multirow{2}{*}{Yes} & \multirow{4}{*}{\textbf{$5 \cdot 10^{-4}$}} \\ 
      & Layer & & & & & & $1 \cdot 10^{-4}$ & & \\
      \cline{7-9}
      \multirow{2}{*}{Penalized} & Model & & & & & - & - & - & \\
       & Layer & & & & & - & - & - & \\
       \hline
    \end{tabular}
    }
\vspace{-2.5ex}
\end{table}

\subsection{ImageNet}
\label{app:exp_details:imagenet}

We employ ResNet50 models for the task of classifying ImageNet \citep{imagenet} images. 
The model's initial convolutional and final fully connected layers are kept fully dense.
The residual connection of each \texttt{Bottleneck} block in the model is kept fully dense, while all other convolutional layers employ output feature-map sparsity. This model thus counts with 48 sparsifiable convolutional layers. 

Due to the high computational cost of ImageNet experiments, and the tunability issues of the penalized method, we do not perform penalized experiments for this dataset.

\begin{table}[h]
    \vspace{-2.5ex}
    \renewcommand{\arraystretch}{1.1}
    \centering
    \caption{Default configurations for ResNet50 experiments on ImageNet.}
    \vspace{1ex}
    \label{tab:optim_r50}
    \resizebox{\textwidth}{!}{%
    \begin{tabular}{cccccccccc}
      \hline
     \multirow{2}{*}{\textbf{Approach}} & \multirow{2}{*}{\textbf{Grouping}} & \multicolumn{2}{c}{\textbf{Weights}} & \multicolumn{2}{c}{\textbf{Gates}} & \multicolumn{3}{c}{\textbf{Lagrange Multipliers}} & \textbf{Weight decay} \\
    \cline{3-9} 
      & & Optim. & $\weightlr$ & Optim. & $\gatelr$ & Optim. & $\duallr$ & Restarts & \textbf{coefficient} \\ 
      \hline
      \hline
      \multirow{2}{*}{Constrained} & Model & \multirow{2}{*}{SGDM} & \multirow{2}{*}{\textbf{$0.1$}} & \multirow{2}{*}{SGDM} & \multirow{2}{*}{\textbf{$1$}} & \multirow{2}{*}{Grad. Ascent} & $3 \cdot 10^{-4}$ & \multirow{2}{*}{Yes} & \multirow{2}{*}{\textbf{$10^{-4}$}} \\ 
      & Layer & & & & & & $3 \cdot 10^{-5}$ & & \\
       \hline
    \end{tabular}
    }
    \vspace{-1ex}
\end{table}

\cref{tab:optim_r50} presents the hyper-parameters used for learning sparse ResNet50s. SGD with a momentum coefficient of 0.9 is used for the weights and the gates. We use a batch size of 256 for 90 epochs. The learning rate of the weights $\weightlr$ is multiplied by 0.1 at 30 and 60 epochs. This mimics the training procedure of previous works \citep{savarese2020winning}.

\textbf{Initialization of the gates}

As discussed in \cref{app:rho_init}, the choice of hyperparameter $\rho_{\text{init}}$ affects the initial sparsity of the network. For their experiments using WideResNet models on the CIFAR10/100 datasets, \citet{louizos2017learning} chose $\rho_{\text{init}} = 0.3$. When executing our experiments with ResNet50 models on ImageNet, we noticed that this hyper-parameter of the $L_0$-reparametrization can have a significant impact in the behavior of the model throughout training. 

\begin{figure}[h]
    \setlength\tabcolsep{0pt}%
    \centering
    \includegraphics[width=0.8\textwidth, trim={2mm 0 2mm 2mm},clip]{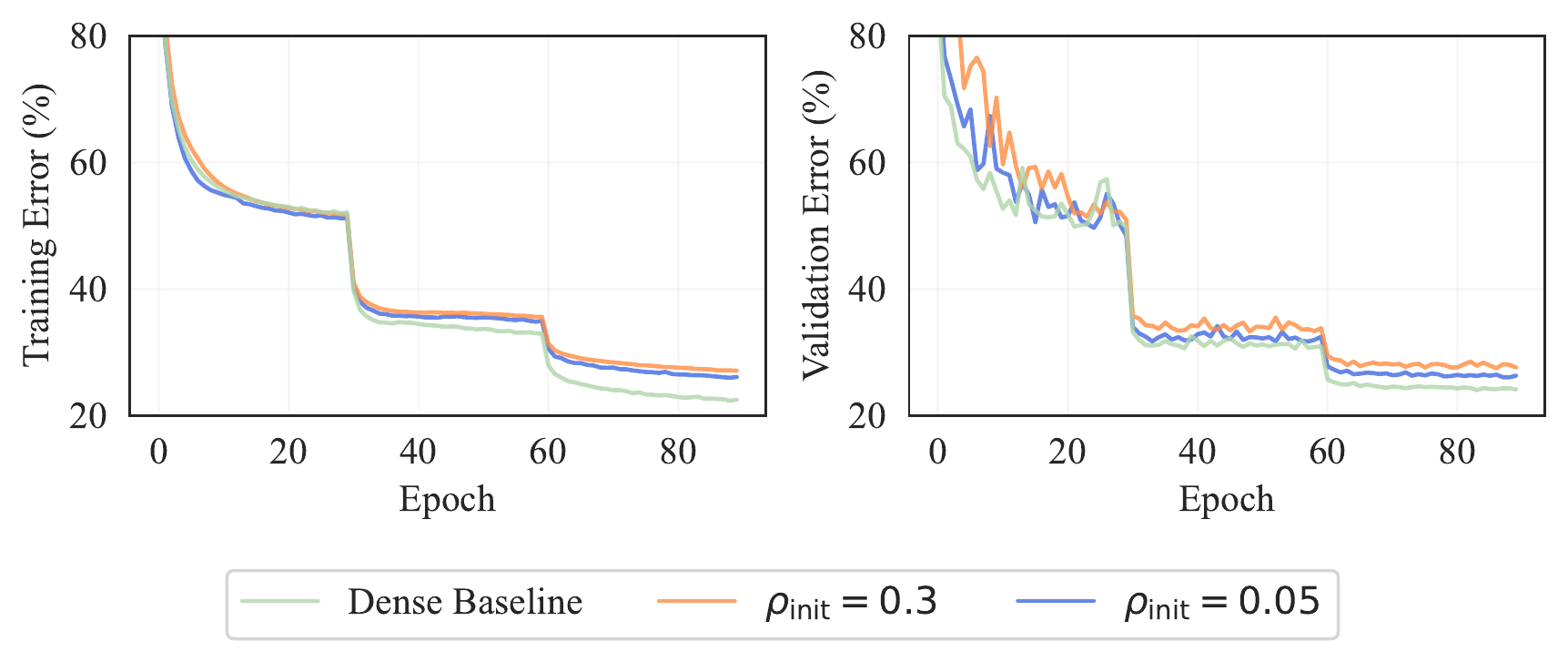}
    \vspace{-1ex}
    \caption[]{Effect of the initialization hyper-parameter $\rho_{\text{init}}$ for training ResNet50 models on ImageNet.}
    \label{fig:droprate}
\end{figure}

\cref{fig:droprate} shows the training and validation error of two different runs at a target density of 70\% with $\rho_{\text{init}} = 0.3$ and $\rho_{\text{init}} = 0.05$. Although both runs achieve the desired sparsity target,  the model initialized at $\rho_{\text{init}} = 0.05$ outperforms the model initialized with $\rho_{\text{init}} = 0.3$. This performance improvement persists throughout training, resulting in a final model with a better validation error.

Recall that we consider fixed values for the parameters $\beta$, $\gamma$ and $\zeta$ associated with the concrete distribution, as detailed in \cref{tab:gates_params}. With these values, \cref{eq:rho_init_prob} yields that the initial $L_0$-density of the model initialized with $\rho_{\text{init}} = 0.3$ is $92.03\%$, while for $\rho_{\text{init}} = 0.05$, the $L_0$-density of the initial model is $98.95\%$. This means that the case $\rho_{\text{init}} = 0.3$ starts from a (in expectation) $\sim7\%$ sparser model; thus unnecessarily restring the model capacity at the beginning of training. This is consistent with the behavior displayed in \cref{fig:droprate}: the model initialized with $\rho_{\text{init}} = 0.05$ follows more closely the validation performance of a dense baseline (i.e. a standard ResNet50 without gates).

Considering this behavior, for all the ImageNet results reported below we use the lower value of $\rho_{\text{init}} = 0.05$. For other models and datasets, we employ the value used in \citet{louizos2017learning} for ease of comparison.

\subsection{Magnitude Pruning Comparison on ImageNet}

We compare the performance of our in-training sparsity method with structured magnitude pruning. We start from a pre-trained ResNet50 model from Pytorch (\texttt{torchvision.models.resnet50}) and apply layer-wise pruning following the procedure of \citet{li2017l1pruning} to \textit{the same layers} that were sparsifiable for our ImageNet models, described in \cref{app:exp_details:imagenet}. After performing magnitude pruning, we fine-tune the models for 20 epochs on the ImageNet dataset, using SGD with momentum of 0.9 and a constant learning rate of 0.001. This matches the fine-tuning setting of \citet{li2017l1pruning}.

\subsection{Sparsity Collapse}

Some of the results for the penalized method presented in \cref{app:all_comparisons} are labeled as ``Failed due to sparsity collapse''. This means that the penalty factor was too high and resulted in all the gates of a layer being turned off.

\section{Comprehensive Experimental Results}
\label{app:all_comparisons}

In this section we provide complete results for all experiments, whose hyper-parameter configurations can be found in \cref{app:exp_details}. To make the navigation of these results easier, we repeat some of the tables and figures provided in the main paper. These repetitions are clearly marked in the caption of the corresponding resource. 

\newpage

\subsection{MNIST}
\label{app:all_comparisons:mnist}

\begin{table*}[h!]
    \renewcommand{\arraystretch}{1.1}
    \centering

    \caption{Achieved density levels and performance for sparse MLP and LeNet5 models     trained on MNIST for 200 epochs. Metrics aggregated over 5 runs. { $^\dag$Results by \citet{louizos2017learning} with $N$ representing the training set size (see \cref{app:normalized_l0}). \textit{This table is the same as \cref{tab:mnist}. We repeat it here for the reader's convenience.}}}
    \label{tab:mnist_APPENDIX}
    \vspace{1ex}
    
    \resizebox{\textwidth}{!}{%
    \begin{tabular}{ccclccc}
      \hline
      \multirow{3}{*}{\textbf{Architecture}} & \multirow{3}{*}{\textbf{Grouping}} & \multirow{3}{*}{\textbf{Method}}  & \multirow{3}{*}{\textbf{\hspace{5mm}Hyper-parameters}} & \textbf{Pruned} & \multicolumn{2}{c}{\textbf{Val. Error} (\%)}  \\
      \cline{6-7} & & &  &  \multirow{2}{*}{\textbf{architecture}} & \multirow{2}{*}{best} & at 200 epochs \\
       & & & & &  & avg {\color{gray} $\pm$ 95\% CI}\\
      \hline
      \hline
       & \multirow{2}{*}{Model} & Pen.  &  $^\dag\lambda_{pen} = 0.1/N$ &  219-214-100 & 1.4 & --\\
       MLP & & Const. &  $\epsilon = 33 \%$ &  198-233-100 & 1.36 & 1.77 {\color{gray} $\pm$ 0.08}\\
       \cline{2-7}
       784-300-100 & \multirow{2}{*}{Layer} & Pen. &  $^\dag\lambda_{pen} = [0.1,0.1,0.1]/N$ &  266-88-33 & 1.8 & -- \\
      & & Const. &  $\epsilon = [30 \%, 30\%, 30\%]$ &  243-89-29 & 1.58 & 2.19 {\color{gray} $\pm$ 0.12 }\\
       \hline
       & \multirow{2}{*}{Model} & Pen. &   $^\dag\lambda_{pen} = 0.1/N$ &  20-25-45-462 & 0.9 & --\\
       LeNet5 & & Const. &  $\epsilon = 10\%$ &  20-21-34-407 & 0.56 & 1.01 {\color{gray} $\pm$ 0.05}\\
       \cline{2-7}
       20-50-800-500 & \multirow{2}{*}{Layer} & Pen. &  $^\dag\lambda_{pen} = [10,0.5,0.1,0.1]/N$ &  9-18-65-25 & 1.0 & -- \\
       & & Const. &  $\epsilon = [50\%, 30\%, 70\%, 10\%]$ &  10-14-224-29 & 0.7 & 0.91 {\color{gray} $\pm$ 0.05}\\
       \hline
    \end{tabular}
    } %

\end{table*}

\vspace{5ex}

\subsubsection*{MLPs}
\vspace{2ex}
\label{app:all_comparisons:mnist:mlp}

\begin{figure}[h!]
\setlength\tabcolsep{0pt}%
\centering
\hspace{-1.5cm}
\begin{tabular}{llccc}
 \centering
 & \hspace{2mm} & \hspace{-2mm} \textbf{$L_0$-density} & \hspace{6mm}\textbf{Parameters} & \hspace{5mm} \textbf{MACs} \\
 
\rotatebox{90}{\hspace{1.2cm} \textbf{Layer-wise}} & & \includegraphics[width=0.4\textwidth, trim={0mm 0mm 0mm 0mm}, clip]{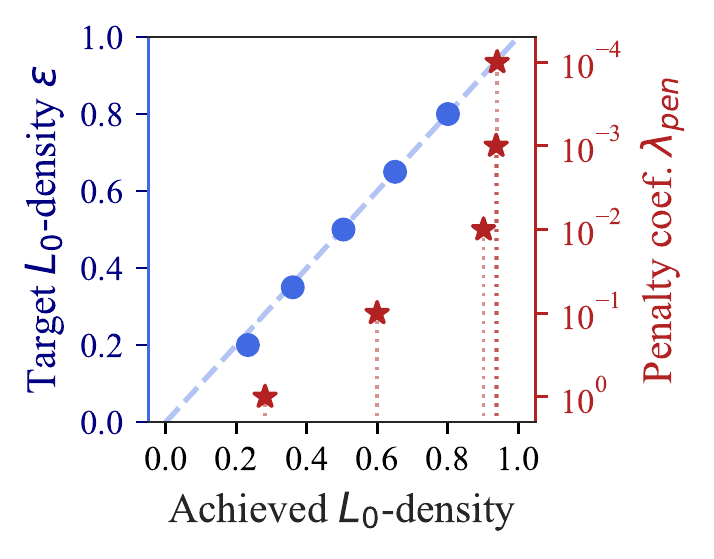}
 & \includegraphics[width=0.32\textwidth]{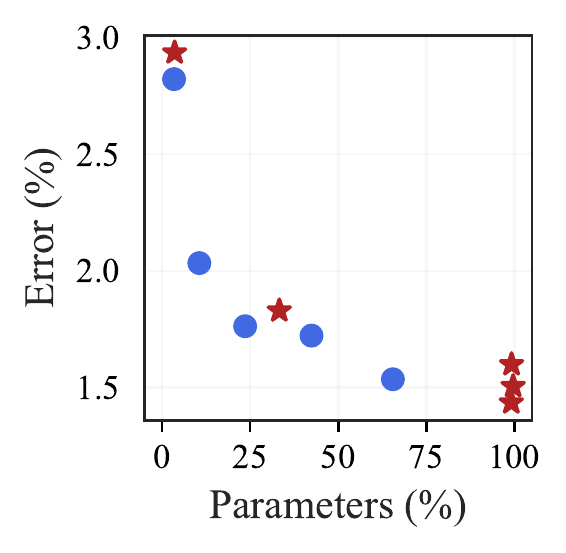}
 & \includegraphics[width=0.32\textwidth]{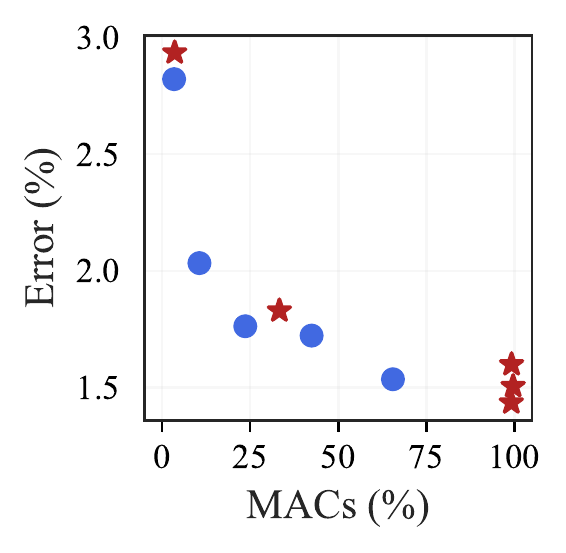} \\
\rotatebox{90}{\hspace{1.2cm} \textbf{Model-wise}} &  & \includegraphics[width=0.4\textwidth, trim={0mm 0mm 0mm 0mm}, clip]{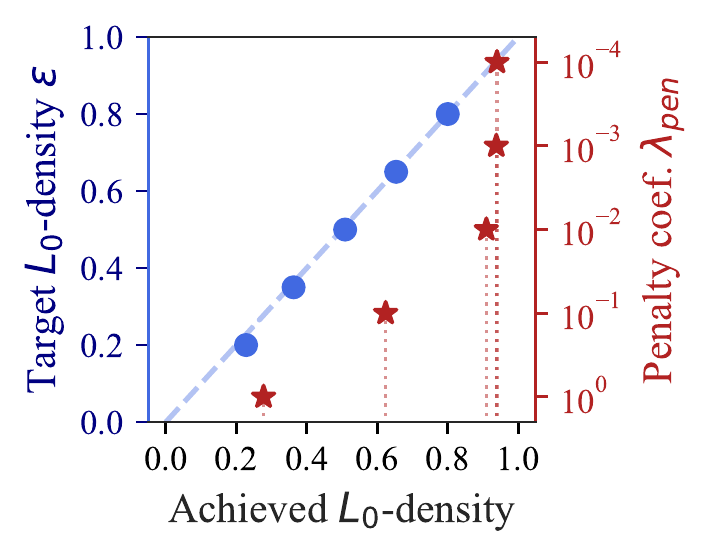}
 & \includegraphics[width=0.32\textwidth]{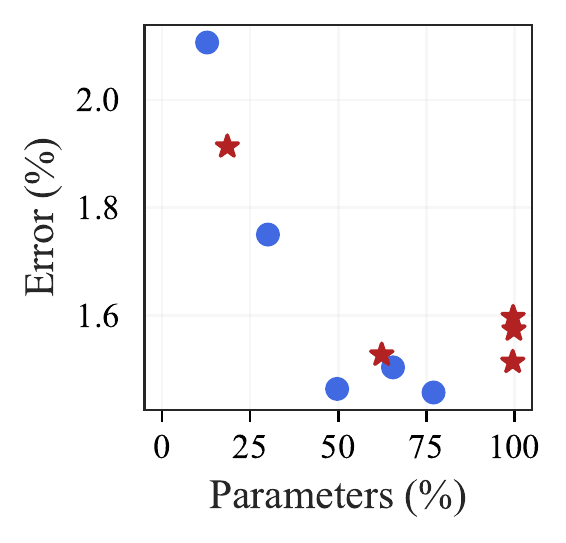}
 & \includegraphics[width=0.32\textwidth]{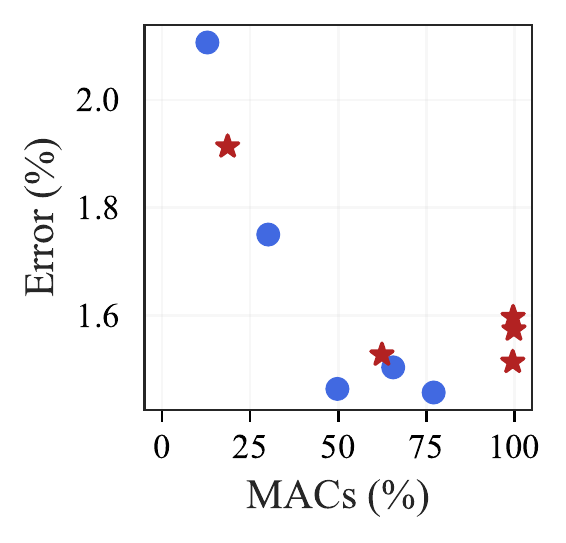} \\
\end{tabular}
    \caption[]{Training sparse MLP models on MNIST.}
    \label{fig:mnist_mlp_comparison}
\end{figure}

\clearpage
\begin{table}[h!]
\centering
\renewcommand{\arraystretch}{1.1}

\caption{Achieved density levels and performance for sparse MLP models trained on MNIST for 200 epochs. Metrics aggregated over 5 runs.}
\label{tab:mlp_control}
\vspace{1ex}

\resizebox{\textwidth}{!}{%
\begin{tabular}{clccccc}
\hline
\multirow{2}{*}{\textbf{Method}} & \multirow{2}{*}{\textbf{Hyper-params}} & \multirow{2}{*}{$L_0$-\textbf{density} (\%)} & \multirow{2}{*}{\textbf{Params} (\%)} & \multirow{2}{*}{\textbf{MACs} (\%)} & \multicolumn{2}{c}{\textbf{Val. Error } (\%)} \\
 & & & & & best & at 200 epochs \\
\hline
\hline
\multirow{5}{*}{\shortstack[l]{Constrained \\ $g \in [1:3]$ \\ {\color{gray} \textit{Layer-wise}}}} & $\epsilon_g = 20 \%$  & $23.28 {\color{gray} \, \pm 0.31}$  & $3.41 {\color{gray} \, \pm 0.16}$  & $3.43 {\color{gray} \, \pm 0.16}$  & 1.65 & $2.82 {\color{gray} \, \pm 0.31}$ \\
& $\epsilon_g = 35 \%$  & $36.02 {\color{gray} \, \pm 0.08}$  & $10.59 {\color{gray} \, \pm 0.17}$  & $10.62 {\color{gray} \, \pm 0.17}$  & 1.66 & $2.03 {\color{gray} \, \pm 0.13}$ \\
& $\epsilon_g = 50 \%$  & $50.42 {\color{gray} \, \pm 0.04}$  & $23.60 {\color{gray} \, \pm 0.14}$  & $23.63 {\color{gray} \, \pm 0.14}$  & 1.58 & $1.76 {\color{gray} \, \pm 0.15}$ \\
& $\epsilon_g = 65 \%$  & $65.07 {\color{gray} \, \pm 0.03}$  & $42.43 {\color{gray} \, \pm 0.51}$  & $42.46 {\color{gray} \, \pm 0.51}$  & 1.42 & $1.72 {\color{gray} \, \pm 0.01}$ \\
& $\epsilon_g = 80 \%$  & $80.08 {\color{gray} \, \pm 0.07}$  & $65.53 {\color{gray} \, \pm 0.76}$  & $65.55 {\color{gray} \, \pm 0.76}$  & 1.38 & $1.54 {\color{gray} \, \pm 0.05}$ \\
\hline
\multirow{5}{*}{\shortstack[l]{Penalized \\ $g \in [1:3]$ \\ {\color{gray} \textit{Layer-wise}}}}  & $\lambda_{pen}^g = 1$  & $28.16 {\color{gray} \, \pm 0.97}$  & $3.58 {\color{gray} \, \pm 0.51}$  & $3.60 {\color{gray} \, \pm 0.51}$  & 2.62 & $2.93 {\color{gray} \, \pm 0.09}$ \\
 & $\lambda_{pen}^g = 0.1$  & $59.97 {\color{gray} \, \pm 0.09}$  & $33.29 {\color{gray} \, \pm 0.06}$  & $33.32 {\color{gray} \, \pm 0.06}$  & 1.37 & $1.83 {\color{gray} \, \pm 0.07}$ \\
& $\lambda_{pen}^g = 0.01$  & $90.18 {\color{gray} \, \pm 0.22}$  & $99.15 {\color{gray} \, \pm 0.94}$  & $99.15 {\color{gray} \, \pm 0.94}$  & 1.29 & $1.44 {\color{gray} \, \pm 0.13}$ \\
& $\lambda_{pen}^g = 0.001$  & $93.77 {\color{gray} \, \pm 0.10}$  & $99.23 {\color{gray} \, \pm 0.22}$  & $99.23 {\color{gray} \, \pm 0.22}$  & 1.33 & $1.60  {\color{gray} \, \pm 0.10}$  \\
& $\lambda_{pen}^g = 0.0001$  & $94.04 {\color{gray} \, \pm 0.23}$  & $99.67 {\color{gray} \, \pm 0.38}$  & $99.67 {\color{gray} \, \pm 0.38}$  & 1.26 & $1.51 {\color{gray} \, \pm 0.04}$ \\
\hline
\multirow{5}{*}{\shortstack[l]{Constrained \\ {\color{gray} \textit{Model-wise}}}} & $\epsilon = 20 \%$  & $22.77 {\color{gray} \, \pm 0.17}$  & $12.80 {\color{gray} \, \pm 0.41}$  & $12.87 {\color{gray} \, \pm 0.41}$  & 1.40  & $2.11 {\color{gray} \, \pm 0.05}$ \\
& $\epsilon = 35 \%$  & $36.25 {\color{gray} \, \pm 0.06}$  & $30.07 {\color{gray} \, \pm 0.39}$  & $30.16 {\color{gray} \, \pm 0.39}$  & 1.37 & $1.75 {\color{gray} \, \pm 0.11}$ \\
& $\epsilon = 50 \%$  & $50.89 {\color{gray} \, \pm 0.12}$  & $49.71 {\color{gray} \, \pm 0.27}$  & $49.78 {\color{gray} \, \pm 0.27}$  & 1.20 & $1.46 {\color{gray} \, \pm 0.16}$ \\
& $\epsilon = 65 \%$  & $65.37 {\color{gray} \, \pm 0.01}$  & $65.57 {\color{gray} \, \pm 0.22}$  & $65.62 {\color{gray} \, \pm 0.22}$  & 1.27 & $1.50 {\color{gray} \, \pm 0.03}$ \\
& $\epsilon = 80 \%$  & $80.02 {\color{gray} \, \pm 0.05}$  & $77.08 {\color{gray} \, \pm 0.86}$  & $77.11 {\color{gray} \, \pm 0.86}$  & 1.16 & $1.46 {\color{gray} \, \pm 0.16}$ \\
\hline
\multirow{5}{*}{\shortstack[l]{Penalized \\ {\color{gray} \textit{Model-wise}}}} & $\lambda_{pen} = 1 $  & $27.73 {\color{gray} \, \pm 0.15}$  & $18.54 {\color{gray} \, \pm 0.51}$  & $18.62 {\color{gray} \, \pm 0.52}$  & 1.48 & $1.91 {\color{gray} \, \pm 0.07}$ \\
& $\lambda_{pen} = 0.1 $  & $62.38 {\color{gray} \, \pm 0.16}$  & $62.37 {\color{gray} \, \pm 0.3}$  & $62.43 {\color{gray} \, \pm 0.3}$  & 1.30 & $1.53 {\color{gray} \, \pm 0.13}$ \\
& $\lambda_{pen} = 0.01$  & $90.98 {\color{gray} \, \pm 0.18}$  & $99.56 {\color{gray} \, \pm 0.43}$  & $99.56 {\color{gray} \, \pm 0.43}$  & 1.34 & $1.51 {\color{gray} \, \pm 0.11}$ \\
& $\lambda_{pen} = 0.001 $  & $93.81 {\color{gray} \, \pm 0.17}$  & $99.89 {\color{gray} \, \pm 0.22}$  & $99.89 {\color{gray} \, \pm 0.22}$  & 1.24 & $1.57 {\color{gray} \, \pm 0.09}$ \\
& $\lambda_{pen} = 0.0001$  & $93.99 {\color{gray} \, \pm 0.06}$  & $99.67 {\color{gray} \, \pm 0.38}$  & $99.67 {\color{gray} \, \pm 0.38}$  & 1.32 & $1.60 {\color{gray} \, \pm 0.17}$ \\
\hline
\end{tabular}
}
\end{table}

\vspace{1cm}
\subsubsection*{LeNet5}
\label{app:all_comparisons:mnist:lenet5}
\begin{figure}[h!]
\setlength\tabcolsep{0pt}%
\centering
\hspace{-1.5cm}
\begin{tabular}{llccc}
 \centering
 & \hspace{2mm} & \hspace{-2mm} \textbf{$L_0$-density} & \hspace{6mm}\textbf{Parameters} & \hspace{5mm} \textbf{MACs} \\
 
\rotatebox{90}{\hspace{1.2cm} \textbf{Layer-wise}} & & \includegraphics[width=0.4\textwidth, trim={0mm 0mm 0mm 0mm}, clip]{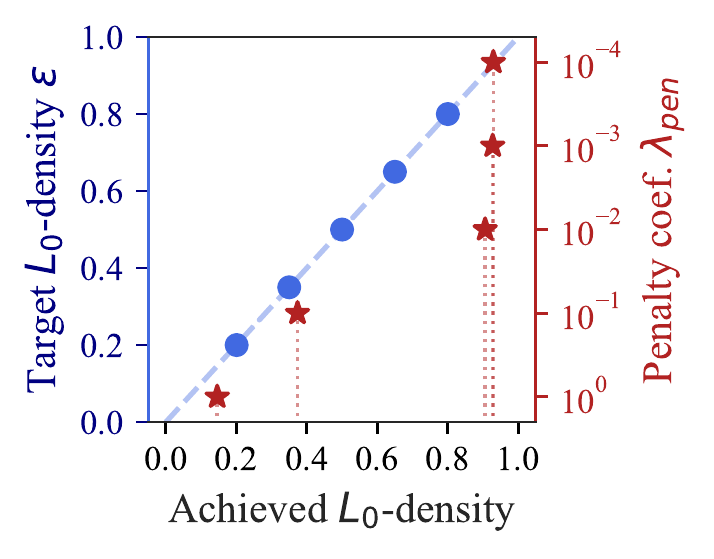}
 & \includegraphics[width=0.32\textwidth]{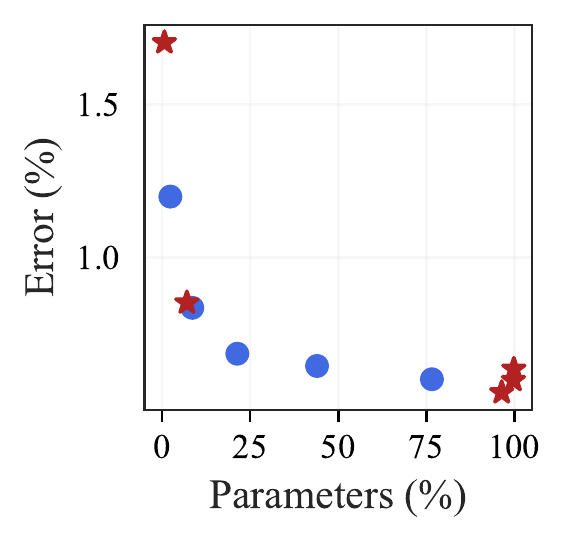}
 & \includegraphics[width=0.32\textwidth]{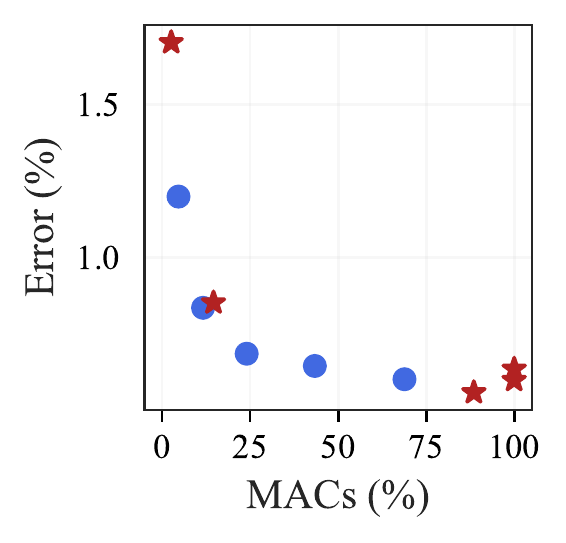} \\
\rotatebox{90}{\hspace{1.2cm} \textbf{Model-wise}} &  & \includegraphics[width=0.4\textwidth, trim={0mm 0mm 0mm 0mm}, clip]{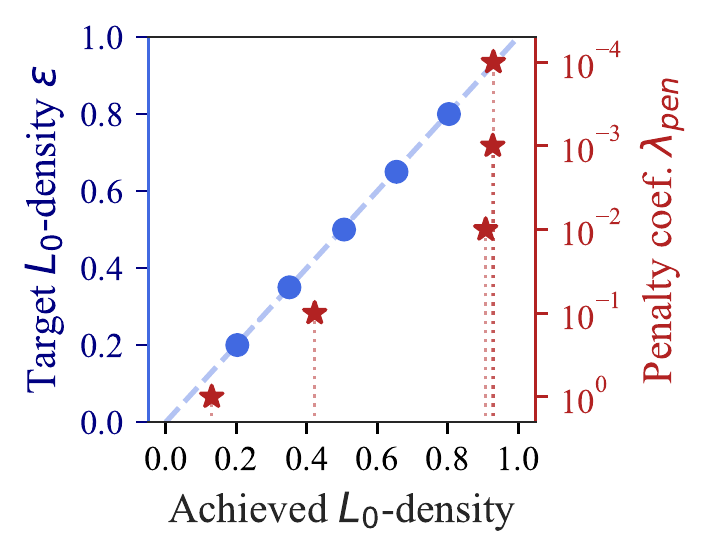}
 & \includegraphics[width=0.32\textwidth]{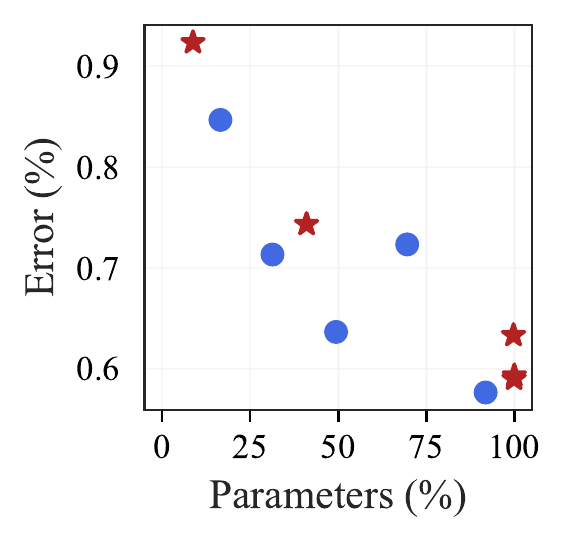}
 & \includegraphics[width=0.32\textwidth]{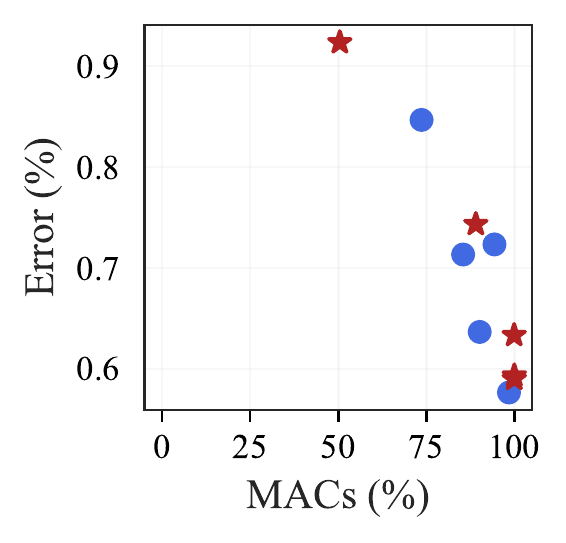} \\
\end{tabular}
    \caption[]{Training sparse LeNet models on MNIST.}
    \label{fig:mnist_lenet_comparison}
\end{figure}

\clearpage
\vspace{2cm}
\begin{table}[h!]
\centering
\renewcommand{\arraystretch}{1.1}

\caption{Achieved density levels and performance for sparse LeNet models trained on MNIST for 200 epochs. Metrics aggregated over 5 runs.}
\label{tab:lenet_control}
\vspace{1ex}

\resizebox{\textwidth}{!}{%
    
\begin{tabular}{clccccc}
\hline
\multirow{2}{*}{\textbf{Method}} & \multirow{2}{*}{\textbf{Hyper-params}} & \multirow{2}{*}{$L_0$-\textbf{density} (\%)} & \multirow{2}{*}{\textbf{Params} (\%)} & \multirow{2}{*}{\textbf{MACs} (\%)} & \multicolumn{2}{c}{\textbf{Val. Error } (\%)} \\
 & & & & & best & at 200 epochs \\
\hline
\hline
\multirow{5}{*}{\shortstack[l]{Constrained \\ $g \in [1:4]$ \\ {\color{gray} \textit{Layer-wise}}}}& $\epsilon_g = 20 \%$ & $20.09  {\color{gray} \, \pm 0.07}$ & $2.36 {\color{gray} \, \pm 0.14}$ & $4.68  {\color{gray} \,  \pm 0.68}$ & 0.81 & $1.20 {\color{gray} \, \pm 0.08}$ \\
& $\epsilon_g = 35 \%$ & $35.01 {\color{gray} \, \pm 0.01}$ & $8.59 {\color{gray} \, \pm 0.03}$ & $11.65 {\color{gray} \, \pm 1.15}$ & 0.69 & $0.84 {\color{gray} \, \pm 0.08}$ \\
& $\epsilon_g = 50 \%$ & $50.02 {\color{gray} \, \pm 0.01}$ & $21.38 {\color{gray} \, \pm 0.77}$ & $24.03 {\color{gray} \, \pm 1.64}$ & 0.56 & $0.69 {\color{gray} \, \pm 0.07}$ \\
& $\epsilon_g = 65 \%$ & $65.02 {\color{gray} \, \pm 0.01}$ & $44.01 {\color{gray} \, \pm 2.16}$ & $43.38 {\color{gray} \, \pm 2.39}$ & 0.53 & $0.65 {\color{gray} \, \pm 0.09}$ \\
& $\epsilon_g = 80 \%$ & $80.04 {\color{gray} \, \pm 0.02}$ & $76.62 {\color{gray} \, \pm 0.99}$ & $68.83  {\color{gray} \, \pm 1.40}$  & 0.45 & $0.60 {\color{gray} \, \pm 0.04}$ \\
\hline
\multirow{5}{*}{\shortstack[l]{Penalized \\ $g \in [1:4]$ \\ {\color{gray} \textit{Layer-wise}}}}  & $\lambda_{pen}^g = 1$ & $14.58 {\color{gray} \, \pm 0.44}$ & $0.67 {\color{gray} \, \pm 0.13}$ & $2.60 {\color{gray} \, \pm 0.17}$ & 0.46 & $1.7 {\color{gray} \, \pm 0.14}$ \\
& $\lambda_{pen}^g = 0.1$ & $37.38 {\color{gray} \, \pm 0.49}$ & $7.06 {\color{gray} \, \pm 0.36}$ & $14.62 {\color{gray} \, \pm 1.68}$ & 0.54 & $0.85 {\color{gray} \, \pm 0.02}$ \\
& $\lambda_{pen}^g = 0.01$ & $90.63 {\color{gray} \, \pm 0.20}$ & $96.43 {\color{gray} \, \pm 0.83}$ & $88.52 {\color{gray} \, \pm 7.17}$ & 0.40 & $0.56 {\color{gray} \, \pm 0.03}$ \\
& $\lambda_{pen}^g = 0.001$ & $92.77 {\color{gray} \, \pm 0.06}$ & $99.75 {\color{gray} \, \pm 0.12}$ & $99.95 {\color{gray} \, \pm 0.02}$ & 0.47 & $0.60 {\color{gray} \, \pm 0.05}$ \\
& $\lambda_{pen}^g = 0.0001$ & $92.96 {\color{gray} \, \pm 0.04}$ & $99.87 {\color{gray} \, \pm 0.12}$ & $99.98 {\color{gray} \, \pm 0.02}$ & 0.47 & $0.64 {\color{gray} \, \pm 0.06}$ \\
\hline
\multirow{5}{*}{\shortstack[l]{Constrained \\ {\color{gray} \textit{Model-wise}}}} & $\epsilon = 20 \%$ & $20.27 {\color{gray} \, \pm 0.30}$ & $16.58 {\color{gray} \, \pm 0.73}$ & $73.67 {\color{gray} \, \pm 0.14}$ & 0.54 & $0.85 {\color{gray} \, \pm 0.09}$ \\
& $\epsilon = 35 \%$ & $35.06 {\color{gray} \, \pm 0.09}$ & $31.36 {\color{gray} \, \pm 0.27}$ & $85.47 {\color{gray} \, \pm 0.85}$ & 0.55 & $0.71 {\color{gray} \, \pm 0.03}$ \\
& $\epsilon = 50 \%$ & $50.58 {\color{gray} \, \pm 0.09}$ & $49.41 {\color{gray} \, \pm 0.29}$ & $90.18 {\color{gray} \, \pm 0.94}$ & 0.44 & $0.64 {\color{gray} \, \pm 0.03}$ \\
& $\epsilon = 65 \%$ & $65.47 {\color{gray} \, \pm 0.05}$ & $69.60 {\color{gray} \, \pm 1.06}$ & $94.37 {\color{gray} \, \pm 0.20}$  & 0.44 & $0.72 {\color{gray} \, \pm 0.03}$ \\
& $\epsilon = 80 \%$ & $80.36 {\color{gray} \, \pm 0.08}$ & $91.86 {\color{gray} \, \pm 0.79}$ & $98.49 {\color{gray} \, \pm 0.15}$ & 0.46 & $0.58 {\color{gray} \, \pm 0.03}$ \\
\hline
\multirow{5}{*}{\shortstack[l]{Penalized \\ {\color{gray} \textit{Model-wise}}}} & $\lambda_{pen} = 1 $ & $13.01 {\color{gray} \, \pm 0.81}$ & $8.79  {\color{gray} \, \pm 0.80}$  & $50.47 {\color{gray} \, \pm 4.74}$ & 0.71 & $0.92 {\color{gray} \, \pm 0.11}$ \\
& $\lambda_{pen} = 0.1 $  & $42.25 {\color{gray} \, \pm 0.83}$ & $41.03 {\color{gray} \, \pm 1.12}$ & $89.08 {\color{gray} \, \pm 0.21}$ & 0.45 & $0.74 {\color{gray} \, \pm 0.04}$ \\
& $\lambda_{pen} = 0.01 $  & $90.78 {\color{gray} \, \pm 0.20}$ & $99.78 {\color{gray} \, \pm 0.20}$ & $99.96 {\color{gray} \, \pm 0.04}$ & 0.44 & $0.63 {\color{gray} \, \pm 0.05}$ \\
& $\lambda_{pen} = 0.001 $ & $92.80 {\color{gray} \, \pm 0.01}$ & $100 {\color{gray} \, \pm 0.00}$ & $100 {\color{gray} \, \pm 0.00}$ & 0.43 & $0.59 {\color{gray} \, \pm 0.01}$ \\
& $\lambda_{pen} = 0.0001 $ & $92.97 {\color{gray} \, \pm 0.03}$ & $99.94 {\color{gray} \, \pm 0.12}$ & $99.99 {\color{gray} \, \pm 0.02}$ & 0.45 & $0.59 {\color{gray} \, \pm 0.07}$ \\
\hline
\end{tabular}

}

\end{table}

\subsection{CIFAR-10}
\label{app:all_comparisons:cifar10}
\vspace{1cm}
\begin{table*}[h!]

    \caption{Achieved density levels and performance for sparse WideResNets-28-10 models trained on CIFAR-10 for 200 epochs. Metrics aggregated over 5 runs. { $^\dag$Result reported by \citet{louizos2017learning}, with $N$ denoting the training set size (see \cref{app:normalized_l0}).}}
    \label{tab:cifar10}
    \vspace{1ex}

    \renewcommand{\arraystretch}{1.1}
    \centering
    \resizebox{\textwidth}{!}{%
    \begin{tabular}{clccccccc}
      \hline
      \multirow{3}{*}{\textbf{Method}} & \multirow{3}{*}{\textbf{Hyper-params}} & \multirow{3}{*}{$\eta_{\text{primal}}^{\vphi}$}  & \multirow{3}{*}{$L_0$-\textbf{density} (\%)} & \multirow{3}{*}{\textbf{Params} (\%)} & \multirow{3}{*}{\textbf{MACs} (\%)} & \multicolumn{2}{c}{\textbf{Val. Error} (\%)}  \\
      \cline{7-8} & & & & & & \multirow{2}{*}{best} & at 200 epochs \\
       & & & & & & & (avg {\color{gray} $\pm$ 95\% CI)}\\
      \hline
       \multirow{5}{*}{Penalized} & $^\dag\lambda_{pen} = 0.001/N$ & 0.1 & \multicolumn{1}{c}{\textbf{--}} &  \multicolumn{1}{c}{\textbf{--}} &  \multicolumn{1}{c}{\textbf{--}} &  \multicolumn{1}{c}{${3.83}$} & \multicolumn{1}{c}{\textbf{--}} \\
       & $^\dag\lambda_{pen} = 0.002/N$ & 0.1 & \multicolumn{1}{c}{\textbf{--}} & \multicolumn{1}{c}{\textbf{--}} &  \multicolumn{1}{c}{\textbf{--}} & \multicolumn{1}{c}{${3.93}$} & \multicolumn{1}{c}{\textbf{--}} \\
       \cline{2-8}
       & $\lambda_{pen} = 0.001$ & 0.1 & $92.30 {\color{gray} \, \pm 0.01}$ & $99.84 {\color{gray} \, \pm 0.00}$ & $100 {\color{gray} \, \pm 0.00}$ & $4.23$ & $4.56 {\color{gray} \, \pm 0.18}$ \\
       & $\lambda_{pen} = 0.001$ & 6 & $91.19 {\color{gray} \, \pm 0.16}$ & $93.98 {\color{gray} \, \pm 0.24}$ & $91.57 {\color{gray} \, \pm 0.35}$ & $3.75$ & $4.04 {\color{gray} \, \pm 0.15}$ \\
       & $\lambda_{pen} = 0.002$ & 6 & $91.36 {\color{gray} \, \pm 0.11}$ & $94.26 {\color{gray} \, \pm 0.22}$ & $92.10 {\color{gray} \, \pm 0.46}$ & $3.62$ & $4.05 {\color{gray} \, \pm 0.14}$ \\
       \cline{1-8}
       \multirow{2}{*}{Constrained} & $\epsilon_g = 100\%$ & 0.1 & $92.34 {\color{gray} \, \pm 0.01}$ & $99.84 {\color{gray} \, \pm 0.00}$ & $100 {\color{gray} \, \pm 0.00}$ & $4.18$ & $4.63 {\color{gray} \, \pm 0.14}$ \\
       \scriptsize \multirow{2}{*}{$g \in [1:12]$} & $\epsilon_g = 100 \%$ & 6 & $91.63 {\color{gray} \, \pm 0.27}$ & $94.52 {\color{gray} \, \pm 0.26}$ & $92.75 {\color{gray} \, \pm 0.58}$ & $3.74$ & $4.12 {\color{gray} \, \pm 0.08}$ \\
       & $\epsilon_g = 70\%$ & 6 & $70.00 {\color{gray} \, \pm 0.00}$ & $69.87 {\color{gray} \, \pm 0.20}$ & $69.55 {\color{gray} \, \pm 0.17}$ & $3.76$ & $4.10 {\color{gray} \, \pm 0.16}$ \\
       \hline
    \end{tabular}
    }
\end{table*}

\clearpage
\begin{figure}[h!]
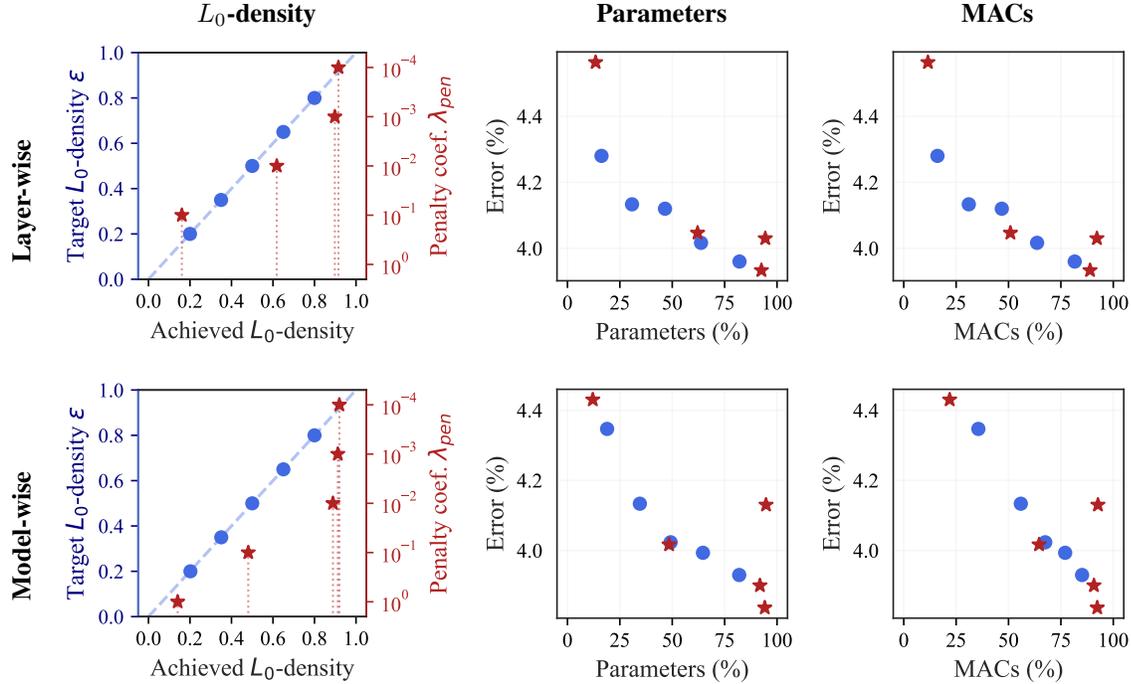

\setlength\tabcolsep{0pt}%
\centering
\hspace{-1.5cm}
\begin{tabular}{llccc}
 \centering
 & \hspace{2mm} & \hspace{-2mm} \textbf{$L_0$-density} & \hspace{6mm}\textbf{Parameters} & \hspace{5mm} \textbf{MACs} \\
 
\rotatebox{90}{\hspace{1.2cm} \textbf{Layer-wise}} & & \includegraphics[width=0.4\textwidth, trim={0mm 0mm 0mm 0mm}, clip]{neurips2022/figs/control/cifar10/ResNet-28-10_layerwise.pdf}
 & \includegraphics[width=0.32\textwidth]{neurips2022/figs/error_vs_params/cifar10/ResNet-28-10_layerwise.pdf}
 & \includegraphics[width=0.32\textwidth]{neurips2022/figs/error_vs_macs/cifar10/ResNet-28-10_layerwise.pdf} \\
\rotatebox{90}{\hspace{1.2cm} \textbf{Model-wise}} &  & \includegraphics[width=0.4\textwidth, trim={0mm 0mm 0mm 0mm}, clip]{neurips2022/figs/control/cifar10/ResNet-28-10_modelwise.pdf}
 & \includegraphics[width=0.32\textwidth]{neurips2022/figs/error_vs_params/cifar10/ResNet-28-10_modelwise.pdf}
 & \includegraphics[width=0.32\textwidth]{neurips2022/figs/error_vs_macs/cifar10/ResNet-28-10_modelwise.pdf} \\
\end{tabular}
\caption[]{Training sparse WideResNet-28-10 models on CIFAR-10. { \textit{This figure is the same as \cref{fig:cifar10_comparison}. We repeat it here for the reader's convenience.}}}
    \label{fig:cifar10_APPENDIX_comparison}
\end{figure}
\begin{table}[h!]
\centering

\caption{Achieved density levels and performance for sparse WideResNet-28-10 models trained on CIFAR-10 for 200 epochs. Metrics aggregated over 5 runs.}
\label{tab:cifar10_control}
\vspace{1ex}

\resizebox{\textwidth}{!}{%
    
\begin{tabular}{llccccc}
\hline
\multirow{2}{*}{\textbf{Method}} & \multirow{2}{*}{\textbf{Hyper-params}} & \multirow{2}{*}{$L_0$-\textbf{density} (\%)} & \multirow{2}{*}{\textbf{Params} (\%)} & \multirow{2}{*}{\textbf{MACs} (\%)} & \multicolumn{2}{c}{\textbf{Val. Error } (\%)} \\
 & & & & & best & at 200 epochs \\
\hline
\hline
\multirow{5}{*}{\shortstack[l]{Constrained \\ $g \in [1:12]$ \\ {\color{gray} \textit{Layer-wise}}}} & $\epsilon_g = 20 \%$ & $20.00 {\color{gray} \, \pm 0.02}$ & $16.24 {\color{gray} \, \pm 0.20}$ & $16.10 {\color{gray} \, \pm 0.23}$ & 4.05 & $4.28 {\color{gray} \, \pm 0.16}$ \\
& $\epsilon_g = 35 \%$ & $34.99 {\color{gray} \, \pm 0.01}$ & $30.80 {\color{gray} \, \pm 0.34}$ & $31.05 {\color{gray} \, \pm 0.23}$ & 3.8 & $4.13 {\color{gray} \, \pm 0.13}$ \\
& $\epsilon_g = 50 \%$ & $50.00 {\color{gray} \, \pm 0.01}$ & $46.54 {\color{gray} \, \pm 0.21}$ & $46.79 {\color{gray} \, \pm 0.28}$ & 3.87 & $4.12 {\color{gray} \, \pm 0.21}$ \\
& $\epsilon_g = 65 \%$ & $64.99 {\color{gray} \, \pm 0.01}$ & $63.74 {\color{gray} \, \pm 0.33}$ & $63.63 {\color{gray} \, \pm 0.27}$ & 3.7 & $4.02 {\color{gray} \, \pm 0.11}$ \\
& $\epsilon_g = 80 \%$ & $80.00 {\color{gray} \, \pm 0.00}$ & $82.10 {\color{gray} \, \pm 0.32}$ & $81.60 {\color{gray} \, \pm 0.34}$ & 3.69 & $3.96 {\color{gray} \, \pm 0.15}$ \\
\hline
\multirow{5}{*}{\shortstack[l]{Penalized \\ $g \in [1:12]$ \\ {\color{gray} \textit{Layer-wise}}}}
& $\lambda_{pen}^g = 1$ & \multicolumn{5}{c}{\color{gray}{------ \;\;\; \textit{Failed \; due \; to \; sparsity \; collapse}} \;\;\; ------ } \\
& $\lambda_{pen}^g = 0.1$ & $16.09 {\color{gray} \, \pm 0.14}$ & $13.43 {\color{gray} \, \pm 0.11}$ & $11.54 {\color{gray} \, \pm 0.29}$ & 4.23 & $4.56 {\color{gray} \, \pm 0.09}$ \\
& $\lambda_{pen}^g = 0.01$ & $61.83 {\color{gray} \, \pm 0.86}$ & $62.10 {\color{gray} \, \pm 0.97}$ & $50.91 {\color{gray} \, \pm 0.62}$ & 3.83 & $4.05 {\color{gray} \, \pm 0.02}$ \\
& $\lambda_{pen}^g = 0.001$ & $89.81 {\color{gray} \, \pm 0.06}$ & $92.55 {\color{gray} \, \pm 0.13}$ & $88.91 {\color{gray} \, \pm 0.15}$ & 3.7 & $3.93 {\color{gray} \, \pm 0.15}$ \\
& $\lambda_{pen}^g = 0.0001$ & $91.54 {\color{gray} \, \pm 0.33}$ & $94.41 {\color{gray} \, \pm 0.48}$ & $92.18 {\color{gray} \, \pm 0.94}$ & 3.81 & $4.03 {\color{gray} \, \pm 0.05}$ \\
\hline
\multirow{5}{*}{\shortstack[l]{Constrained \\ {\color{gray} \textit{Model-wise}}}} & $\epsilon = 20 \%$ & $20.22 {\color{gray} \, \pm 0.01}$ & $18.91 {\color{gray} \, \pm 0.03}$ & $35.62 {\color{gray} \, \pm 0.84}$ & 3.95 & $4.35 {\color{gray} \, \pm 0.25}$ \\
& $\epsilon = 35 \%$  & $35.07 {\color{gray} \, \pm 0.03}$ & $34.55 {\color{gray} \, \pm 0.08}$ & $55.87 {\color{gray} \, \pm 1.23}$ & 3.76 & $4.13 {\color{gray} \, \pm 0.27}$ \\
& $\epsilon = 50 \%$ & $50.00 {\color{gray} \, \pm 0.01}$ & $49.21 {\color{gray} \, \pm 0.33}$ & $67.53 {\color{gray} \, \pm 0.08}$ & 3.79 & $4.02 {\color{gray} \, \pm 0.06}$ \\
& $\epsilon = 65 \%$ & $65.01 {\color{gray} \, \pm 0.01}$ & $64.65 {\color{gray} \, \pm 0.34}$ & $77.02 {\color{gray} \, \pm 0.43}$ & 3.80  & $3.99 {\color{gray} \, \pm 0.14}$ \\
& $\epsilon = 80 \%$ & $80.01 {\color{gray} \, \pm 0.00}$ & $81.96 {\color{gray} \, \pm 0.12}$ & $85.11 {\color{gray} \, \pm 0.58}$ & 3.83 & $3.93 {\color{gray} \, \pm 0.09}$ \\
\hline
\multirow{5}{*}{\shortstack[l]{Penalized \\ {\color{gray} \textit{Model-wise}}}} & $\lambda_{pen} = 1$ & $14.07 {\color{gray} \, \pm 0.23}$ & $12.09 {\color{gray} \, \pm 0.24}$ & $21.92 {\color{gray} \, \pm 0.59}$ & 4.23 & $4.43 {\color{gray} \, \pm 0.10}$ \\
& $\lambda_{pen} = 0.1$ & $48.08 {\color{gray} \, \pm 2.05}$ & $48.50 {\color{gray} \, \pm 2.27}$ & $64.58 {\color{gray} \, \pm 1.57}$ & 3.87 & $4.02 {\color{gray} \, \pm 0.22}$ \\
& $\lambda_{pen} = 0.01$ & $88.89 {\color{gray} \, \pm 0.31}$ & $91.85 {\color{gray} \, \pm 0.47}$ & $90.76 {\color{gray} \, \pm 0.77}$ & 3.72 & $3.90 {\color{gray} \, \pm 0.18}$ \\
& $\lambda_{pen} = 0.001$ & $91.24 {\color{gray} \, \pm 0.20}$ & $94.17 {\color{gray} \, \pm 0.24}$ & $92.34 {\color{gray} \, \pm 0.22}$ & 3.59 & $3.84 {\color{gray} \, \pm 0.25}$ \\
& $\lambda_{pen} = 0.0001$ & $91.98 {\color{gray} \, \pm 0.42}$ & $94.75 {\color{gray} \, \pm 0.31}$ & $92.68 {\color{gray} \, \pm 0.36}$ & 3.91 & $4.13 {\color{gray} \, \pm 0.09}$ \\
\hline
\end{tabular}
}
\end{table}

\clearpage
\subsection{CIFAR-100}
\label{app:all_comparisons:cifar100}
\begin{table*}[h]

    \caption{Achieved density levels and performance for sparse WideResNets-28-10 models trained on CIFAR-100 for 200 epochs. Metrics aggregated over 5 runs. {$^\dag$Result reported by \citet{louizos2017learning}, with $N$ denoting the training set size (see \cref{app:normalized_l0}).}}
    \label{tab:cifar100}
    \vspace{1ex}

    \renewcommand{\arraystretch}{1.1}
    \centering
    \resizebox{\textwidth}{!}{%
    \begin{tabular}{clcccccc}
      \hline
      \multirow{3}{*}{\textbf{Method}} & \multirow{3}{*}{\textbf{Hyper-params}} & \multirow{3}{*}{$\eta_{\text{primal}}^{\vphi}$}  & \multirow{3}{*}{$L_0$-\textbf{density} (\%)} & \multirow{3}{*}{\textbf{Params} (\%)} & \multirow{3}{*}{\textbf{MACs} (\%)} & \multicolumn{2}{c}{\textbf{Val. Error} (\%)}  \\
      \cline{7-8} & & & & & & \multirow{2}{*}{best} & at 200 epochs \\
       & & & & & & & (avg {\color{gray} $\pm$ 95\% CI)}\\
      \hline
      \hline
       \multirow{5}{*}{Penalized} & $^\dag\lambda_{pen} = 0.001/N$ & 0.1 & \multicolumn{1}{c}{\textbf{--}} &  \multicolumn{1}{c}{\textbf{--}} &  \multicolumn{1}{c}{\textbf{--}} &  \multicolumn{1}{c}{${18.75}$} & \multicolumn{1}{c}{\textbf{--}} \\
       & $^\dag\lambda_{pen} = 0.002/N$ & 0.1 & \multicolumn{1}{c}{\textbf{--}} & \multicolumn{1}{c}{\textbf{--}} &  \multicolumn{1}{c}{\textbf{--}} & \multicolumn{1}{c}{${19.04}$} & \multicolumn{1}{c}{\textbf{--}} \\
       \cline{2-8}
       & $\lambda_{pen} = 0.001$ & 0.1 & $93.20 {\color{gray} \, \pm 0.01}$ & $100.00 {\color{gray} \, \pm 0.00}$ & $100.00 {\color{gray} \, \pm 0.00}$ & $21.01$ & $21.70 {\color{gray} \, \pm 0.19}$ \\
       & $\lambda_{pen} = 0.001$ & 6 & $90.64 {\color{gray} \, \pm 0.32}$ & $90.88 {\color{gray} \, \pm 0.41}$ & $89.94 {\color{gray} \, \pm 0.71}$ & $18.51$ & $19.14 {\color{gray} \, \pm 0.21}$ \\
       & $\lambda_{pen} = 0.002$ & 6 & $90.13 {\color{gray} \, \pm 0.45}$ & $90.19 {\color{gray} \, \pm 0.38}$ & $89.52 {\color{gray} \, \pm 0.57}$ & $18.99$ & $19.24 {\color{gray} \, \pm 0.14}$ \\
       \cline{1-8}
       \multirow{2}{*}{Constrained} & $\epsilon_g = 100\%$ & 0.1 & $93.20 {\color{gray} \, \pm 0.01}$ & $100.00 {\color{gray} \, \pm 0.00}$ & $100.00 {\color{gray} \, \pm 0.00}$ & $21.02$ & $21.66 {\color{gray} \, \pm 0.21}$  \\
       \scriptsize \multirow{2}{*}{$g \in [1:12]$} & $\epsilon_g = 100 \%$ & 6 & $90.77 {\color{gray} \, \pm 0.31}$ & $90.99 {\color{gray} \, \pm 0.25}$ & $89.74 {\color{gray} \, \pm 0.29}$ & $18.68$ & $19.08 {\color{gray} \, \pm 0.16}$\\
       & $\epsilon_g = 70\%$ & 6 & $69.99 {\color{gray} \, \pm 0.01}$ & $68.62 {\color{gray} \, \pm 0.08}$ & $68.59 {\color{gray} \, \pm 0.22}$ & $18.88$ & $19.37 {\color{gray} \, \pm 0.15}$ \\
       \hline
    \end{tabular}
    }
\end{table*}

\vspace{2cm}
\begin{figure}[h!]
\setlength\tabcolsep{0pt}%
\centering
\hspace{-1.5cm}
\begin{tabular}{llccc}
 \centering
 & \hspace{2mm} & \hspace{-2mm} \textbf{$L_0$-density} & \hspace{6mm}\textbf{Parameters} & \hspace{5mm} \textbf{MACs} \\
 
\rotatebox{90}{\hspace{1.2cm} \textbf{Layer-wise}} & & \includegraphics[width=0.4\textwidth, trim={0mm 0mm 0mm 0mm}, clip]{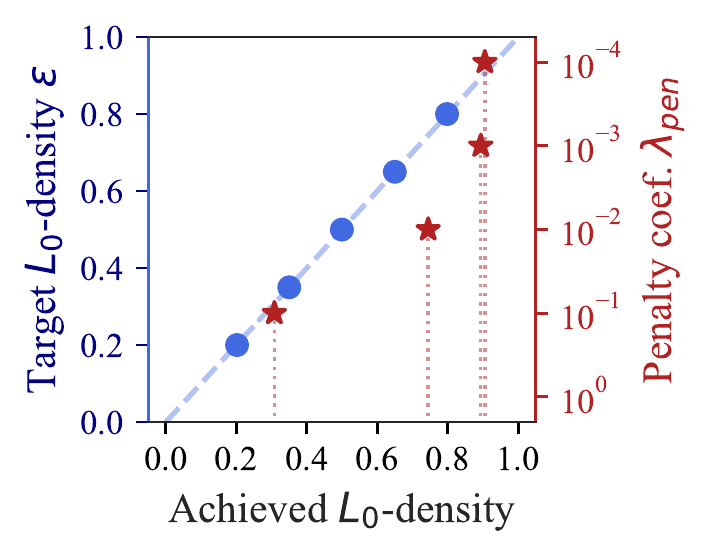}
 & \includegraphics[width=0.32\textwidth]{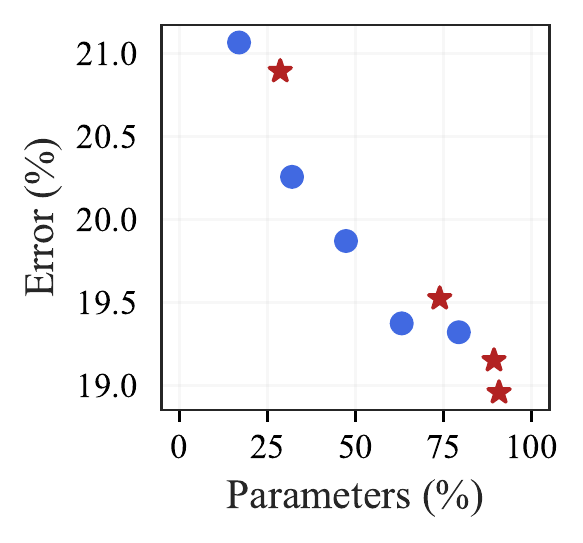}
 & \includegraphics[width=0.32\textwidth]{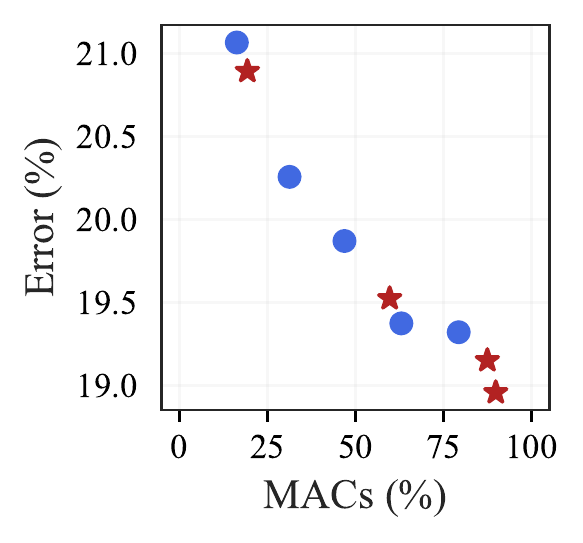} \\
\rotatebox{90}{\hspace{1.2cm} \textbf{Model-wise}} &  & \includegraphics[width=0.4\textwidth, trim={0mm 0mm 0mm 0mm}, clip]{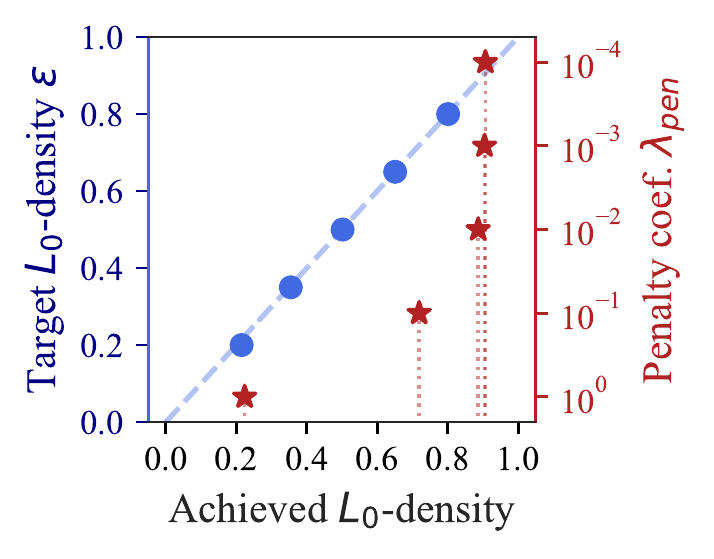}
 & \includegraphics[width=0.32\textwidth]{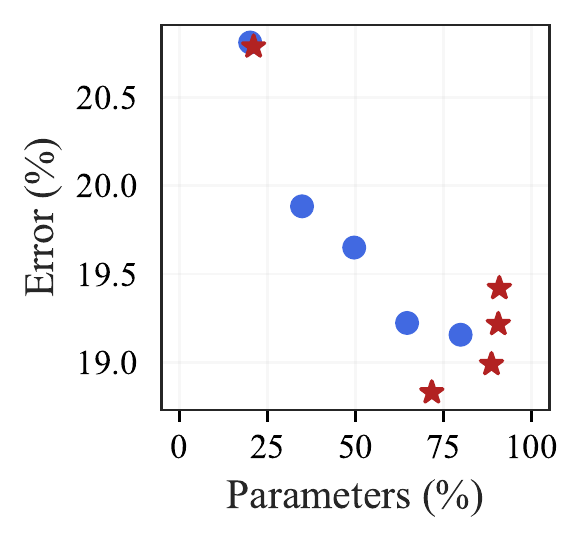}
 & \includegraphics[width=0.32\textwidth]{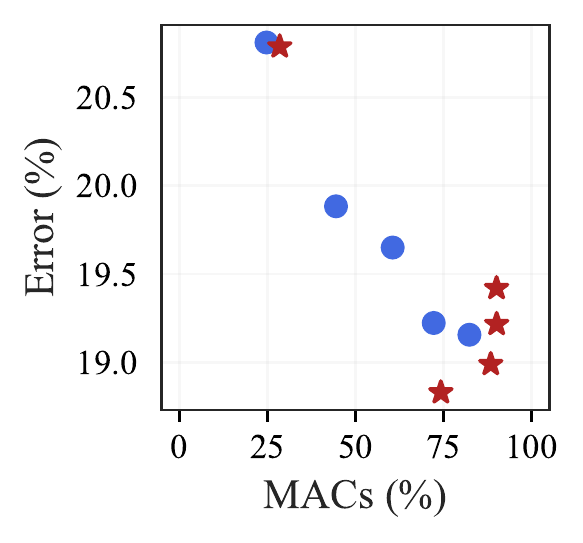} \\
\end{tabular}
    \caption[]{Training sparse WideResNet-28-10 models on CIFAR-100.}
    \label{fig:cifar100_comparison}
\end{figure}

\clearpage
\begin{table}[h!]
\centering
\renewcommand{\arraystretch}{1.1}

\caption{Achieved density levels and performance for sparse WideResNet-28-10 models trained on CIFAR-100 for 200 epochs. Metrics aggregated over 5 runs.}
\label{tab:cifar100_control}
\vspace{1ex}

\resizebox{\textwidth}{!}{%
    
\begin{tabular}{llccccc}
\hline
\multirow{2}{*}{\textbf{Method}} & \multirow{2}{*}{\textbf{Hyper-params}} & \multirow{2}{*}{$L_0$-\textbf{density} (\%)} & \multirow{2}{*}{\textbf{Params} (\%)} & \multirow{2}{*}{\textbf{MACs} (\%)} & \multicolumn{2}{c}{\textbf{Val. Error } (\%)} \\
 & & & & & best & at 200 epochs \\
\hline
\hline
\multirow{5}{*}{\shortstack[l]{Constrained \\ $g \in [1:12]$ \\ {\color{gray} \textit{Layer-wise}}}}& $\epsilon_g = 20 \%$ & $20.20 {\color{gray} \, \pm 0.00}$ & $16.94 {\color{gray} \, \pm 0.06}$ & $16.30 {\color{gray} \, \pm 0.08}$ & 20.67 & $21.07 {\color{gray} \, \pm 0.19}$ \\
& $\epsilon_g = 35 \%$ & $35.03 {\color{gray} \, \pm 0.02}$ & $31.97 {\color{gray} \, \pm 0.14}$ & $31.27 {\color{gray} \, \pm 0.08}$ & 19.71 & $20.26 {\color{gray} \, \pm 0.37}$ \\
& $\epsilon_g = 50 \%$ & $49.97 {\color{gray} \, \pm 0.03}$ & $47.30 {\color{gray} \, \pm 0.31}$ & $46.87 {\color{gray} \, \pm 0.31}$ & 19.58 & $19.87 {\color{gray} \, \pm 0.21}$ \\
& $\epsilon_g = 65 \%$ & $64.99 {\color{gray} \, \pm 0.02}$ & $63.11 {\color{gray} \, \pm 0.05}$ & $63.01 {\color{gray} \, \pm 0.17}$ & 18.89 & $19.37 {\color{gray} \, \pm 0.27}$ \\
& $\epsilon_g = 80 \%$ & $79.82 {\color{gray} \, \pm 0.15}$ & $79.33 {\color{gray} \, \pm 0.30}$ & $79.29 {\color{gray} \, \pm 0.43}$ & 18.93 & $19.32 {\color{gray} \, \pm 0.21}$ \\
\hline
\multirow{5}{*}{\shortstack[l]{Penalized \\ $g \in [1:12]$ \\ {\color{gray} \textit{Layer-wise}}}}  & $\lambda_{pen}^g = 1$ & \multicolumn{5}{c}{\color{gray}{------ \;\;\; \textit{Failed \; due \; to \; sparsity \; collapse}} \;\;\; ------ } \\
& $\lambda_{pen}^g = 0.1$ & $30.80 {\color{gray} \, \pm 0.54}$ & $28.63 {\color{gray} \, \pm 0.52}$ & $19.30 {\color{gray} \, \pm 0.20}$ & 20.40 & $20.89 {\color{gray} \, \pm 0.29}$ \\
& $\lambda_{pen}^g = 0.01$ & $74.46 {\color{gray} \, \pm 0.41}$ & $73.90 {\color{gray} \, \pm 0.39}$ & $59.69 {\color{gray} \, \pm 0.36}$ & 19.03 & $19.52 {\color{gray} \, \pm 0.15}$ \\
& $\lambda_{pen}^g = 0.001$ & $89.38 {\color{gray} \, \pm 0.12}$ & $89.31 {\color{gray} \, \pm 0.17}$ & $87.42 {\color{gray} \, \pm 0.40}$ & 18.72 & $19.15 {\color{gray} \, \pm 0.33}$ \\
& $\lambda_{pen}^g = 0.0001$ & $90.55 {\color{gray} \, \pm 0.34}$ & $90.74 {\color{gray} \, \pm 0.34}$ & $89.81 {\color{gray} \, \pm 0.42}$ & 18.67 & $18.96 {\color{gray} \, \pm 0.22}$ \\
\hline
\multirow{5}{*}{\shortstack[l]{Constrained \\ {\color{gray} \textit{Model-wise}}}} & $\epsilon = 20 \%$ & $21.50 {\color{gray} \, \pm 0.19}$ & $20.10 {\color{gray} \, \pm 0.29}$ & $24.72 {\color{gray} \, \pm 0.31}$ & 20.42 & $20.81 {\color{gray} \, \pm 0.07}$ \\
& $\epsilon = 35 \%$ & $35.46 {\color{gray} \, \pm 0.01}$ & $34.82 {\color{gray} \, \pm 0.11}$ & $44.46 {\color{gray} \, \pm 0.76}$ & 19.29 & $19.88 {\color{gray} \, \pm 0.16}$ \\
& $\epsilon = 50 \%$ & $50.16 {\color{gray} \, \pm 0.01}$ & $49.63 {\color{gray} \, \pm 0.09}$ & $60.54 {\color{gray} \, \pm 0.88}$ & 19.25 & $19.65 {\color{gray} \, \pm 0.17}$ \\
& $\epsilon = 65 \%$ & $65.08 {\color{gray} \, \pm 0.04}$ & $64.64 {\color{gray} \, \pm 0.14}$ & $72.20 {\color{gray} \, \pm 0.60}$ & 18.86 & $19.22 {\color{gray} \, \pm 0.25}$ \\
& $\epsilon = 80 \%$ & $80.12 {\color{gray} \, \pm 0.03}$ & $79.82 {\color{gray} \, \pm 0.04}$ & $82.33 {\color{gray} \, \pm 0.46}$ & 18.91 & $19.16 {\color{gray} \, \pm 0.25}$ \\
\hline
\multirow{5}{*}{\shortstack[l]{Penalized \\ {\color{gray} \textit{Model-wise}}}} & $\lambda_{pen} = 1 $ & $22.37 {\color{gray} \, \pm 0.39}$ & $21.07 {\color{gray} \, \pm 0.48}$ & $28.49 {\color{gray} \, \pm 0.62}$ & 20.34 & $20.79 {\color{gray} \, \pm 0.32}$ \\ 
& $\lambda_{pen} = 0.1 $ & $71.90 {\color{gray} \, \pm 0.73}$ & $71.66 {\color{gray} \, \pm 0.71}$ & $74.22 {\color{gray} \, \pm 0.43}$ & 18.37 & $18.83 {\color{gray} \, \pm 0.47}$ \\
& $\lambda_{pen} = 0.01 $ & $88.65 {\color{gray} \, \pm 0.73}$ & $88.60 {\color{gray} \, \pm 0.80}$ & $88.39 {\color{gray} \, \pm 0.83}$ & 18.41 & $18.99 {\color{gray} \, \pm 0.37}$ \\
& $\lambda_{pen} = 0.001 $ & $90.46 {\color{gray} \, \pm 0.23}$ & $90.53 {\color{gray} \, \pm 0.45}$ & $90.06 {\color{gray} \, \pm 0.62}$ & 18.75 & $19.22 {\color{gray} \, \pm 0.16}$ \\
& $\lambda_{pen} = 0.0001 $ & $90.70 {\color{gray} \, \pm 0.33}$ & $90.83 {\color{gray} \, \pm 0.35}$ & $90.03 {\color{gray} \, \pm 0.32}$ & 19.02 & $19.42 {\color{gray} \, \pm 0.10}$ \\
\hline
\end{tabular}

}

\end{table}

\subsection{TinyImageNet}
\label{app:all_comparisons:tiny_imagenet}
\begin{figure}[h!]
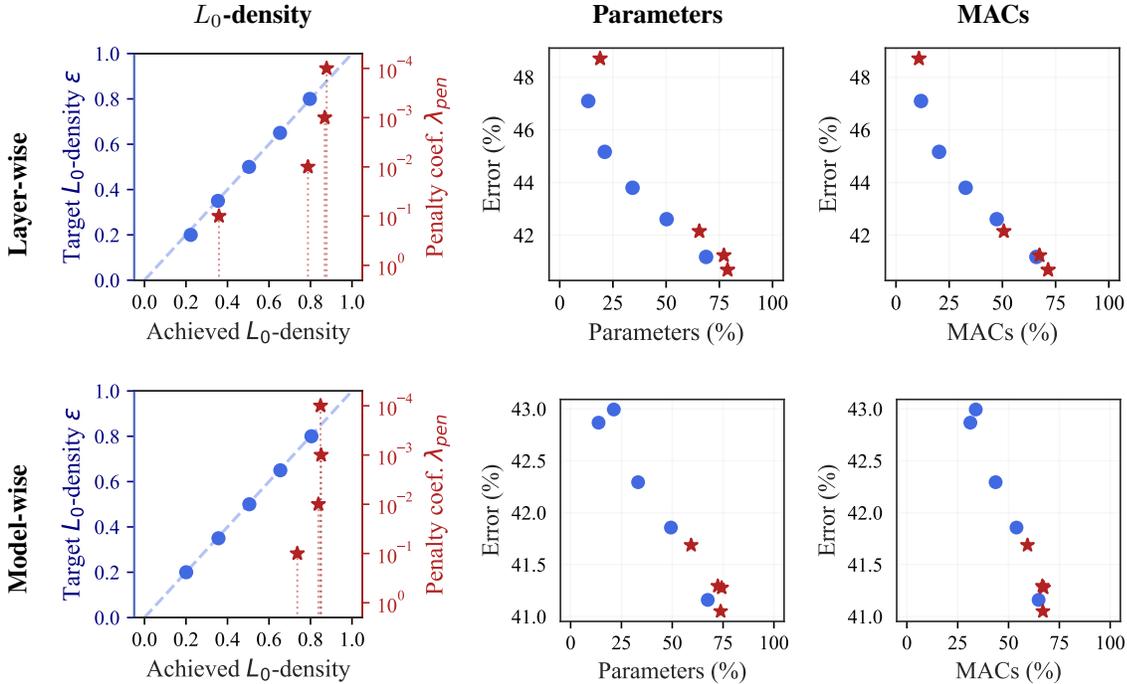

\setlength\tabcolsep{0pt}%
\hspace{-1.5cm}
\centering
\begin{tabular}{llccc}
 \centering
 & \hspace{2mm} & \hspace{-2mm} \textbf{$L_0$-density} & \hspace{6mm}\textbf{Parameters} & \hspace{5mm} \textbf{MACs} \\
 
\rotatebox{90}{\hspace{1.3cm} \textbf{Layer-wise}} & & \includegraphics[width=0.4\textwidth, trim={0mm 0mm 0mm 0mm}, clip]{neurips2022/figs/control/tiny_imagenet/L0ResNet18_layerwise.pdf}
 & \includegraphics[width=0.32\textwidth]{neurips2022/figs/error_vs_params/tiny_imagenet/L0ResNet18_layerwise_rebuttal.pdf}
 & \includegraphics[width=0.32\textwidth]{neurips2022/figs/error_vs_macs/tiny_imagenet/L0ResNet18_layerwise_rebuttal.pdf} \\
\rotatebox{90}{\hspace{1.3cm} \textbf{Model-wise}} &  & \includegraphics[width=0.4\textwidth, trim={0mm 0mm 0mm 0mm}, clip]{neurips2022/figs/control/tiny_imagenet/L0ResNet18_modelwise.pdf}
 & \includegraphics[width=0.32\textwidth]{neurips2022/figs/error_vs_params/tiny_imagenet/L0ResNet18_modelwise_rebuttal.pdf}
 & \includegraphics[width=0.32\textwidth]{neurips2022/figs/error_vs_macs/tiny_imagenet/L0ResNet18_modelwise_rebuttal.pdf}\\
\end{tabular}
    \caption[]{Training sparse ResNet18 models on TinyImageNet. \textit{This figure is the same as \cref{fig:tiny_imagenet_comparison}. We repeat it here for the reader's convenience.}}
    \label{fig:tiny_imagenet_APPENDIX_comparison}
\end{figure}

\clearpage
\vspace{-3cm}
\begin{table}[h!]

\centering
\renewcommand{\arraystretch}{1.1}

\caption{Achieved density levels and performance for sparse ResNet18 models trained on TinyImageNet for 120 epochs. Metrics aggregated over 3 runs.}
\vspace{1ex}

\resizebox{\textwidth}{!}{%

\begin{tabular}{llllllllllll}
\hline
\multirow{2}{*}{\textbf{Method}} & \multirow{2}{*}{\textbf{Hyper-params}} & \multirow{2}{*}{$L_0$-\textbf{density} (\%)} & \multirow{2}{*}{\textbf{Params} (\%)} & \multirow{2}{*}{\textbf{MACs} (\%)} & \multicolumn{2}{c}{\textbf{Val. Error } (\%)} \\
 & & & & & best & at 200 epochs \\
\hline
\hline
\multirow{5}{*}{\shortstack[l]{Constrained \\ $g \in [1:16]$ \\ {\color{gray} \textit{Layer-wise}}}} & $\epsilon_g = 20 \%$ & $22.99 {\color{gray} \, \pm 0.53}$ & $13.40 {\color{gray} \, \pm 0.54}$ & $11.75 {\color{gray} \, \pm 0.45}$ & 46.12  & $47.09 {\color{gray} \, \pm 0.15}$ \\
& $\epsilon_g = 35 \%$ & $35.41 {\color{gray} \, \pm 0.07}$ & $21.16 {\color{gray} \, \pm 0.03}$ & $20.25 {\color{gray} \, \pm 0.16}$ & 43.87 & $45.16 {\color{gray} \, \pm 0.31}$ \\
& $\epsilon_g = 50 \%$ & $50.40 {\color{gray} \, \pm 0.04}$ & $34.22 {\color{gray} \, \pm 0.28}$ & $32.73 {\color{gray} \, \pm 0.49}$ & 42.52 & $43.80 {\color{gray} \, \pm 0.44}$ \\
& $\epsilon_g = 65 \%$ & $65.32 {\color{gray} \, \pm 0.42}$ & $50.22 {\color{gray} \, \pm 0.54}$ & $47.34 {\color{gray} \, \pm 0.66}$ & 41.80 & $42.61 {\color{gray} \, \pm 0.10}$ \\
& $\epsilon_g = 80 \%$ & $79.68 {\color{gray} \, \pm 0.88}$ & $68.78 {\color{gray} \, \pm 1.70}$ & $65.98 {\color{gray} \, \pm 1.53}$ & 40.55 & $41.17 {\color{gray} \, \pm 0.61}$ \\
\hline
\multirow{5}{*}{\shortstack[l]{Penalized \\ $g \in [1:16]$ \\ {\color{gray} \textit{Layer-wise}}}}  & $\lambda_{pen}^g = 1$ & \multicolumn{5}{c}{\color{gray}{------ \;\;\; \textit{Failed \; due \; to \; sparsity \; collapse}} \;\;\; ------ } \\
& $\lambda_{pen}^g = 0.1$ & $35.90 {\color{gray} \, \pm 0.57}$ & $18.98 {\color{gray} \, \pm 0.57}$ & $10.77 {\color{gray} \, \pm 0.90}$ & 46.34 & $48.70 {\color{gray} \, \pm 0.73}$ \\
& $\lambda_{pen}^g = 0.01$ & $78.77 {\color{gray} \, \pm 0.84}$ & $65.59 {\color{gray} \, \pm 1.94}$ & $50.66 {\color{gray} \, \pm 1.77}$ & 41.36 & $42.15 {\color{gray} \, \pm 0.27}$ \\
& $\lambda_{pen}^g = 0.001$ & $86.87 {\color{gray} \, \pm 0.38}$ & $77.14 {\color{gray} \, \pm 1.29}$ & $67.45 {\color{gray} \, \pm 3.88}$ & 40.75 & $41.23 {\color{gray} \, \pm 0.58}$ \\
& $\lambda_{pen}^g = 0.0001$ & $87.77 {\color{gray} \, \pm 0.42}$ & $78.85 {\color{gray} \, \pm 0.91}$ & $71.49 {\color{gray} \, \pm 2.36}$ & 40.32 & $40.68 {\color{gray} \, \pm 0.04}$ \\
\hline
\multirow{5}{*}{\shortstack[l]{Constrained \\ {\color{gray} \textit{Model-wise}}}} & $\epsilon = 20 \%$ & $20.11 {\color{gray} \, \pm 0.03}$ & $13.75 {\color{gray} \, \pm 0.31}$ & $31.22 {\color{gray} \, \pm 1.25}$ & 42.02 & $42.87 {\color{gray} \, \pm 1.00}$ \\
& $\epsilon = 35 \%$ & $35.66 {\color{gray} \, \pm 0.12}$ & $21.25 {\color{gray} \, \pm 0.70}$ & $33.88 {\color{gray} \, \pm 1.25}$ & 41.28 & $42.99 {\color{gray} \, \pm 0.81}$ \\
& $\epsilon = 50 \%$ & $50.53 {\color{gray} \, \pm 0.09}$ & $33.19 {\color{gray} \, \pm 0.08}$ & $43.65 {\color{gray} \, \pm 0.94}$ & 40.70 & $42.29 {\color{gray} \, \pm 0.17}$ \\
& $\epsilon = 65 \%$ & $65.46 {\color{gray} \, \pm 0.19}$ & $49.34 {\color{gray} \, \pm 0.55}$ & $53.92 {\color{gray} \, \pm 0.97}$ & 41.23 & $41.86 {\color{gray} \, \pm 0.54}$ \\
& $\epsilon = 80 \%$ & $80.45 {\color{gray} \, \pm 0.19}$ & $67.45 {\color{gray} \, \pm 0.67}$ & $64.80 {\color{gray} \, \pm 1.18}$ & 40.71 & $41.16 {\color{gray} \, \pm 0.28}$ \\
\hline
\multirow{5}{*}{\shortstack[l]{Penalized \\ {\color{gray} \textit{Model-wise}}}} & $\lambda_{pen} = 1$ & \multicolumn{5}{c}{\color{gray}{------ \;\;\; \textit{Failed \; due \; to \; sparsity \; collapse}} \;\;\; ------ } \\
& $\lambda_{pen} = 0.1$ & $73.64 {\color{gray} \, \pm 0.72}$ & $59.27 {\color{gray} \, \pm 0.76}$ & $59.33 {\color{gray} \, \pm 2.19}$ & 40.84 & $41.69 {\color{gray} \, \pm 0.16}$ \\
& $\lambda_{pen} = 0.01$ & $83.80 {\color{gray} \, \pm 0.08}$ & $72.41 {\color{gray} \, \pm 0.67}$ & $66.64 {\color{gray} \, \pm 1.26}$ & 40.33 & $41.29 {\color{gray} \, \pm 0.50}$ \\
& $\lambda_{pen} = 0.001$ & $85.12 {\color{gray} \, \pm 0.89}$ & $74.31 {\color{gray} \, \pm 1.72}$ & $67.30 {\color{gray} \, \pm 1.62}$ & 40.71 & $41.28 {\color{gray} \, \pm 0.56}$ \\
& $\lambda_{pen} = 0.0001$ & $84.84 {\color{gray} \, \pm 0.90}$ & $73.80 {\color{gray} \, \pm 1.53}$ & $66.91 {\color{gray} \, \pm 1.65}$ & 40.57 & $41.05 {\color{gray} \, \pm 0.33}$ \\
\hline
\end{tabular}

}
\end{table}

\subsection{ImageNet}
\label{app:all_comparisons:imagenet}

\vspace{-2ex}

\begin{table}[h!]
\small
\centering
\renewcommand{\arraystretch}{1.0}

\caption{Sparse ResNet50 models on ImageNet. 
``Fine-tuning'' for zero epochs means \textit{no} fine-tuning. \textit{This table is the same as \cref{tab:mp_comparison_imagenet}. We repeat it here for the reader's convenience.}}

\label{tab:mp_comparison_imagenet_APPENDIX}
\vspace{1ex}

\resizebox{\textwidth}{!}{%
\begin{tabular}{ccccccccc}
\hline
\multirow{2}{*}{\textbf{Target}} & \multirow{3}{*}{\textbf{Method}}  & \multirow{2}{*}{$L_0$-\textbf{density}} & \multirow{2}{*}{\textbf{Params}} & \multirow{2}{*}{\textbf{MACs}} & \multicolumn{4}{c}{\textbf{Best Val. Error} (\%)} \\
\cline{6-9}
\multirow{2}{*}{\textbf{Density}} & & \multirow{2}{*}{(\%)} & \multirow{2}{*}{(\%)} & \multirow{2}{*}{(\%)} &  \multicolumn{4}{c}{After fine-tuning for \# epochs} \\
& & & & & 0 & 1 & 10 & 20 \\
\hline
\hline
  \multirow{2}{*}{$-$} & \multirow{2}{*}{Pre-trained Baseline} & \multirow{2}{*}{$100$} & \multirow{2}{*}{[25.5M]}  & \multirow{2}{*}{[$4.12 \cdot 10^9$]}  & {\multirow{2}{*}{$23.90$}} & \multicolumn{3}{c}{\multirow{2}{*}{{\color{gray} --------------------}}} \\
  & & & & & & & \\
 \hline
\multirow{6}{*}{$\epsilon = 90 \%$} & Constrained & \multirow{2}{*}{$90.36$} & \multirow{2}{*}{$88.06$} & \multirow{2}{*}{$91.62$} & {\multirow{2}{*}{$\textbf{24.68}$}} & \multicolumn{3}{c}{\multirow{2}{*}{{\color{gray} --------------------}}} \\
& {\color{gray} \textit{Model-wise}} & & & & & & & \\
\cline{2-9}
& Constrained & \multirow{2}{*}{$90.58$} & \multirow{2}{*}{$87.07$} & \multirow{2}{*}{$85.97$} & {\multirow{2}{*}{$24.97$}} & \multicolumn{3}{c}{\multirow{2}{*}{{\color{gray} --------------------}}} \\
& {\color{gray} \textit{Layer-wise}} & & & & & & & \\
\cline{2-9}
 & L1 - Mag. Prune & \multirow{2}{*}{$-$} & \multirow{2}{*}{$85.94$} & \multirow{2}{*}{$84.99$} & \multirow{2}{*}{$38.74$} & \multirow{2}{*}{$25.38$} & \multirow{2}{*}{$24.69$} &\multirow{2}{*}{$\textbf{24.68}$} \\
 & {\color{gray} \textit{Layer-wise}} & & & & & & & \\
\hline
\multirow{6}{*}{$\epsilon = 70 \%$} & Constrained & \multirow{2}{*}{$70.78$} & \multirow{2}{*}{$64.41$} & \multirow{2}{*}{$76.50$} & {\multirow{2}{*}{$\textbf{25.53}$}} & \multicolumn{3}{c}{\multirow{2}{*}{{\color{gray} --------------------}}} \\
& {\color{gray} \textit{Model-wise}} & & & & & & & \\
\cline{2-9}
& Constrained & \multirow{2}{*}{$70.36$} & \multirow{2}{*}{$61.91$} & \multirow{2}{*}{$58.59$} & {\multirow{2}{*}{$26.98$}} & \multicolumn{3}{c}{\multirow{2}{*}{{\color{gray} --------------------}}} \\
& {\color{gray} \textit{Layer-wise}} & & & & & & & \\
\cline{2-9}
& L1 - Mag. Prune & \multirow{2}{*}{$-$} & \multirow{2}{*}{$62.15$} & \multirow{2}{*}{$59.85$} & \multirow{2}{*}{$97.78$} & \multirow{2}{*}{$29.04$} & \multirow{2}{*}{$26.80$} & \multirow{2}{*}{$26.14$} \\
& {\color{gray} \textit{Layer-wise}} & & & & & & & \\
\hline
\multirow{6}{*}{$\epsilon = 50 \%$} & Constrained & \multirow{2}{*}{50.18} & \multirow{2}{*}{42.47} & \multirow{2}{*}{58.00} & {\multirow{2}{*}{\textbf{27.51}}} & \multicolumn{3}{c}{\multirow{2}{*}{{\color{gray} --------------------}}} \\
& {\color{gray} \textit{Model-wise}} & & & & & & & \\
\cline{2-9}
& Constrained & \multirow{2}{*}{50.70} & \multirow{2}{*}{43.15} & \multirow{2}{*}{38.25} & {\multirow{2}{*}{27.89}} & \multicolumn{3}{c}{\multirow{2}{*}{{\color{gray} --------------------}}}\\
& {\color{gray} \textit{Layer-wise}} & & & & & & & \\
\cline{2-9}
& L1 - Mag. Prune & \multirow{2}{*}{$-$} & \multirow{2}{*}{43.47} & \multirow{2}{*}{39.76} & \multirow{2}{*}{99.75} & \multirow{2}{*}{36.21} & \multirow{2}{*}{29.98} & \multirow{2}{*}{29.16} \\
& {\color{gray} \textit{Layer-wise}} & & & & & & & \\
\hline
\multirow{6}{*}{$\epsilon = 30 \%$} & Constrained & \multirow{2}{*}{30.31} & \multirow{2}{*}{31.81} & \multirow{2}{*}{42.05} & {\multirow{2}{*}{\textbf{29.65}}} & \multicolumn{3}{c}{\multirow{2}{*}{{\color{gray} --------------------}}} \\
& {\color{gray} \textit{Model-wise}} & & & & & & & \\
\cline{2-9}
& Constrained & \multirow{2}{*}{31.44} & \multirow{2}{*}{30.16} & \multirow{2}{*}{23.74} & {\multirow{2}{*}{31.71}} & \multicolumn{3}{c}{\multirow{2}{*}{{\color{gray} --------------------}}}\\
& {\color{gray} \textit{Layer-wise}} & & & & & & & \\
\cline{2-9}
& L1 - Mag. Prune & \multirow{2}{*}{$-$} & \multirow{2}{*}{29.86} & \multirow{2}{*}{24.80} & \multirow{2}{*}{99.89} & \multirow{2}{*}{56.11} & \multirow{2}{*}{36.90} & \multirow{2}{*}{34.74} \\
& {\color{gray} \textit{Layer-wise}} & & & & & & & \\
\hline
\end{tabular}
}
\end{table}

Here we provide further results for ResNet50 models on ImageNet, including experiments with model-wise constraints and the fine-tuned performance of the magnitude pruning method. We highlight that the controllability properties of our proposed constrained formulation extend to this large-scale setting (compare target density and $L_0$-density columns). The dense, pre-trained baseline, used as the starting point for magnitude pruning, corresponds to the \href{https://pytorch.org/vision/0.13/models/generated/torchvision.models.resnet50.html}{\texttt{ResNet50\_Weights.IMAGENET1K\_V1}} model made publicly available by Pytorch \citep{pytorch}.

As expected, model-wise constraints allow for a more flexible allocation of the parameter budget throughout the network, thus leading to a better validation error. However, this flexibility can also result in models with larger memory and computational footprints. For example, with $\epsilon = 70\%$, the model learned with model-wise constraints has a very similar parameter count ($64.41\%$ vs $61.19\%$) but a significantly higher MAC count ($76.50\%$ vs $58.59\%$).

The results on magnitude pruning confirm the importance of the fine-tuning stage for this technique. The accuracy improves dramatically after a few epochs of retraining compared to the ``just-pruned'' model. Rather than a ``pure'' pruning technique, one can think of magnitude pruning as a method that, given a pretrained model, provides an \textit{initialization} for a smaller model which needs to be trained (i.e. fine-tuned). At high density levels (e.g. 70-90\%) the performance of the constrained $L_0$ formulation and fine-tuned magnitude pruning methods are located within a similar range. However, for harsher sparsity levels (30-50\% density) the performance of models obtained using the constrained approach is significantly better than for magnitude pruning, even after fine-tuning.

\section{Unstructured Sparsity}
\label{app:unstructured}

In this section we demonstrate that our constrained approach transfers successfully between the structured and unstructured sparsity regimes without major modifications. We carry out experiments using MLP and convolutional models on MNIST, and ResNet18 models on TinyImageNet. 

Our unstructured experiments consider one gate \textit{per model parameter}. This means that the number of gates in the unstructured setting is much larger than in the structured one, since it scales with the total number of model parameters and not with the number of units/output maps. The $L_0$ density of a layer with unstructured sparsity corresponds to the expected number of active gates within that layer. 

We compare to a magnitude pruning baseline where a dense model is pre-trained, pruned in an unstructured way and fine-tuned. Since magnitude pruning is typically applied independently at each layer, we concentrate on experiments with layer-wise constraints.

\subsection{Experimental Setting}
\label{app:unstructured:experimental_setting}

Throughout this section, the model architectures we use for experiments with unstructured sparsity match those of structured experiments detailed in \cref{tab:model_stats}.

\subsubsection{MNIST}
\label{app:unstructured:mnist_exp}

\cref{tab:optim_mnist_unstructured} presents the hyper-parameters used for training MLP and LeNet models on MNIST with unstructured sparsity. We train using a batch size of 128 and do not apply weight decay.

\begin{table}[h]
    \vspace{-2.5ex}
    \renewcommand{\arraystretch}{1.25}
    \centering
    \caption{Configurations for MNIST experiments with unstructured sparsity.}
    \label{tab:optim_mnist_unstructured}
    \vspace{1ex}
    \resizebox{0.9\textwidth}{!}{%
    \begin{tabular}{cccccccc}
    \hline
      \multirow{2}{*}{\textbf{Approach}} & \multicolumn{2}{c}{\textbf{Weights}} & \multicolumn{2}{c}{\textbf{Gates}} & \multicolumn{3}{c}{\textbf{Lagrange Multipliers}} \\
      \cline{2-8} 
      & Optim. & $\weightlr$ & Optim. & $\gatelr$ & Optim. & $\duallr$ & Restarts \\ 
      \hline
      \hline
      Constrained  & Adam & $7 \cdot 10^{-4}$ & Adam & $1 \cdot 10^{-3}$ & Grad. Ascent & $10^{-3}$ & Yes \\ 
      Magnitude Pruning  & Adam & $7 \cdot 10^{-4}$ & - & - &  -  & - & - \\ 
      \hline
      
    \end{tabular}
    }
\end{table}

\textbf{Models with $L_0$ gates.}       
All the layers in these models have unstructured gates (one gate per weight entry and one gate per bias). The gate parameters are initialized using $\rho_{\text{init}} = 0.05$ (see  \cref{app:rho_init}). We train these models for 200 epochs.

\textbf{Magnitude pruning.} For our magnitude pruning experiments we first trained a fully dense model for 200 epochs. We then apply unstructured pruning with the pre-determined target density, and retrain the resulting sparse model for another 200 epochs. We apply magnitude pruning to each of the layers in these models. Our magnitude pruning implementation keeps the biases fully dense.

\subsubsection{TinyImageNet}
\label{app:unstructured:tiny_exp}

\cref{tab:optim_tiny_unstructured} presents the hyper-parameters used for learning sparse ResNet18 models on TinyImageNet. We use SGD with a momentum coefficient of 0.9 for the weights. The learning rate of the weights $\weightlr$ is multiplied by 0.1 at 30, 60 and 90 epochs. We use a batch size of 100.

\begin{table}[h]
    \vspace{-2.5ex}
    \renewcommand{\arraystretch}{1.25}
    \centering
    \caption{Default configurations for TinyImageNet experiments with unstructured sparsity.}
    \vspace{1ex}
    \label{tab:optim_tiny_unstructured}
    \resizebox{\textwidth}{!}{%
    \begin{tabular}{cccccccccc}
      \hline
     \multirow{2}{*}{\textbf{Approach}}  & \multicolumn{2}{c}{\textbf{Weights}} & \multicolumn{2}{c}{\textbf{Gates}} & \multicolumn{3}{c}{\textbf{Lagrange Multipliers}} & \textbf{Target} & \textbf{Weight} \\
    \cline{2-8} 
      &  Optim. & $\weightlr$ & Optim. & $\gatelr$ & Optim. & Restarts & $\duallr$  & \textbf{density} & \textbf{decay} \\ 
      \hline
      \hline
       \multirow{4}{*}{Constrained} & \multirow{4}{*}{SGDM} & \multirow{4}{*}{\textbf{$0.1$}} & \multirow{4}{*}{Adam} & \multirow{4}{*}{$3 \cdot 10^{-2}$} &  & \multirow{4}{*}{Yes} & $9 \cdot 10^{-5}$  & 20\% & \multirow{4}{*}{$5 \cdot 10^{-4}$} \\ 
      & & & & & Gradient & & $2 \cdot 10^{-4}$ & 10\% & \\
      & & & & & Ascent & & $7 \cdot 10^{-4}$ & 5\% & \\
      & & & & & & & $2 \cdot 10^{-3}$ & 1\% & \\
       \hline
      Magnitude Pruning  & SGDM & \textbf{$1 \cdot 10^{-4}$} & - & - & - & - & - & & $5 \cdot 10^{-4}$ \\ 
       \hline
    \end{tabular}
    }
\end{table}

\textbf{Models with $L_0$ gates.}       
The model's initial convolutional and final fully connected layers are kept fully dense.
The residual connection of each \texttt{BasicBlock} in the model is kept fully dense, while all other convolutional layers use unstructured sparsity. This results in 16 sparsifiable convolutional layers. The gate parameters are initialized using $\rho_{\text{init}} = 0.05$ (see  \cref{app:rho_init}) and optimized with Adam. 
We train these models for 120 epochs.

\textbf{Magnitude pruning.} For our magnitude pruning experiments we first trained a fully dense ResNet18 model for 120 epochs, using the weights learning rate schedule mentioned above. We then apply unstructured pruning with the pre-determined target density, and fine-tune the resulting sparse model for another 120 epochs using a fixed learning rate of $1 \cdot 10^{-4}$. We apply magnitude pruning to \textit{the same layers} that were sparsifiable in models with $L_0$ gates. Our magnitude pruning implementation keeps the biases fully dense.

\subsubsection{Gates Optimizer}

We originally tried the same optimization setup as with structured experiments (see \cref{app:exp_details:tiny_imagenet}) for our unstructured experiments on TinyImageNet. Note that in the unstructured setting, there is a significantly larger number of gates whose parameters need to be optimized. Moreover, the influence of each individual gate on the sparsity of a layer is much smaller compared to the structured experiments. Thus the learning rates for the gates and the dual variables required to be tuned for these new unstructured tasks.

It was difficult to find a value of the gates learning rate that allowed the gates to move appropriately: (1) models trained with small learning rates would not achieve any sparsity (see \cref{app:gates} for similar behavior in the structured sparsity regime); while (2) for larger gates learning rates we observed no decrease in model density for a long portion of training, followed by a sudden drop to the target density level. However, this sudden sparsification of the network caused a significant accuracy degradation. This behavior is consistent with the observations documented by \citet{gale2019state}.

We hypothesize that training models with \textit{unstructured} $L_0$ gates is highly susceptible to noise. 
In the unstructured case the information available for determining whether a gate should be active is mediated by its single associated parameter. 
Since we use mini-batch estimates of the model gradients, the training signal coming from this single parameter can be very noisy.  
In contrast, the structured setting has lower variance since each gate aggregates information across a large group of parameters.
Therefore, using an optimizer that is robust to this training noise is desirable.

We performed experiments with Adam as the optimizer for the model gates and this choice successfully delivered sparse models without breaking their predictive capacity. 

Note that the experiments reported by \citet{gale2019state} did not use an adaptive optimizer for the gates. Exploring whether an adaptive optimizer like Adam would be sufficient to resolve the shortcomings of the $L_0$ reparametrization framework of \citet{louizos2017learning} documented by \citet{gale2019state} is an interesting direction for future research.

\subsection{Training Dynamics}

In this section we explore the training dynamics of our proposed constrained formulation in the unstructured sparsity setting. We train a ResNet18 model on TinyImageNet with layer-wise constraints of 5\% density. For other experimental settings see \cref{app:unstructured:tiny_exp}.

\begin{figure*}[h]
    \centering
    \vspace{-2ex}
    \includegraphics[scale=0.725]{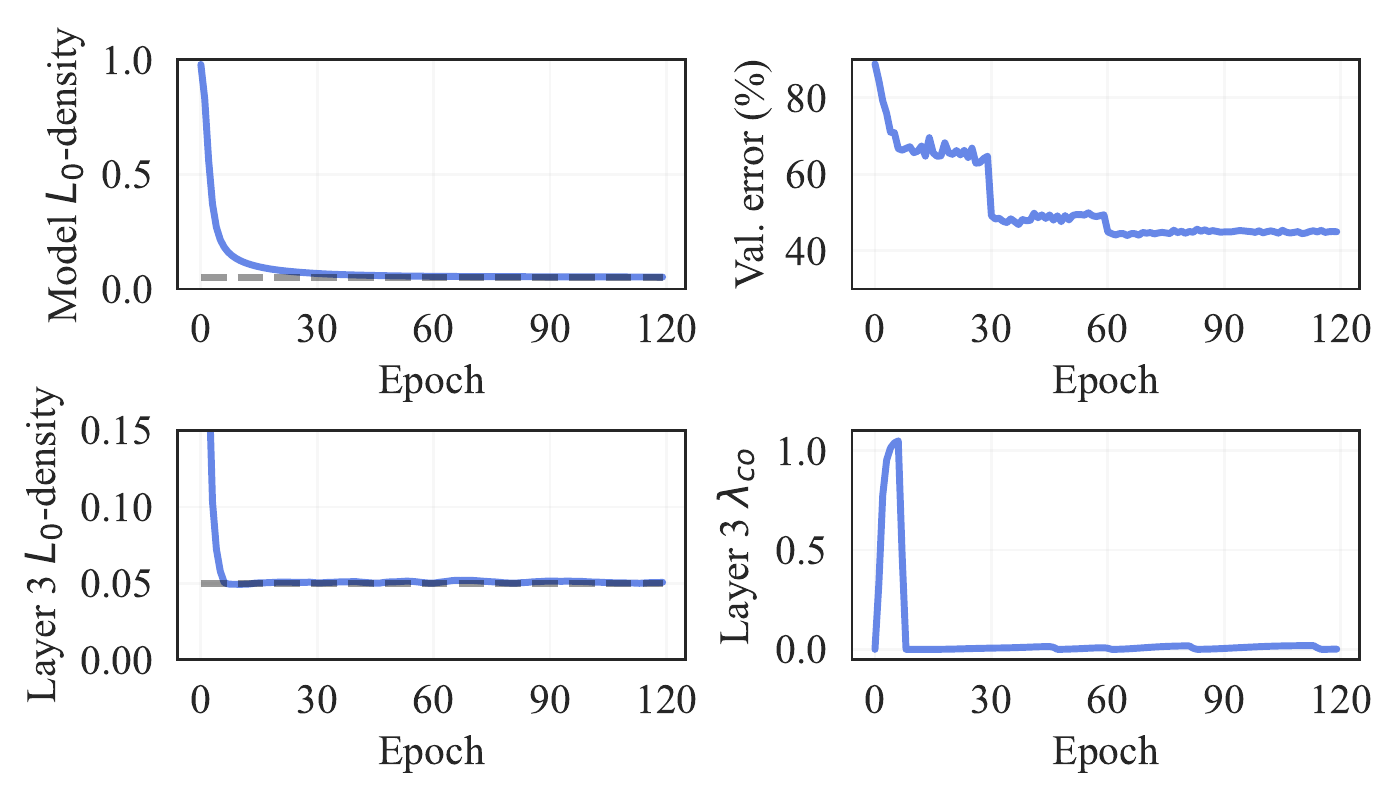}
    \vspace{-0.15in}
    \caption{
    Training dynamics for a ResNet18 model on TinyImageNet. This unstructured sparsity experiment uses layer-wise constraints with a target density of $5\%$. \textit{Layer 3} corresponds to the first convolutional layer of the second \texttt{BasicBlock} of the first "layer/stage" of the ResNet18 model.}
    \label{fig:unstructured_dynamics}
\end{figure*}

\cref{fig:unstructured_dynamics} shows the overall model density, the validation error, the density for a specific layer, and the Lagrange multiplier associated with the constraint for this layer. The behavior for the chosen layer is representative of that of other layers in the model.
The density at the model and layer levels decreases stably to the desired target. Note that a high sparsity of 95\% is achievable without causing irreparable damage to the model accuracy. 

Overall, the training dynamics of our unstructured experiments are qualitatively equivalent to those of the structured experiments presented in \cref{fig:dynamics}. This demonstrates that our constrained approach transfers successfully between structured and unstructured experiments.

\subsection{Performance Comparison}

\cref{tab:mnist_unstructured_comparison_mlp,tab:mnist_unstructured_comparison_lenet} present the sparsity and performance statistics of MLP and LeNet models trained on MNIST at different target densities. We also include a (100\%) dense model and layer-wise magnitude pruning experiments as baselines. For details on the experimental settings see \cref{app:unstructured:experimental_setting}.

\textbf{MNIST.} 
Our proposed constrained approach consistently outperforms magnitude pruning even after fine tuning the magnitude pruning models for 200 epochs. Our approach reliably produces models with the desired density for experiments at $20\%$, $10\%$, and $5\%$ density. The 1\% density setting is challenging for both methods. 

The constrained approach achieves high accuracy, although it incurs in a small violation of the sparsity constraint. 
Note that we employ the same dual learning rate across all densities. More extensive tuning of the dual learning rate can resolve this unfeasibility. 

On the other hand, magnitude pruning experiments achieve the desired density at 1\% (by design) but drastically fail in terms of performance. 
Note that the accuracy for magnitude pruning does not improve to an acceptable level even after fine-tuning for a large number of epochs. This behavior can be explained by the fact that our MNIST models have some layers with very few parameters. For example, at a 1\% sparsity, the first convolutional layer of the model has only 5 active parameters.

The very low number of parameters make the high-sparsity pruned models difficult to fine-tune. This observation is consistent with the poor gradient dynamics reported by \citet{evci2022GradientFlowa} when training highly sparse networks. We would like to highlight that our proposed method applies sparsity to the same layers as magnitude pruning, yet achieves high levels of unstructured sparsity per layer with minimal accuracy reduction. This shows that very sparse models with high accuracy do exist, however they seem to be out of reach when simply fine-tuning a magnitude-pruned model.

\begin{table}[h!]
\small
\centering
\renewcommand{\arraystretch}{1.1}

\caption{MLP models trained with unstructured sparsity on MNIST. Unstructured magnitude pruning is applied independently at each layer to retain the desired target density. ``Fine-tuning'' for zero epochs means \textit{no} fine-tuning. Metrics averaged across 3 runs.}

\label{tab:mnist_unstructured_comparison_mlp}
\vspace{1ex}

\resizebox{0.9\textwidth}{!}{%
\begin{tabular}{ccccccc}
\hline
\multirow{2}{*}{\textbf{Target Density}} & \multirow{3}{*}{\textbf{Method}}  & \multirow{3}{*}{$L_0$-\textbf{density} (\%)} & \multicolumn{4}{c}{\textbf{Best Val. Error} (\%)} \\
\cline{4-7}
\multirow{2}{*}{$g \in [1:3]$} & &  &  \multicolumn{4}{c}{After fine-tuning for \# epochs} \\
& & & 0 & 50 & 100 & 200 \\
\hline
\hline
  \multirow{1}{*}{$-$} & \multirow{1}{*}{Dense Baseline} & \multirow{1}{*}{$100.00$} & {\multirow{1}{*}{$1.72$}} & \multicolumn{3}{c}{\multirow{1}{*}{{\color{gray} --------------}}} \\
 \hline
\multirow{2}{*}{$\epsilon_g = 20 \%$} & Constrained & \multirow{1}{*}{$20.00$} & \multirow{1}{*}{$1.45$} & \multicolumn{3}{c}{\multirow{1}{*}{{\color{gray} --------------}}} \\
 & Magnitude Pruning & \multirow{1}{*}{$-$} & \multirow{1}{*}{$3.81$} & \multirow{1}{*}{$2.03$} & \multirow{1}{*}{$1.93$} &\multirow{1}{*}{$1.89$} \\
\hline
\multirow{2}{*}{$\epsilon_g = 10 \%$} & Constrained & \multirow{1}{*}{$10.02$} & {\multirow{1}{*}{$1.51$}} & \multicolumn{3}{c}{\multirow{1}{*}{{\color{gray} --------------}}} \\
& Magnitude Pruning & \multirow{1}{*}{$-$} & \multirow{1}{*}{$9.31$} & \multirow{1}{*}{$2.63$} & \multirow{1}{*}{$2.57$} & \multirow{1}{*}{$2.48$} \\
\hline
\multirow{2}{*}{$\epsilon_g = 5 \%$} & Constrained & \multirow{1}{*}{$5.05$} & {\multirow{1}{*}{$1.64$}} & \multicolumn{3}{c}{\multirow{1}{*}{{\color{gray} --------------}}} \\
& Magnitude Pruning & \multirow{1}{*}{$-$} & \multirow{1}{*}{$30.68$} & \multirow{1}{*}{$3.69$} & \multirow{1}{*}{$3.69$} & \multirow{1}{*}{$3.69$} \\
\hline
\multirow{2}{*}{$\epsilon_g = 1 \%$} & Constrained & \multirow{1}{*}{$2.62$} & {\multirow{1}{*}{$1.92$}} & \multicolumn{3}{c}{\multirow{1}{*}{{\color{gray} --------------}}} \\
& Magnitude Pruning & \multirow{1}{*}{$-$} & \multirow{1}{*}{$90.45$} & \multirow{1}{*}{$60.82$} & \multirow{1}{*}{$60.69$} & \multirow{1}{*}{$55.45$} \\
\hline
\end{tabular}
}
\end{table}

\begin{table}[h!]
\small
\centering
\renewcommand{\arraystretch}{1.1}

\caption{LeNet models trained with unstructured sparsity on MNIST. Unstructured magnitude pruning is applied independently at each layer to retain the desired target density. ``Fine-tuning'' for zero epochs means \textit{no} fine-tuning. Metrics averaged across 3 runs.}

\label{tab:mnist_unstructured_comparison_lenet}
\vspace{1ex}

\resizebox{0.9\textwidth}{!}{%
\begin{tabular}{ccccccc}
\hline
\multirow{2}{*}{\textbf{Target Density}} & \multirow{3}{*}{\textbf{Method}}  & \multirow{3}{*}{$L_0$-\textbf{density} (\%)} & \multicolumn{4}{c}{\textbf{Best Val. Error} (\%)} \\
\cline{4-7}
\multirow{2}{*}{$g \in [1:4]$} & &  &  \multicolumn{4}{c}{After fine-tuning for \# epochs} \\
& & & 0 & 50 & 100 & 200 \\
\hline
\hline
  \multirow{1}{*}{$-$} & \multirow{1}{*}{Dense Baseline} & \multirow{1}{*}{$100.00$} & {\multirow{1}{*}{$0.85$}} & \multicolumn{3}{c}{\multirow{1}{*}{{\color{gray} --------------}}} \\
 \hline
\multirow{2}{*}{$\epsilon_g = 20 \%$} & Constrained & \multirow{1}{*}{$19.99$} & \multirow{1}{*}{$0.73$} & \multicolumn{3}{c}{\multirow{1}{*}{{\color{gray} --------------}}} \\
 & Magnitude Pruning & \multirow{1}{*}{$-$} & \multirow{1}{*}{$2.28$} & \multirow{1}{*}{$0.98$} & \multirow{1}{*}{$0.92$} &\multirow{1}{*}{$0.92$} \\
\hline
\multirow{2}{*}{$\epsilon_g = 10 \%$} & Constrained & \multirow{1}{*}{$10.00$} & {\multirow{1}{*}{$0.78$}} & \multicolumn{3}{c}{\multirow{1}{*}{{\color{gray} --------------}}} \\
& Magnitude Pruning & \multirow{1}{*}{$-$} & \multirow{1}{*}{$5.23$} & \multirow{1}{*}{$1.38$} & \multirow{1}{*}{$1.38$} & \multirow{1}{*}{$1.32$} \\
\hline
\multirow{2}{*}{$\epsilon_g = 5 \%$} & Constrained & \multirow{1}{*}{$5.01$} & {\multirow{1}{*}{$0.89$}} & \multicolumn{3}{c}{\multirow{1}{*}{{\color{gray} --------------}}} \\
& Magnitude Pruning & \multirow{1}{*}{$-$} & \multirow{1}{*}{$12.53$} & \multirow{1}{*}{$2.39$} & \multirow{1}{*}{$2.39$} & \multirow{1}{*}{$2.39$} \\
\hline
\multirow{2}{*}{$\epsilon_g = 1 \%$} & Constrained & \multirow{1}{*}{$1.55$} & {\multirow{1}{*}{$1.26$}} & \multicolumn{3}{c}{\multirow{1}{*}{{\color{gray} --------------}}} \\
& Magnitude Pruning & \multirow{1}{*}{$-$} & \multirow{1}{*}{$88.76$} & \multirow{1}{*}{$88.76$} & \multirow{1}{*}{$88.76$} & \multirow{1}{*}{$88.76$} \\
\hline
\end{tabular}
}
\end{table}

\vspace*{2ex}

\textbf{TinyImageNet.} \cref{tab:tiny_unstructured_comparison} displays the result for TinyImageNet experiments at 1\%, 5\%, 10\% and 20\% unstructured sparsity. We observe similar patterns as in the MNIST experiments: the constrained approach reliably achieves the desired sparsity targets and preserves reasonable performance.

\begin{table}[h!]
\small
\centering
\renewcommand{\arraystretch}{1.1}

\caption{ResNet18 models trained with unstructured sparsity on TinyImageNet. ``Fine-tuning'' for zero epochs means \textit{no} fine-tuning.}

\label{tab:tiny_unstructured_comparison}
\vspace{1ex}

\resizebox{0.9\textwidth}{!}{%
\begin{tabular}{ccccccc}
\hline
\multirow{2}{*}{\textbf{Target Density}} & \multirow{3}{*}{\textbf{Method}}  & \multirow{3}{*}{$L_0$-\textbf{density} (\%)} & \multicolumn{4}{c}{\textbf{Best Val. Error} (\%)} \\
\cline{4-7}
\multirow{2}{*}{$g \in [1:16]$} & &  &  \multicolumn{4}{c}{After fine-tuning for \# epochs} \\
& & & 0 & 40 & 80 & 120 \\
\hline
\hline
  \multirow{1}{*}{$-$} & \multirow{1}{*}{Dense Baseline} & \multirow{1}{*}{$100.00$} & {\multirow{1}{*}{$38.64$}} & \multicolumn{3}{c}{\multirow{1}{*}{{\color{gray} --------------}}} \\
 \hline
\multirow{2}{*}{$\epsilon_g = 20 \%$} & Constrained & \multirow{1}{*}{$20.26$} & \multirow{1}{*}{$42.06$} & \multicolumn{3}{c}{\multirow{1}{*}{{\color{gray} --------------}}} \\
 & Magnitude Pruning & \multirow{1}{*}{$-$} & \multirow{1}{*}{$43.45$} & \multirow{1}{*}{$39.81$} & \multirow{1}{*}{$39.35$} &\multirow{1}{*}{$39.27$} \\
\hline
\multirow{2}{*}{$\epsilon_g = 10 \%$} & Constrained & \multirow{1}{*}{$10.57$} & {\multirow{1}{*}{$42.54$}} & \multicolumn{3}{c}{\multirow{1}{*}{{\color{gray} --------------}}} \\
& Magnitude Pruning & \multirow{1}{*}{$-$} & \multirow{1}{*}{$54.06$} & \multirow{1}{*}{$42.19$} & \multirow{1}{*}{$41.45$} & \multirow{1}{*}{$41.25$} \\
\hline
\multirow{2}{*}{$\epsilon_g = 5 \%$} & Constrained & \multirow{1}{*}{$5.20$} & {\multirow{1}{*}{$43.98$}} & \multicolumn{3}{c}{\multirow{1}{*}{{\color{gray} --------------}}} \\
& Magnitude Pruning & \multirow{1}{*}{$-$} & \multirow{1}{*}{$75.44$} & \multirow{1}{*}{$46.00$} & \multirow{1}{*}{$44.21$} & \multirow{1}{*}{$43.75$} \\
\hline
\multirow{2}{*}{$\epsilon_g = 1 \%$} & Constrained & \multirow{1}{*}{$1.87$} & {\multirow{1}{*}{$47.24$}} & \multicolumn{3}{c}{\multirow{1}{*}{{\color{gray} --------------}}} \\
& Magnitude Pruning & \multirow{1}{*}{$-$} & \multirow{1}{*}{$99.21$} & \multirow{1}{*}{$81.67$} & \multirow{1}{*}{$72.95$} & \multirow{1}{*}{$69.00$} \\
\hline
\end{tabular}
}
\end{table}

\newpage

In this task magnitude pruning outperforms the constrained approach, except for the case of 1\% density. We hypothesize that the improvement of magnitude pruning at relatively larger density targets (10\% and 20\%) may be linked to the fact that the ResNet18 model is significantly larger than those models used for MNIST, and thus easier to fine-tune.

Finally, note that the performance of the constrained approach in the harsh 1\% density setting is significantly better (albeit with a small violation of the constraint target) than that of magnitude pruning, even after fine-tuning the magnitude pruning model for 120 epochs.

\end{document}